%% file: main.tex
\documentclass[10pt,twocolumn,letterpaper]{article}
\pdfoutput=1
\ifdefined\pdfsuppressptexinfo\pdfsuppressptexinfo=-1\fi
\usepackage[T1]{fontenc}

\usepackage[pagenumbers]{preprint}

\input{preamble}

\input{macros}

\definecolor{pprblue}{rgb}{0.21,0.49,0.74}
\usepackage[pagebackref,breaklinks,colorlinks,allcolors=pprblue]{hyperref}

\title{From a Billion Poses to One: \\ Global Registration of 2D Histology to MRI Volumes}

\author{Stefan Brandst\"atter$^{1,2,3,4}$ \quad Lin Tian$^{5}$ \quad Adri\`a Casamitjana$^{6}$ \quad Alexandra G. Roberts$^{4}$\\
Tonmoy Hossain$^{4}$ \quad Helmut Prosch$^{2,7}$ \quad Georg Langs$^{1,2,3,\dagger}$ \quad Juan Eugenio Iglesias$^{4,8,9,\dagger}$\\[4pt]
\small $^{1}$Computational Imaging Research Lab, Medical University of Vienna, Vienna, Austria\\
\small $^{2}$Christian Doppler Laboratory for Machine Learning Driven Precision Imaging, Medical University of Vienna, Vienna, Austria\\
\small $^{3}$Comprehensive Center for Artificial Intelligence in Medicine, Medical University of Vienna, Vienna, Austria\\
\small $^{4}$Athinoula A. Martinos Center for Biomedical Imaging,\\
\small Massachusetts General Hospital and Harvard Medical School, Boston, MA, USA\\
\small $^{5}$Hong Kong Baptist University, Hong Kong\\
\small $^{6}$Universitat de Girona, Girona, Spain\\
\small $^{7}$Division of General and Pediatric Radiology, Medical University of Vienna, Vienna, Austria\\
\small $^{8}$Department of Medical Physics and Biomedical Engineering, University College London, London, UK\\
\small $^{9}$Computer Science and Artificial Intelligence Laboratory, Massachusetts Institute of Technology, Cambridge, MA, USA\\[2pt]
\small $^{\dagger}$Shared last authorship}

\begin{document}
\maketitle

\input{sections/00_abstract}

\input{sections/01_introduction}
\input{sections/02_related}
\input{sections/04_method}

\input{sections/03_setup}

\input{sections/05_nextbrain}

\input{sections/06_adni}
\input{sections/07_analysis}

\input{sections/08_limitations}
\input{sections/08b_llm_usage}

Data collection and sharing for the Alzheimer's Disease Neuroimaging Initiative (ADNI) is funded by the National Institute on Aging (National Institutes of Health Grant U19AG024904). The grantee organization is the Northern California Institute for Research and Education. In the past, ADNI has also received funding from the National Institute of Biomedical Imaging and Bioengineering, the Canadian Institutes of Health Research, and private sector contributions through the Foundation for the National Institutes of Health (FNIH) including generous contributions from the following: AbbVie, Alzheimer's Association, Alzheimer's Drug Discovery Foundation, Araclon Biotech, BioClinica, Inc., Biogen, Bristol-Myers Squibb Company, CereSpir, Inc., Cogstate, Eisai Inc., Elan Pharmaceuticals, Inc., Eli Lilly and Company, EuroImmun, F. Hoffmann-La Roche Ltd and its affiliated company Genentech, Inc., Fujirebio, GE Healthcare, IXICO Ltd., Janssen Alzheimer Immunotherapy Research \& Development, LLC., Johnson \& Johnson Pharmaceutical Research \& Development LLC., Lumosity, Lundbeck, Merck \& Co., Inc., Meso Scale Diagnostics, LLC., NeuroRx Research, Neurotrack Technologies, Novartis Pharmaceuticals Corporation, Pfizer Inc., Piramal Imaging, Servier, Takeda Pharmaceutical Company, and Transition Therapeutics.

{
    \small
    \bibliography{references}
}

\clearpage
\makeappendixtitle
\appendix
\input{sections/10_appendix}

\input{sections/09_statements}

\end{document}

%% file: preamble.tex
\usepackage{cuted}

\usepackage{currfile}

\usepackage{caption}

%% file: macros.tex
\usepackage{graphicx}
\usepackage{booktabs}
\usepackage{array}
\usepackage{xcolor}
\usepackage{enumitem}
\usepackage{amsmath,amssymb}
\usepackage{calc}
\graphicspath{{figs/}}

%% file: sections/00_abstract.tex
\begin{abstract}
Localizing 2D histological sections within 3D magnetic resonance images (MRI) could transform neuropathology and brain banking by enabling spatial population analyses on a common 3D coordinate frame. 
This is a challenging problem due to differences in image appearance, tissue deformation, and the vast 6-DoF search space. Existing methods typically operate within a single modality or require pose initialization.  
We present an initialization-free, cross-modal method that recovers the full 6-DoF pose of a histological section in an MRI of the same specimen. 
Our approach uses two lightweight encoders to extract histology and MRI features, scoring candidate poses by comparing histological features with planar samples of the MRI feature volume, encoded only once. 
Crucially, the score is linear in the encoded volume, enabling efficient evaluation of $5.4{\times}10^{8}$ candidate poses per section using fast Fourier transforms. 
We train the encoders contrastively with hard negatives mined by the same global search. 
When multiple parallel sections are available, we aggregate their evidence to estimate a shared orientation (``block mode'').
Experiments on synthetic sections 
demonstrate robustness to arbitrary orientations and to random in-plane displacement fields of up to 3\,mm RMS.
On a real histological dataset (NextBrain),
 86.7\% of sections stained with LFB and 78.8\% with H\&E are localized within 5$^\circ$ and 2\,mm of the reference pose in block mode, outperforming a state-of-the-art, training-free approach (MIND-SSC) by 20 points on LFB. The single-section mode achieves 71.7\% and 63.1\%, respectively.    
Our code and trained models will be made publicly available. 
\par\vspace{0pt plus 2\baselineskip}
\end{abstract}

%% file: sections/01_introduction.tex
\section{Introduction}\label{sec:intro}
\begin{figure*}[!t]
\centering
\includegraphics[width=\textwidth]{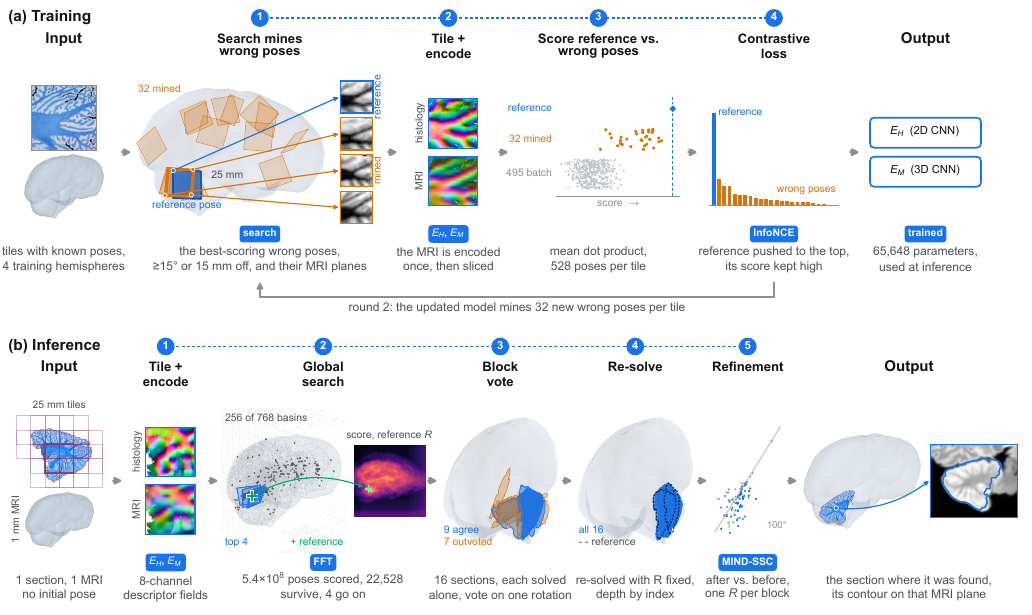}
\caption{\textbf{Method overview.} Every pose is drawn inside the MRI volume with an arrow to its MRI plane. \textbf{(a) Training}, for one real tile: a contrastive loss ranks the reference pose above the best-scoring wrong poses the search mines. \textbf{(b) Inference}, the steps of Section~\ref{sec:inference}.}
\label{fig:teaser}
\end{figure*}
Human brain banks use histology to assess brain tissue at cellular resolution: a block cut from the brain is sliced into thin sections, stained, and photographed under a microscope. Cutting discards each section's 3D context: its position and orientation in the brain~\cite{pichat2018survey,luciano2016printing}.  Brain banks sample a few small blocks from standard regions~\cite{montine2012niaaa,webster2021sampling} and record each sampling location only approximately, by gross anatomical landmarks, or often not at all~\cite{webster2021sampling,webster2026atlases}.  Localizing a section in an MRI volume of the same specimen restores this context by estimating the pose of its cutting plane.  Further registering the MRI to a population atlas~\cite{iglesias2023easyreg,casamitjana2025nextbrain}, a common coordinate frame, enables cross-sectional analysis in one bank or across banks~\cite{webster2026atlases}.  This supports pathology-MRI association studies~\cite{athalye2025sampling}, the construction of histological atlases~\cite{casamitjana2025nextbrain}, and neuropathology mapping, as done for tau tangles, a hallmark of Alzheimer's disease~\cite{montine2012niaaa,ortegacruz2024path2mr}, in a study of 18 donors~\cite{yushkevich2021tangles}.     

Despite its utility, 3D histology remains rare outside a few select sites~\cite{mancini2020pipeline,ortegacruz2024path2mr}. Their pipelines constrain the section poses with photographs of the block's cut face (blockface photographs), cutting molds that fix the cutting planes, or manual initialization~\cite{casamitjana2025nextbrain,huszar2023tirl,athalye2026joint,absinta2014cuttingbox}.   General slice-to-volume methods, which register a 2D image to a 3D volume, do not eliminate these prerequisites: most operate within a single modality~\cite{brandstaetter2024slivreg,frolov2025needles,xu2022svort} or require an initial pose~\cite{ferrante2017svr}.  We aim to enable scalable and comparable 3D histology analysis in brain banks by automating histology-to-MRI registration without user interaction.  

Three issues make the task challenging.  First, without an initial pose, the search must cover all six degrees of freedom (three rotations, three translations).  In X-ray to CT~\cite{gopalakrishnan2024diffpose,gopalakrishnan2026xvr,chen2026spherical,xpos3r2026}, the clinical setup can bound them and the search remains near its initialization~\cite{markelj2012review,unberath2021impact}.   Serial histology of a whole brain or hemisphere~\cite{amunts2013bigbrain,carey2023deepslice,ding2016allen} also provides an initialization, since each section's depth, i.e., its position along the stack, follows from its cutting order.  However, in brain banks, serial whole-brain histology is rare~\cite{ortegacruz2024path2mr,webster2021sampling}.      

Second, the similarity between a section and an MRI plane must be learned, because no current physical model renders a stained section from MRI.  One section alone weakly constrains its own cutting plane~\cite{ferrante2017svr}. Its out-of-plane rotation (tilt) and depth are ambiguous near the optimum, as anatomy varies slowly along the plane normal.       

Third, cutting and mounting deform, fold, and tear the tissue~\cite{iglesias2018jointsynth}, so no human brain MRI-histology dataset is registered without error.    Even NextBrain~\cite{casamitjana2025nextbrain}, built to minimize this distortion, incurs registration errors of about 1\,mm.  Our main results use its released sections, which are already warped onto the MRI.  

In this paper, we propose a fully automatic method that maps histology sections to MRI volumes (Figure~\ref{fig:teaser}). Our specific contributions are:   
\begin{enumerate}[wide=0pt, label=(\arabic*)]
\item We develop a global 6-DoF histology-to-MRI registration method that requires no pose initialization.  Two encoders are trained on LFB and H\&E simultaneously and extract directly comparable features from histology and from MRI. We call these features descriptors and the two encoders together the dual encoder. We train it with a contrastive loss so that the correct pose scores above the wrong poses that the global search finds.
\item We identify the sources of its efficiency and accuracy (Section~\ref{sec:serial}).  The global search is tractable because the MRI is encoded once and the score is linear in the encoded MRI. At each rotation, the fast Fourier transform (FFT) therefore scores all translations in a single pass.     When multiple parallel sections of one tissue block are available (block mode), they vote on their shared orientation, following DeepSlice~\cite{carey2023deepslice}.   The search then narrows to translations at that orientation, and the established modality-independent descriptor MIND-SSC~\cite{heinrich2012mind,heinrich2013mindssc} refines the poses.  
\item We characterize its limits.  Controlled experiments on synthetic sections measure how accuracy falls as deformation grows. Native slides are scans NextBrain did not warp, and we mask and rescale them ourselves and measure how much accuracy drops (Section~\ref{sec:mainresults}).
\end{enumerate}

%% file: sections/02_related.tex
\section{Related work}\label{sec:related}
\noindent\textbf{Slice-to-volume registration.} Most methods register slices within a single modality or start from a nearby pose~\cite{ferrante2017svr}, as in SVoRT~\cite{xu2022svort} and EUReg~\cite{eureg2025}.  SLIV-Reg~\cite{brandstaetter2024slivreg} is global and requires no initial pose, but is same-modality and estimates only the rotation in a known subvolume.   LoFTR-S2V~\cite{frolov2025needles} matches learned features between the whole section, taken as one 128-pixel image, and a volume of the same modality. The global multimodal method of Markova et~al.~\cite{markova2022global} recovers a coarse pose of tracked ultrasound sweeps from keypoint correspondences, which require local features visible in both modalities.  In computer vision, image-to-point-cloud registration~\cite{li2025implicit} and visual localization~\cite{mao2026color,cheng2026pilot,sarlin2021pixloc} find the 6-DoF pose of an image in a 3D scene from 2D-3D correspondences or near a pose prior. Object pose networks rank rendered candidate poses within one modality~\cite{labbe2022megapose,wen2024foundationpose}, but no model renders a stained section from MRI. We match across modalities with no pose prior or keypoints.   

\noindent\textbf{Section-to-atlas placement.} Regressors such as SVRnet~\cite{hou2018svr,salehi2019svr} and the ultrasound method of Yeung et~al.~\cite{yeung2021planeinvol} require no initial pose and predict a position in an atlas learned from many aligned volumes of one modality.  DeepSlice~\cite{carey2023deepslice} does the same for coronal mouse sections, and Fang and Tward~\cite{fang2026atlas} for mouse and marmoset. We consider this case easier than ours, because the sections are usually cut along a standard atlas plane with small deviations~\cite{puchades2019quicknii} and DeepSlice is trained on hundreds of thousands of real whole-brain sections aligned to the atlas.  DeepSlice's authors advise against aligning single images and average the two cutting angles over a series, and Xiong et~al.~\cite{xiong2018mapping} fit both angles to a whole series, the published precedents of our block vote.  Our target is instead the specimen's own MRI. An MRI-to-atlas registration~\cite{iglesias2023easyreg,hoffmann2022synthmorph} can then map the section into an atlas, as in~\cite{yushkevich2021tangles}.   

\noindent\textbf{Histology to MRI.} NextBrain~\cite{casamitjana2025nextbrain} and other 3D-histology pipelines~\cite{wang2025bbq,goubran2013histomri} rely on a photograph of every cut face of the block~\cite{mancini2020pipeline} or on manual initialization, which takes time and anatomical knowledge~\cite{alegro2016multimodal,yushkevich2021tangles,huszar2023tirl}.  The closest pipelines for human tissue register sections deformably to postmortem MRI from a nearby initial pose~\cite{osechinskiy2011slice,tward2020diffeomorphic}.  TIRL~\cite{huszar2023tirl} relies on blockface photographs and a 2\,cm initial brain slab set manually.  Athalye et al.~\cite{athalye2026joint} align each pair from an automatic initialization inside a 1\,cm slab that a cutting mold fixes, and such a mold must be printed for each brain~\cite{absinta2014cuttingbox,luciano2016printing}.  Their method requires this slab, so it is not a baseline here.  Our method estimates only a rigid pose and uses none of these aids, but our test sections come from NextBrain's reconstruction.   Block mode takes sections already aligned in plane~\cite{ourselin2001reconstructing}, as stack-to-volume registration does~\cite{malandain2004fusion}, but reconstructs no volume from them.   

\noindent\textbf{Modality-invariant similarity.} MIND-SSC~\cite{heinrich2012mind,heinrich2013mindssc}, CoMIR~\cite{pielawski2020comir}, and learned similarities~\cite{simonovsky2016deep} and descriptors such as DISA~\cite{ronchetti2023disa} narrow the gap between modalities.  DISA imitates the training-free LC2 similarity~\cite{wein2008lc2}, a local linear fit of one modality to the other, with a dot product to keep the search computationally tractable, and has initialized histology-to-micro-CT registration from a known cutting face~\cite{chen2025disahisto}.   For bone, Chicherova et al.~\cite{chicherova2014histology} place histological sections in micro-CT from keypoints matched to every slice. Lundin et al.~\cite{bone2017histo2microct} place them with no prior on the cutting direction, but only after segmenting bone in both modalities, which reduces the search to a same-modality problem.   Pretrained networks and a classical solver won a PET/CT challenge by matching CT to CT and PET to PET~\cite{dey2026representation}, also a same-modality pairing.   We learn descriptors for the global search, where MIND-SSC accuracy drops sharply cross-modally (Table~\ref{tab:baselines_untouched}), and keep MIND-SSC for the local refinement.  

\noindent\textbf{Global FFT search over poses.} Protein docking~\cite{katchalski1992molecular} and cryo-electron tomography~\cite{frangakis2002identification} enumerate rotations and search translations of a 3D body by FFT, and Equivariant Scalar Fields (ESF)~\cite{jing2024esf} learns the two fields this search correlates, for docking.   A globally optimal search also recovers pose from silhouettes, with no initial pose~\cite{sengupta2026silhouettes}. Map localization learns embeddings through an FFT search over position and heading in a 2D map~\cite{barsan2018lidar,sarlin2023orienternet,chen2020dpcn}.  We apply this search to a 2D section as a plane through a once-encoded 3D volume, over 3D rotations and translations, and mine negatives with it. 

%% file: sections/04_method.tex
\section{Method}\label{sec:method}
\noindent\textbf{Task.}\label{sec:task} We are given one histological section $I$, the MRI volume $V$ of the same specimen, and, when available, the other sections of its tissue block, in cutting order and aligned to each other in one shared 2D frame.  We seek the section's rigid pose in the MRI, a $3{\times}3$ rotation matrix $R\in\mathrm{SO}(3)$ and a translation $t\in\mathbb{R}^3$. They place a point $u=(u_1,u_2)$ of the section at $x_{R,t}(u) = t + R\,(u_1,u_2,0)^{\top}$ in $V$'s frame, with $u$ and $t$ in mm.  When the block is given, we assume its sections are parallel, so they share one rotation $R$, and the section with index $i$ in the cutting order has the pose $(R,t_i)$. We also assume their depth along the plane normal is linear in $i$ (even spacing). 

 \noindent\textbf{Overview.} We learn two encoders with parameters $\theta=(\theta_H,\theta_M)$ and call them the dual
  encoder. We divide the section $I$ into overlapping square tiles $P_k$, indexed by $k$. $E_H$ encodes a tile, and $E_M$
  encodes $V$. The tile score $s_k$ of Equation~(\ref{eq:score}) compares the two encodings on the plane where a pose places
  the tile, and the section score $S$ averages it over the tiles. We train $\theta$ to rank a tile's reference pose above
  wrong poses the search mines (Section~\ref{sec:training}). Because no encoder exists before the first mining round, that round uses a pretrained image translator $F$, kept fixed, to synthesize an MRI view of each tile.  The inference steps add no learned
  parameters (Section~\ref{sec:inference}, Figure~\ref{fig:teaser}). The global search evaluates $S$ on a pose grid and
  keeps a candidate set $\mathcal{C}_i$ per section $i$. In block mode, the block vote selects $R_B$ from these sets, the
  re-solve gives each $t_i$ at $R_B$, the line fit corrects outlier $t_i$, and the refinement adds a rigid block correction.
  Single-section mode ends after the search.

  \subsection{Training: the dual encoder}\label{sec:training}
  \noindent\textbf{Encoders and score.} The tiles $P_k$ lie in one plane at known offsets (sizes in
  Section~\ref{sec:setup}). The encoder $E_H$ maps a tile to a feature-descriptor field $h_k = E_H(P_k)$, which assigns a unit vector
  $h_k(u)\in\mathbb{R}^8$, the feature descriptor, to every pixel $u$. $E_M$ maps $V$ to the feature-descriptor field $v =
  E_M(V)$, which assigns a unit-length feature descriptor to every voxel and is interpolated between voxels. Let $\Omega_k$
  be the foreground pixels of $P_k$. A pose's score for tile $k$ is the mean agreement of the two fields on its plane:
  \begin{equation}\label{eq:score}
  s_k(R,t) \,=\, \frac{1}{|\Omega_k|}\sum_{u\in\Omega_k}\big\langle h_k(u),\, v\big(x_{R,t}(u)\big)\big\rangle .
  \end{equation}
  The score is linear in $v$, and the entire MRI is encoded once. At a fixed $R$, one cross-correlation of $v$ with the
  rotated tile (one FFT per channel) scores every $t$. Because $E_H$ is trained on tiles, we score a section tile by tile.
  $S(R,t)$ is the mean of $s_k$ over the section's tiles, which all share its pose.

\noindent\textbf{Loss and negatives.} Let $s^{+}$ and $s_n$ be a training tile's scores, Equation~(\ref{eq:score}), at its reference pose and at wrong poses $n\in\mathcal{N}$. We minimize over $\theta$
\begin{equation}\label{eq:loss}
\mathcal{L} \,=\, -\log\frac{e^{s^{+}/\tau}}{e^{s^{+}/\tau}+\sum_{n\in\mathcal{N}}e^{s_n/\tau}} \,+\, \lambda\,(1-s^{+}) ,
\end{equation}
with temperature $\tau = 0.07$ and weight $\lambda = 1$. The contrastive (InfoNCE) term constrains only the differences between scores~\cite{yan2022scalecalib}, so we add a second term that pulls $s^{+}$ toward 1, as Wang and Isola's alignment term does~\cite{wang2020alignment}.   The set $\mathcal{N}$ contains the global search's best poses under the current model at least 15$^\circ$ or 15\,mm from the reference pose, as in hard example mining~\cite{shrivastava2016ohem} and ANCE~\cite{xiong2021ance}.  

\noindent\textbf{Two mining rounds.} We first train the translator $F$ on tiles paired with their reference MRI plane,
  because mining requires a cross-modal score before any dual encoder exists. Round 1 scores the MRI view image $F(P_k)$ of
  each tile against $V$ with masked normalized cross-correlation (NCC) and keeps 8 wrong poses per tile. We train a first
  dual encoder on these poses. Round 2 scores with this dual encoder and keeps 32 wrong poses per tile, and we train the
  final dual encoder on them. Each tile's $\mathcal{N}$ also contains the reference and mined poses of the other tiles of
  its batch and has up to 527 poses (Appendix~\ref{app:method}, Section~\ref{sec:serial}).

\subsection{Inference}\label{sec:inference}
\noindent\textbf{Global search.} The global search evaluates $S$ on a grid over $V$, 2048 rotations times $64^3$ translations, $5.4{\times}10^{8}$ poses per section, with the same score function the loss is defined on. A coarse-to-fine funnel applies non-maximum suppression, shortlists 256, improves them by local ascent (the finishing stage), and returns its top 4 as $\mathcal{C}_i$ (Appendix~\ref{app:method}).   

\noindent\textbf{Serial sections (block mode).} The block vote pools the rotations in all $\mathcal{C}_i$: the largest group of sections agreeing within a tolerance sets $R_B$, the estimate of $R$. If too few agree, each section keeps its single-section pose.  The re-solve searches each section over $t$ only, at $R_B$, giving $t_i$. The $t_i$ should be linear in $i$, so the line fit replaces an outlier by the line's value, given at least 8 inliers (Appendix~\ref{app:method}). 

\noindent\textbf{Refinement.} The refinement operates on the images and runs only for blocks with sufficient tissue (Appendix~\ref{app:data}).   From the poses $(R_B,t_i)$, it estimates one rigid block correction by local ascent on MIND-SSC~\cite{heinrich2012mind,heinrich2013mindssc} between evenly spaced sections and $V$ (Appendix~\ref{app:method}).

%% file: sections/03_setup.tex
\section{Experiments}\label{sec:experiments}\label{sec:nextbrain}

\subsection{Experimental setup}\label{sec:setup}

\noindent\textbf{Real data.} NextBrain~\cite{casamitjana2025nextbrain} is a public dataset of five postmortem human
  hemispheres, each with an ex vivo MRI and serial sections. The sections are stained with LFB, a myelin stain, and have
  matching H\&E images. NextBrain's 3D reconstruction, built with blockface photographs and a joint registration of all
  blocks to the MRI, is our reference standard, the role ground truth plays in other pose benchmarks. One affine map per
  tissue block defines the reference pose of every section and the pixel size of the block. The release has also warped each
  section onto its MRI plane, which removes the deformation from cutting except for that alignment's residual error
  (Appendix~\ref{app:adni}). The sections that block mode uses lie a median 0.71\,mm apart and span a median 10.5\,mm.
  Unless stated otherwise, the MRI is a 1\,mm isotropic volume synthesized from each hemisphere's ex vivo scan with
  SuperSynth~\cite{iglesias2025supersynth,iglesias2023synthsr,billot2023synthseg}, a network that maps every scan to one
  standard contrast and grid. NextBrain's data record also contains the native slides, scans of the glass slides that
  NextBrain did not warp, and Section~\ref{sec:mainresults} tests our method on them.

\noindent\textbf{Synthetic data.} Our pose is rigid, but cutting and mounting deform real sections. The released sections cannot show how accuracy falls as deformation grows, because they are already warped onto the MRI, and their reference poses have about 1\,mm of error.   Synthetic sections have neither problem: we set the deformation and know every pose exactly. They are not histology, so every claim about histology rests on the real data.    From in vivo MRI of 15 training and 4 held-out subjects of the Alzheimer's Disease Neuroimaging Initiative (ADNI)~\cite{jack2008adni}\footnote{Data used in the preparation of this article were obtained from the Alzheimer's Disease Neuroimaging Initiative (ADNI) database (adni.loni.usc.edu). The ADNI was launched in 2003 as a public-private partnership, led by Principal Investigator Michael W. Weiner, MD. The original goal of ADNI was to test whether serial magnetic resonance imaging (MRI), positron emission tomography (PET), other biological markers, and clinical and neuropsychological assessment can be combined to measure the progression of mild cognitive impairment (MCI) and early Alzheimer's disease (AD). The current goals include validating biomarkers for clinical trials, improving the generalizability of ADNI data by increasing diversity in the participant cohort, and to provide data concerning the diagnosis and progression of Alzheimer's disease to the scientific community. For up-to-date information, see adni.loni.usc.edu.}, we digitally cut NextBrain-like blocks of 8 sections, each block at a uniformly random 3D orientation, and warp them, elastic or free-form, by 0 to 3\,mm (Appendix~\ref{app:adni}).  Severity is the root-mean-square (RMS) displacement of a warp after removing the best rigid motion. FLAIR-like sections are registered into a T2-like volume (cross-contrast condition), or T2-like into T2-like (same-contrast control), all synthesized with SuperSynth. 

\noindent\textbf{Implementation details.} A section has pixels of about 0.1\,mm, and its square tiles (image patches) are 256 pixels across, about 25\,mm at the histology's pixel size.  $E_H$ takes each tile resampled to square pixels of about 0.2\,mm, and $E_M$ the MRI on an isotropic grid of about 1.4\,mm.  Histology has a black background: a pixel is foreground when its brightest channel exceeds 2\% of the intensity range.  Tiles with at least 15\% foreground are scored. 

\noindent\textbf{Evaluation.} On NextBrain, learned results are leave-one-hemisphere-out: each hemisphere is scored by models trained on the other four, with weights taken at a fixed final step (Appendix~\ref{app:loso}).  We call this the clean protocol, because no data of the held-out hemisphere reach training, negative mining, or checkpoint selection.   We chose inference settings on a development set of 150 sections (30 per hemisphere) in 68 blocks, solved with up to 16 sections each. The test set is 300 real NextBrain sections (60 per hemisphere) from 68 other blocks, scored under decision rules fixed in advance (pre-registered, Appendix~\ref{app:rules}).  Two choices were informed by the test set: the two-stain training of the dual encoder we report, chosen after we had seen single-section and block results on it, and the refinement, developed partly on its 15 hemisphere-5 blocks (Appendix~\ref{app:data}).
 A section is considered correct when its rotation error, the angle to the rigid pose that best fits NextBrain's alignment over the tissue, is below 5$^\circ$ and its translation error, measured at the tissue center, is below 2\,mm (joint accuracy). Differences are in points (0.01 is 1 point). Every learned result has two training seeds: the text reports their mean, taken from exact counts, and a pair gives seed~0~/~seed~1.   Confidence intervals (95\%, bootstrap) and $p$ values (paired sign-flip permutation test, significant when $p<0.05$) are over blocks, since a block's sections share their rotation, unless another test is named (Appendix~\ref{app:protocol}).

\noindent\textbf{Baselines.} Six baselines run on the same sections (Table~\ref{tab:baselines_untouched}, Appendix~\ref{app:baselines}). Two are training-free similarities inside our search at our settings, with the pose score swapped: masked NCC and MIND-SSC~\cite{heinrich2012mind,heinrich2013mindssc}.  Mutual information~\cite{viola1997alignment,maes1997multimodality,studholme1999overlap,pluim2003survey}, the classical similarity for multimodal registration, runs in our search but did not pass its pre-registered check within one modality on the development sections under either scoring (Appendix~\ref{app:baselines}), so Table~\ref{tab:baselines_untouched} has no row for it.  Three are learned: the feature matcher LoFTR-S2V~\cite{frolov2025needles}, our translator (Section~\ref{sec:training}) followed by masked NCC, and anatomix~\cite{anatomix2025}, a general-purpose 3D encoder pretrained on randomized synthetic volumes, frozen as our MRI encoder with a linear probe.   A fourth regresses the pose: EUReg~\cite{eureg2025}.  All but the translator also run within one modality, where the section is the MRI's own plane on the histology grid.

%% file: sections/05_nextbrain.tex
\subsection{Main results}\label{sec:mainresults}
  \noindent\textbf{How accurate is the method on the test set?} On the 300 LFB test sections (Table~\ref{tab:main}), block
  mode places 0.880~/~0.853 within 5$^\circ$ and 2\,mm (95\% confidence intervals [0.797, 0.948] / [0.764, 0.929]), and
  0.764~/~0.811 of the 297 test sections with a usable H\&E image ([0.653, 0.869] / [0.707, 0.908]). On native slides
  (Appendix~\ref{app:native}), it places 0.712~/~0.686 of 299 LFB and 0.557~/~0.647 of 289 H\&E sections.

\noindent\textbf{Does the method require the synthesized MRI?}  The synthesis reduces differences between scans and could inflate our results.  We retrain on the acquired 0.4\,mm MRI instead (Appendix~\ref{app:realmri}). With it, block mode gains 7 points on the LFB test sections and 1 on H\&E (Table~\ref{tab:main}). We treat the synthesized MRI as a design    
  choice. It gives every scan one contrast and grid (Section~\ref{sec:setup}), so a trained dual encoder could take scans   
  from other scanners without retraining, which we do not test.  

\noindent\textbf{Can one dual encoder serve both stains?} Sharing one dual encoder could reduce the accuracy on each stain.  The control is one dual encoder per stain, in block mode on the same test sections. An H\&E-only dual encoder stays 20 points below an LFB-only one (Table~\ref{tab:he} in the supplement).   Training on both stains costs LFB 5.2 points against an LFB-only dual encoder and raises H\&E by 0.3~/~13.1 points against an H\&E-only one (Appendix~\ref{app:results}).   Each change is significant on seed~1 only (exact block sign test).      One dual encoder therefore serves both stains. 

\noindent\textbf{How accurate is the method on native slides?} Brain banks store slides that no reconstruction has warped onto an MRI. A pre-registered test (Appendix~\ref{app:native}) replaces 299 LFB test sections with their native slides. We place each slide by one similarity transform fitted to its released image. This is the best case of a rigid placement. Against the released sections of the same runs, single-section accuracy falls by 19 points, from 0.714 to 0.527 (block-level $p<10^{-4}$).  In block mode, where the slides keep the release's alignment between them, the drop is 16 points, from 0.861 to 0.699 ($p<0.05$ on both seeds). The placement gives each slide an in-plane angle, a shift, a mirror state, and one scale factor (median 1.177, 0.986 to 1.489 over 282 accepted placements), so that the reference pose and the scoring stay those of the release. The released sections carry the same angle.   The search can represent the angle, the shift, and the mirror state, but this test shows every slide in the release's image frame, in which the encoders were trained.  The test does not separate the missing warp from our masking and rescaling.    

\begin{table*}[t]
\centering\scriptsize\setlength{\tabcolsep}{3.5pt}
\caption{\textbf{Our accuracy on NextBrain.} Seed 0 / seed 1, two-stain dual encoder. Errors are measured against the rigid pose that best fits NextBrain's alignment over the tissue, translation at the tissue center.  Bold: joint accuracy. Indented: acquired 0.4\,mm MRI. The LFB native row uses 299 native LFB test slides in their best rigid placement (Section~\ref{sec:mainresults}), and block mode keeps the release's alignment between slides. The H\&E native row uses 289 native H\&E test slides, each a separate slide next to its LFB slide, in their best rigid placement (Appendix~\ref{app:native}), and block mode keeps the release's alignment between slides.}
\label{tab:main}
\begin{tabular}{@{}lr ccccc ccccc@{}}
\toprule
 & & \multicolumn{5}{c}{Block mode} & \multicolumn{5}{c}{Single section}\\
\cmidrule(lr){3-7}\cmidrule(l){8-12}
Setting & $n$ & Joint$\uparrow$ & \shortstack{Rot.\\$<5^\circ$\,$\uparrow$} & \shortstack{Trans.\\$<2$\,mm\,$\uparrow$} & \shortstack{Med.\\rot.\ ($^\circ$)\,$\downarrow$} & \shortstack{Med.\\trans.\ (mm)\,$\downarrow$} & Joint$\uparrow$ & \shortstack{Rot.\\$<5^\circ$\,$\uparrow$} & \shortstack{Trans.\\$<2$\,mm\,$\uparrow$} & \shortstack{Med.\\rot.\ ($^\circ$)\,$\downarrow$} & \shortstack{Med.\\trans.\ (mm)\,$\downarrow$}\\
\midrule
LFB, test & 300 & \textbf{0.880/0.853} & 0.910/0.867 & 0.967/0.920 & 1.91/1.65 & 0.61/0.60 & \textbf{0.707/0.727} & 0.727/0.743 & 0.897/0.880 & 2.89/3.04 & 0.58/0.59\\
\quad acquired 0.4\,mm MRI & 300 & \textbf{0.953/0.920} & 0.980/0.923 & 0.973/0.960 & 1.63/1.61 & 0.57/0.56 & \textbf{0.780/0.780} & 0.783/0.783 & 0.917/0.897 & 2.44/2.26 & 0.52/0.53\\
\midrule
H\&E, test & 297 & \textbf{0.764/0.811} & 0.801/0.848 & 0.892/0.872 & 2.02/1.86 & 0.77/0.73 & \textbf{0.626/0.636} & 0.653/0.653 & 0.778/0.774 & 3.64/3.49 & 0.67/0.70\\
\quad acquired 0.4\,mm MRI & 297 & \textbf{0.774/0.818} & 0.808/0.862 & 0.862/0.875 & 2.20/2.15 & 0.76/0.72 & \textbf{0.734/0.694} & 0.747/0.707 & 0.828/0.801 & 2.95/2.99 & 0.64/0.63\\
\midrule
LFB, native slides & 299 & \textbf{0.712/0.686} & 0.756/0.732 & 0.890/0.876 & 3.41/2.96 & 0.87/0.86 & \textbf{0.512/0.542} & 0.522/0.572 & 0.826/0.819 & 4.62/4.32 & 0.87/0.91\\
H\&E, native slides & 289 & \textbf{0.557/0.647} & 0.619/0.716 & 0.737/0.806 & 3.52/3.48 & 1.21/1.04 & \textbf{0.436/0.495} & 0.467/0.522 & 0.723/0.761 & 5.21/4.83 & 1.17/1.01\\
\bottomrule
\end{tabular}
\end{table*}

\subsection{Comparison with baselines}\label{sec:baselines}

\noindent\textbf{Is a learned score needed?} The dual encoder must outperform training-free similarities to be justified.  Masked NCC and MIND-SSC replace $S$ in both modes. Masked NCC solves the task within one modality only, so our search is effective and the similarity is the limiting factor.  Our learned translator gives it no significant gain (0.237,   and 0.365 in block mode, below MIND-SSC). With the search alone, MIND-SSC reaches 0.377 cross-modally, against 0.717 for our single-section mode (block-level $p<10^{-4}$). Given the 256 poses that our learned score shortlists, exact MIND-SSC places 0.533: of its 34-point deficit, our finishing stage accounts for 18 points and the shortlist for 16 (Appendix~\ref{app:baselines}).   In block mode, MIND-SSC, which our refinement maximizes, reaches 0.620 before the refinement (Appendix~\ref{app:baselines}) and 0.667 after it (Table~\ref{tab:baselines_untouched}), and our block mode remains 20 points higher ($p<0.01$ on both seeds).    The learned score is therefore needed, but the blocks carry much of the accuracy, 29 points for MIND-SSC, close to the 34 that our descriptor adds in single-section mode.   

\noindent\textbf{What does the modality cost?} If the search were the limiting factor, the modality gap would be small.  We train and test on the MRI's own plane at each section's reference pose, with the LFB-only dual encoder's training tiles.   Our single-section mode then reaches 0.962 against 0.717 cross-modally with the two-stain dual encoder (Table~\ref{tab:baselines_untouched}). The modality costs single-section mode 25 points (block-level $p<10^{-4}$) and 20 points on the development sections.    The cross-modal score therefore most likely limits accuracy. 

\noindent\textbf{Do other learned methods perform better?}  Trained on NextBrain, existing methods might solve the task. LoFTR-S2V and EUReg also run on their own benchmarks. Direct pose regression, a ResNet-18 retrained on NextBrain tiles, did not fit its own training tiles, so Table~\ref{tab:baselines_untouched} has no row for it (Appendix~\ref{app:baselines}).  With a linear probe, anatomix reaches 0.700 within one modality and 0.357~/~0.210 cross-modally. LoFTR-S2V solves about one section in five within one modality (0.205) and almost none cross-modally (0.017).  On its own data and metric, our method reaches 87.9\% mean average accuracy against 51.5\% for its released model (two inference runs, Appendix~\ref{app:nhbench}). EUReg~\cite{eureg2025} places no section without an initial pose and reaches 0.058 within one modality given one.   On its own benchmark, with our search limited to its training poses, its mean distance error is 2.12\,mm and ours 0.30\,mm, but it is considerably faster.  It places 0.48 of the test frames within 5$^\circ$ and 2\,mm at the frame's center, ours all (Appendix~\ref{app:euregbench}).

\noindent\textbf{How large are the misses?}  Joint accuracy treats near and gross misses alike, and we call a miss gross when its error exceeds $30^\circ$ or 20\,mm.  Figure~\ref{fig:qualitative} shows six test sections chosen by a pre-registered rule. On all 300 test sections, block mode has no gross miss on seed~0 and 13 on seed~1, all in two blocks of one hemisphere. Its worst misses are at most $9.6^\circ$~/~$93.7^\circ$ and 3.6~/~139.2\,mm. The other methods have considerably more gross misses, on 21~/~24 of the 300 sections for our single-section mode, 128 for MIND-SSC, 224 for masked NCC, and 121~/~184 for anatomix.    The blocks and our score therefore both reduce gross misses. 

\begin{table}[t]
\centering\footnotesize\setlength{\tabcolsep}{2.3pt}
\caption{\textbf{Baselines on the 300 test sections.} Joint accuracy, clean protocol, seed~0~/~seed~1 (one value: deterministic). Errors are measured as in Table~\ref{tab:main}. $\dagger$: our pipeline with the pose score swapped. A dash: not applicable or not run. \emph{Initialized}: within $\pm$10$^\circ$ per axis and $\pm$4/$\pm$4/$\pm$8\,mm, as in EUReg's prostate benchmark.  The translator and the same-modality column are trained on LFB, the cross-modal columns' other learned rows on both stains. Bold: best per column.}
\label{tab:baselines_untouched}
\begin{tabular}{@{}lccc@{}}
\toprule
 & \multicolumn{2}{c}{Single section} & Block mode\\
\cmidrule(lr){2-3}\cmidrule(l){4-4}
Method & Same modality & Cross-modal & Cross-modal\\
\midrule
Masked NCC & 0.760\rlap{$^{\dagger}$} & 0.187\rlap{$^{\dagger}$} & 0.260\rlap{$^{\dagger}$}\\
MIND-SSC & 0.883\rlap{$^{\dagger}$} & 0.377\rlap{$^{\dagger}$} & 0.667\rlap{$^{\dagger}$}\\
Translator + masked NCC &--& 0.223/0.250\rlap{$^{\dagger}$} & 0.353/0.377\rlap{$^{\dagger}$}\\
anatomix + linear probe & 0.727/0.673 & 0.357/0.210 & --\\
LoFTR-S2V (fine-tuned) & 0.187/0.223 & 0.017/0.017 & --\\
EUReg & 0.000/0.000 & 0.000/0.000 & --\\
EUReg, initialized & 0.057/0.060 & 0.007/0.010 & --\\
\midrule
Ours & \textbf{0.963/0.960} & \textbf{0.707/0.727} & \textbf{0.880/0.853}\\
\bottomrule
\end{tabular}
\end{table}

\input{tables/fig_qualitative}

\subsection{Ablation study and analysis}\label{sec:serial}
\noindent\textbf{Do serial sections help?}    One section weakly constrains its own cutting plane (Section~\ref{sec:intro}).   With the same global search, block mode adds 15.0 points of joint accuracy over single-section mode on the development sections (block-level $p<10^{-4}$). On the test set it adds 15.0 (Table~\ref{tab:main}, and Table~\ref{tab:main_devfull} in the supplement).    

\noindent\textbf{Which design choices matter?} Each has a simpler or published alternative. We test each on the development sections.  In single-section mode, encoding the whole section at once loses 16 points of rotation accuracy on LFB and 16 on H\&E (Appendix~\ref{app:tiling}).  With $\lambda=0$, the dual encoder is as accurate (single-section joint accuracy 0.740~/~0.767 against 0.747~/~0.760, Appendix~\ref{app:method}).   With an LFB-only dual encoder the refinement adds 2.0 points of joint accuracy, not significant (block-level $p \geq 0.5$, Appendix~\ref{app:realmri}).   Replacing the 32 mined poses per tile with 32 random poses, drawn in the tissue, costs 14 points of single-section joint accuracy, and training on the first mining round alone costs 15. With 8 poses per tile, random poses match those mined in that round (0.600 against 0.603, Appendix~\ref{app:mining}).    That round mines with the translator and masked NCC, a weak score (Table~\ref{tab:baselines_untouched}).  Training with only the other tiles of the batch as negatives (CoMIR-style, about 15 per tile against 527) costs 15 points (exact block sign test, $p = 0.0013$~/~$0.0015$), and its gap to MIND-SSC is not resolved (Appendix~\ref{app:mining}).  A dual encoder trained so that its dot product imitates the LC2 similarity (DISA-style, 32 random poses per tile) costs 72 points (same test, $p<10^{-4}$~/~$<10^{-4}$) and falls below MIND-SSC (Appendix~\ref{app:mining}).   Tiling and mined negatives matter: mined poses lie nearer the reference rotation than random ones. 

\noindent\textbf{Can blocks compensate for missing content?} Most rotation-failed test blocks contain little tissue (Section~\ref{sec:analysis}).   On the development sections, before the refinement, we remove tiles, tissue, or sections and measure rotation accuracy, averaged over blocks (Figure~\ref{fig:content} in the supplement).   Four sections instead of sixteen cost block mode 5 points of rotation accuracy, and it still leads by 13 points.     As tiles drop to 6 with the tissue intact, the block's lead grows from 18 to 26 points.    Down to half the tissue, the lead stays between 17 and 24 points on both seeds, and at a quarter both modes fail.    The test compares the step at which each mode's accuracy halves.  On tiles and tissue, a constant lead alone would produce 15 to 74\% of the effect the pre-registered analysis interprets, under both scorings, as the block \emph{compensating} on both seeds.   Both remainders are significant on one seed only, so we consider them suggestive.     

\begin{figure*}[t]
\centering
\includegraphics[width=\textwidth]{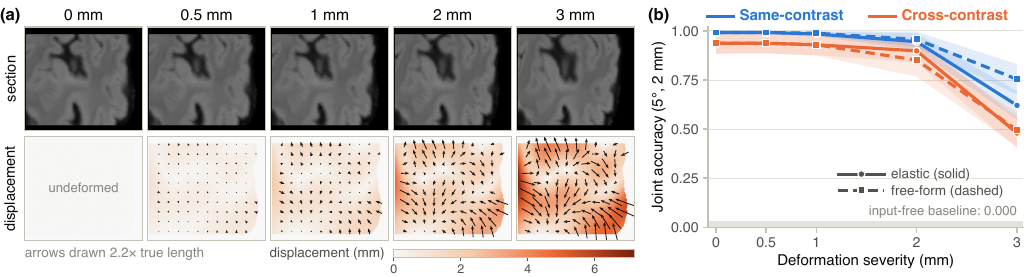}
\caption{\textbf{Deformation stress test.} \textbf{(a)}~One ADNI section at each severity (cross-contrast condition) and the field that warped it. \textbf{(b)}~Joint accuracy after the refinement against severity, seed 0, bands $\pm$1 standard error (Table~\ref{tab:adni} in the supplement). Input-free baseline: every block is assigned the most central rotation of the other subjects' blocks. Errors are measured against the exact pose.   }
\label{fig:adni}
\end{figure*}
\noindent\textbf{How many sections are needed?} Serial whole-brain histology is rare (Section~\ref{sec:intro}).   A pre-registered test solves each development section with $K{-}1$ evenly spaced other sections of its block, then refines it. At $K=4$, joint accuracy is 0.847, above its threshold of 0.70 on both seeds and about 6 points below the full block (Appendix~\ref{app:results}).  At $K=8$, the line fit (Section~\ref{sec:inference}), expected to be inactive at this size, replaced one of the 300 scored poses (150 sections, two seeds).  The conclusion holds with that pose counted as a miss.   Four evenly spaced sections therefore suffice.  

\noindent\textbf{Does a human cue help?} Brain banks may record a block's approximate position (Section~\ref{sec:intro}).  We simulate such a cue, which bounds the search, on the test sections.  A simulated cue of 10\,mm and 10$^\circ$, removes all gross misses, 21 / 24 of single sections and 13 of block mode on seed 1.   However, no cue size adds the pre-registered 5~points on both seeds (Appendix~\ref{app:cue}). In single-section mode most gross misses become near misses, so the absence of an initial pose is not the limiting factor.

\noindent\textbf{What does the search cost?} The search must be fast enough to run on every section. On one A100 GPU, the global search takes 25\,s per section and under 5\,GiB, against 72\,s for MIND-SSC and about 90\,s with direct convolution in place of the FFT (Appendix~\ref{app:runtime}).    The once-encoded MRI and the FFT make this possible (Section~\ref{sec:training}).

%% file: tables/fig_qualitative.tex
\begin{figure*}[t]
\centering
\includegraphics[width=\textwidth]{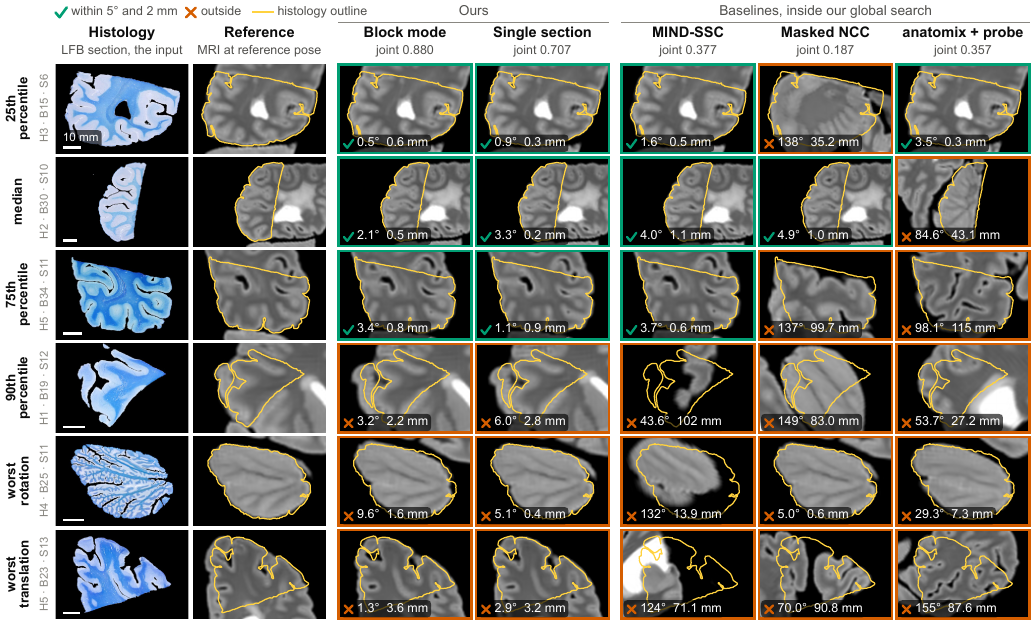}
\caption{\textbf{Registrations on the test set, failures included.} Rows: six of its 300 sections, chosen by the rank of their error in block mode under a pre-registered rule (Appendix~\ref{app:rules}), seed 0. Each cell: the 1\,mm MRI re-sliced on the plane that method returned, under the histology outline, with that section's rotation and translation error. \emph{Block mode} uses the block's serial sections, all other columns this section alone. All failed blocks and the second seed: Figures~\ref{fig:failures} and~\ref{fig:qualitative_s1} in the supplement. H, B, S: hemisphere, block, section. Scale bars: 10\,mm.}
\label{fig:qualitative}
\end{figure*}

%% file: sections/06_adni.tex
\subsection{Controlled experiments on synthetic data}\label{sec:adni}

\noindent\textbf{How much deformation does the method tolerate?} On NextBrain we cannot set the deformation.     We retrain the dual encoder on ADNI and run block mode.  Joint accuracy stays above 0.92 up to 1\,mm of deformation, drops at 2\,mm, and falls sharply at 3\,mm (Figure~\ref{fig:adni}).  Accuracy under elastic and free-form warps falls at the same severity.  The same-contrast control falls less at 3\,mm.   Histology need not follow this curve (Section~\ref{sec:mainresults}).

\noindent\textbf{Does accuracy hold on new subjects?} The development level is based on 4 held-out subjects, so a pre-registered rule tests 12 new ones (Appendix~\ref{app:rules}).   Accuracy is at or above the development level, or within two standard errors below it, for both conditions and seeds under either scoring. Cross-contrast accuracy is 0.988 undeformed and 0.909 at 2\,mm.   The level therefore holds on new subjects.  

\noindent\textbf{Do acquired scans match synthetic ones?} Every contrast above is synthesized.  We register acquired FLAIR and T1-weighted sections into the acquired T1-weighted volume (3~subjects, seed~0).   With the mining round chosen on 6 other subjects, rotation accuracy is within 4.2 points of the synthetic pair in all six settings.    With the synthetic pair's round, cross-contrast rotation accuracy is 25 to 67 points below it (Appendix~\ref{app:adni}).  The agreement depends on the round. 

%% file: sections/07_analysis.tex
\section{Discussion and conclusion}\label{sec:analysis}

\noindent\textbf{Remaining failures.}  Most failures are rotation errors. Among the 300 test sections, 36 / 44 fail (seed~0~/~seed~1): 27 / 40 on rotation, in 7 / 10 of the 68 blocks, and 9 / 4 on translation only (Appendix~\ref{app:failures}). Over half of the rotation-failed blocks (4 of 7 / 6 of 10) contain so little tissue that the refinement skips them.    In most rotation-failed blocks the rotation error is mainly out-of-plane: tilt exceeds the in-plane rotation in 5 of 7~/~10 of 10.

%% file: sections/08_limitations.tex
\noindent\textbf{Limitations.}\label{sec:limitations} 
Our evaluation uses one real dataset of five hemispheres, reflecting the scarcity of such data given the time, cost, and difficulty of acquisition. NextBrain provides a reference reconstruction through a highly tailored registration pipeline with multiple intermediate steps, yet even this reference remains approximate: its residual landmark error of 0.99 $\pm$ 0.51\,mm~\cite{casamitjana2025nextbrain} precludes verifying translation accuracy below about 1\,mm, while our 2\,mm criterion exceeds that error. This reliance on the reconstruction also extends to our native-slide evaluation (LFB and H\&E), where accuracy is lower and slides retain the released placement and pixel size, remaining within the encoders' training image frame (Section~\ref{sec:mainresults}). Generalization beyond these conditions therefore remains to be established: inter-section tilt and uneven spacing are untested, and training and testing cover only LFB and H\&E. In such settings, the method offers no confidence estimate to flag unreliable poses, and voting can correct only errors that differ between sections.

\noindent\textbf{Conclusion.}\label{sec:conclusion}
  We show that a stained section warped onto its MRI plane can be placed in that MRI with no initial pose, landmarks, or    
  manual step. Native slides are placed less accurately (Section~\ref{sec:mainresults}). The method also carries over to two
  other 2D/3D benchmarks within one modality (Appendices~\ref{app:nhbench} and~\ref{app:euregbench}). With it, a brain bank moves closer to placing every LFB and H\&E section in 3D, to validate MRI  
  markers of disease against histology and to map pathology across its donors.

%% file: sections/08b_llm_usage.tex
\section*{Acknowledgments}
We used Claude (Anthropic), through the Claude Code tool, throughout the research and the writing, including the design and running of experiments, the code, the figures, the related work, and the text.  The authors directed and verified all of this work and take full responsibility for the paper.

%% file: sections/10_appendix.tex
\noindent\textbf{How to use this supplement.} Appendices~\ref{app:data} to~\ref{app:loso} describe the data, the test set, and the protocol behind every number, and Appendix~\ref{app:method} gives the main settings.  Appendix~\ref{app:realmri} covers the acquired MRI. Appendix~\ref{app:results} expands on the main results, the second stain, and several ablations of Section~\ref{sec:serial}, and Appendix~\ref{app:settings} explains how the inference settings were chosen. Appendices~\ref{app:baselines} to~\ref{app:euregbench} detail the baselines of Table~\ref{tab:baselines_untouched}, Appendix~\ref{app:negative} lists the ideas we discarded, and Appendix~\ref{app:failures} shows every failed block of seed~0.  Appendix~\ref{app:runtime} gives the runtime, Appendix~\ref{app:adni} details the controlled experiments, Appendix~\ref{app:cue} describes the human cue, and Appendix~\ref{app:rules} gives pre-registered rules with their outcomes.   Appendix~\ref{app:native} reports the pre-registered test of Section~\ref{sec:mainresults} on native LFB slides and its repeat on native H\&E slides.   Appendix~\ref{app:release} covers the data use, ethics, and code release. 

\input{sections/supp_setup}

\begin{table*}[t]
\centering\footnotesize\setlength{\tabcolsep}{4pt}
\caption{Clean-protocol runs on the 150 development sections with an LFB-only dual encoder, seed 0 / seed 1. They belong to the pre-registered analysis of Appendix~\ref{app:realmri}, not to the two-stain rows of Table~\ref{tab:main_devfull}. Joint accuracy before and after the refinement, and per held-out hemisphere after it (H1 to H5, 30 sections each). The overall accuracy is the mean of the five. Errors are measured as in Table~\ref{tab:main}.}
\label{tab:clean}
\begin{tabular}{@{}lccccccc@{}}
\toprule
Volume & \shortstack{Before\\refinement} & \shortstack{After\\refinement} & H1 & H2 & H3 & H4 & H5\\
\midrule
\input{tables/tab_clean_rows}\end{tabular}
\end{table*}

\section{Leave-one-hemisphere-out protocol}\label{app:loso}
The accuracies of this appendix and of Table~\ref{tab:clean} are measured as in Section~\ref{sec:setup} (Appendix~\ref{app:protocol}).  

\noindent\textbf{The clean protocol.} A fold is one held-out hemisphere together with the models trained on the other four. For each fold, every step runs with no access to its held-out hemisphere: its own translator (Appendix~\ref{app:method}), its own two rounds of negative mining, and its own dual encoders. Each network is trained on the other four hemispheres with no validation hemisphere and kept at the weights of its final step. A guard inside each training and mining run refuses any read of the held-out hemisphere, and it recorded no such read in any fold. 

\noindent\textbf{What the paper reports.} Table~\ref{tab:main} reports the clean protocol on the test set, and Table~\ref{tab:main_devfull} on the development set and the full pool. The development and full-pool rows come from the same runs. The test rows come from a second pass in which each fold's own clean dual encoders, with no retraining, score the section list of the test set's first run. The native-slide test of Appendix~\ref{app:native}, the LFB native row of Table~\ref{tab:main}, is a further pass of each fold's clean two-stain dual encoders, with no retraining, over 299 of the LFB test sections. The H\&E native row of Table~\ref{tab:main} comes from the same kind of pass over native H\&E slides of the test set (Appendix~\ref{app:native}).    The baseline table (Table~\ref{tab:baselines_untouched}) uses the clean protocol in every column. The anatomix baseline is trained per fold exactly as our dual encoders are, separately for each modality (the fold's own training negatives, the weights of the final step, the same guard). Our same-modality model is trained the same way, with every training tile at the same pose as its cross-modal twin. It reaches 0.953~/~0.960 in single-section mode on the development set.  No other baseline trains on, or selects a checkpoint on, the held-out hemisphere. Masked NCC and MIND-SSC are training-free, and the translator row uses translators trained under the clean protocol. The direct pose regressor (Appendix~\ref{app:baselines}), LoFTR-S2V, and EUReg train on the other four hemispheres and never choose a checkpoint on the held-out one.  Every ablation of Section~\ref{sec:serial} uses the clean protocol. Figure~\ref{fig:teaser} comes from earlier LFB-only dual encoders. Row (a) replays one training step of such a dual encoder. In row (b), the refinement panel plots the 300 test sections of seed 0, and the other panels show one block of the development set.    

\input{sections/supp_method}

\section{The acquired 0.4\,mm MRI instead of the synthesized MRI}\label{app:realmri}
\noindent\textbf{Summary.} In joint accuracy, with the two-stain dual encoder (Tables~\ref{tab:main} and~\ref{tab:main_devfull}), the acquired MRI matches or exceeds the synthesized MRI on every released LFB row and on the released H\&E rows of the test set. On the H\&E test sections, it reaches a joint accuracy of 0.774 / 0.818 against 0.764 / 0.811 in block mode. Its joint accuracy is up to 4 points lower on some H\&E development and full-pool rows.     With an LFB-only dual encoder, in the pre-registered analysis below, the acquired MRI does not outperform the synthesized MRI. These two results are why Section~\ref{sec:mainresults} treats the synthesized MRI as a design choice.  These runs do not determine whether the difference of that analysis, 3.3 / 2.0 points, is real. The difference is not significant at block level. It is based on 7 / 4 lost sections in 3 / 2 blocks, mostly in hemisphere 5, and the seeds differ on whether it is already present before the refinement.  

\noindent\textbf{Design (pre-registered).} Only the volume changes: the acquired 0.4\,mm ex vivo MRI of the same hemispheres replaces the MRI synthesized with SuperSynth. It is resampled onto the same 1\,mm grid with its intensities median-matched, so only the contrast differs and every reference pose and millimeter conversion is unchanged. Every model is retrained on it under the clean protocol. Hemisphere 5 was scored first as a go/no-go check: the acquired MRI was required to reach at least half the accuracy on the synthesized MRI. The check gives a pass, at 0.767 / 0.800 against 0.867 / 0.900.    

\noindent\textbf{Result.} With an LFB-only dual encoder (the pre-registered analysis), block mode on the acquired MRI reaches a joint accuracy of 0.920 / 0.940 against 0.953 / 0.960 on the synthesized MRI.   The difference is $-3.3$ / $-2.0$ points (block-level $p = 0.63$ / $0.75$, and exact McNemar $p = 0.18$ / $0.38$ on the 150 paired sections). The pre-registered rule calls the two volumes \emph{equivalent} when the difference lies inside the equivalence band (at most 5 points in either direction) on both seeds. Applied to these accuracies, the rule gives \emph{equivalent}. The acquired MRI is lower on both seeds, and neither difference is significant at block level.  Accuracy within $5^\circ$ goes from 0.960 / 0.973 to 0.927 / 0.947, and within 2\,mm from 0.973 / 0.967 to 0.927 / 0.967. The lost sections fail mostly on rotation. Of the 7 / 4 sections the acquired MRI loses, 7 / 3 fail on rotation. The remaining one misses the 2\,mm criterion by a quarter of a millimeter (2.25\,mm).   The lost sections lie in 3 / 2 blocks, all in hemisphere 5 except one block of hemisphere 1 (Table~\ref{tab:clean}), and one block loses sections on both seeds. In single-section mode, the volumes differ by up to 2.7 points (from 0.733 / 0.753 to 0.760 / 0.753). On the 300 LFB test sections, with the two-stain dual encoder of Table~\ref{tab:main}, the acquired MRI is higher on both seeds in both modes. It reaches 0.953 / 0.920 against 0.880 / 0.853 in block mode and 0.780 / 0.780 against 0.707 / 0.727 in single-section mode.  No rule was pre-registered for that comparison, and it is reported without a test.

\noindent\textbf{Refinement.} It contributes little on either volume in this LFB-only analysis: $+2.0$ / $+2.0$ points on the synthesized MRI and $+1.3$ / $0.0$ on the acquired MRI. It fixes 1 to 3 sections and breaks at most 1, and no change is significant (block-level $p \geq 0.5$). It acts on 56 of the 68 blocks, and its size and tissue gate skips the rest. 

\input{sections/supp_results}
\input{sections/supp_adni}

\input{sections/supp_cue}
\input{sections/supp_rules}
\input{sections/supp_native}

%% file: sections/supp_setup.tex
\section{Data and evaluation sets}\label{app:data}
\noindent\textbf{Real data.} NextBrain provides five postmortem human hemispheres with serial Luxol fast blue (LFB) and matching hematoxylin and eosin (H\&E) histology at about 0.1\,mm per pixel (Section~\ref{sec:setup}). A quality check keeps a section only if the MRI re-sliced at its reference pose matches the MRI slice the release provides for it, with a Pearson correlation of at least 0.7. It removes 535 of the 5033 sections (10.6\%) and 12 of the 213 blocks entirely, and leaves 4498 sections in 201 tissue blocks. Every NextBrain accuracy in this paper is therefore measured only on sections whose reference pose passes this check.  Two tests also use native slides, the scans of the LFB and H\&E glass slides that NextBrain did not warp. NextBrain's data record contains these scans and, for many LFB slides, label maps in slide space (Appendix~\ref{app:native}).  

\noindent\textbf{Evaluation sets.} The 150 development sections, 30 per hemisphere, are selected by a fixed stride over each hemisphere's blocks, and block mode still solves every section of the full pool (Table~\ref{tab:main_devfull}).  The test set is 300 sections from 68 other blocks. It is selected by the same stride rule over each hemisphere's remaining blocks, 60 sections per hemisphere in 14, 13, 14, 12, and 15 blocks.  Its first run, after the method was frozen as it then stood, evaluated an earlier LFB-only dual encoder once, under a rule and a block selection, both pre-registered. 

\noindent\textbf{Two choices made partly on the test set.} The refinement's gate skips blocks whose sections are small or contain less than 10\% tissue. This gate and the number of sections the refinement uses were chosen in pilot runs. These runs initialized the refinement from perturbed reference poses of 29 blocks of hemisphere 5 outside the development set. These include all 15 test blocks of that hemisphere (60 of the 300 sections). Outside the development set, the settings were chosen on seven blocks, and two of them, blocks 31 and 34, are test blocks.   The two-stain dual encoder that the paper reports was adopted afterward under a second rule, written down after its expected gain and cost had been measured on the test set, in single-section and block mode. The test set is therefore not fully held out for that dual encoder (Section~\ref{sec:setup}).     The native-slide test of Appendix~\ref{app:native} scored 299 of the LFB test sections once more and chose no setting of the method. While the test was designed, read-only probes fitted between 26 and 100 test sections to their native slides, and we measured the appearance of 10 test sections. No accuracy of the method on native slides had been seen at that time.      The H\&E native-slide test in the last paragraph of Appendix~\ref{app:native} scored 289 of the H\&E test sections once more and chose no setting of the method, and its plan and amendments list its other contacts with the test set. These 289 are the 297 test sections with a usable H\&E image (Table~\ref{tab:main}) without the eight sections of two blocks that have no native H\&E scan in NextBrain's data record. The excluded sections were fixed by name before any native H\&E slide of the test set was read. 

\noindent\textbf{Synthetic data.} Appendix~\ref{app:adni} gives the setup of the ADNI experiments, the full deformation ladder, the residual alignment error of real sections, the confirmation on new subjects, and the acquired pair.  

\section{Protocol details}\label{app:protocol}
\noindent\textbf{Held-out hemispheres.} On NextBrain, all learned results are leave-one-hemisphere-out: the model for each hemisphere is trained on the other four (Appendix~\ref{app:loso}).

\noindent\textbf{Scoring.} A pose is scored against NextBrain's alignment of its block, an affine map from the section to the MRI. The rotation error is the angle to the rigid pose that best fits this map over the section's tissue. The translation error is the distance between where the estimate and the map place the center of the tissue. A section is correct when the rotation error is below 5$^\circ$ and the translation error is below 2\,mm (joint accuracy, Section~\ref{sec:setup}). This measure holds for every NextBrain result scored on sections, in the paper and in this supplement. On the synthetic sections the reference is the exact pose (Appendix~\ref{app:adni}). The benchmark of Appendix~\ref{app:euregbench} measures translation at the center of the ultrasound frame, as its own protocol does.  The reference pose that training uses is read from the block's affine map, in the plane of the pose that errors are measured against. Training reads it at each tile's center. Where Appendix~\ref{app:baselines} reads a score or a nearest rotation at this pose, it is read at the center of one of the section's tiles. For the runs initialized at or near it (Appendices~\ref{app:data}, \ref{app:baselines}, and~\ref{app:failures}), it is read at the center of the section image. For the cue that Appendix~\ref{app:cue} draws around it, the pose is read at a tile's center and moved in that plane to the center of the section's tiles.   

\noindent\textbf{Earlier scoring.} An earlier version of this measure scored a pose where our search stores it, at the center of the section's anchor tile, its first tile, row by row, with at least 15\% foreground. The translation error was the distance between where the estimate and the map put that point, and the reference rotation was read from the map along the image's horizontal axis, not fitted over the tissue. That point lies off the tissue in 251 of 300 test sections, a median 25.4\,mm from the tissue center. A rotation alone of just under 5$^\circ$ about that center therefore moves it by up to 2.2\,mm in the median section, past the 2\,mm threshold. The measure above follows NextBrain's alignment more faithfully. Its reference pose lies a median 0.17\,mm from where the alignment puts the tissue, against 0.40\,mm for the earlier one, and is independent of where a method stores a pose. We adopted it after measuring its effect on the test set and native LFB slides, so the test set informed this choice too. Both versions use the same thresholds and, on synthetic sections, differ only in where translation is measured. The pre-registered rules were written for the earlier version, and we apply each rule that scores sections to numbers under the measure above. Where the main text says ``under either scoring'' or ``under both scorings'', its numbers use the measure above, and the earlier version gives the same conclusion. 

\noindent\textbf{Seeds, intervals, and tests.}
Every learned result has two training seeds. For our method, the two seeds of each held-out hemisphere are two training runs of the final dual encoder. They share one translator and, for each stain, both mining rounds and the first dual encoder.  The main text gives their mean, taken from exact counts, or the pair, and the tables and the supplement give seed 0 / seed 1, never the better seed. Confidence intervals and permutation tests resample blocks rather than sections, because the sections of a block share their rotation. Any other test is named where it is used. An exact sign test over the blocks appears in Appendices~\ref{app:results}, \ref{app:baselines}, and~\ref{app:rules}, and an exact McNemar test over paired sections in Appendix~\ref{app:realmri}. An unpaired $z$ test, for two sets that share no block, appears in Appendices~\ref{app:baselines} and~\ref{app:rules}. Appendices~\ref{app:nhbench} and~\ref{app:euregbench} use an exact sign test over the volumes or subjects of their benchmark, and on ADNI each standard error comes from a bootstrap over subjects, then blocks (Appendix~\ref{app:adni}).

%% file: tables/tab_clean_rows.tex
Synthesized (SuperSynth) & 0.933/0.940 & 0.953/0.960 & 1.00/1.00 & 0.93/0.93 & 1.00/1.00 & 0.97/0.97 & 0.87/0.90\\
Acquired 0.4\,mm MRI & 0.907/0.940 & 0.920/0.940 & 0.93/1.00 & 0.93/0.93 & 1.00/1.00 & 0.97/0.97 & 0.77/0.80\\
\bottomrule

%% file: sections/supp_method.tex
\section{Method details and hyperparameters}\label{app:method}
This appendix gives the stage sizes and settings of Section~\ref{sec:method}. Every setting is the one used for Table~\ref{tab:main} unless stated otherwise.

\noindent\textbf{Encoders.} The two CNN encoders output 8-channel descriptors of unit length, with 65,648 parameters in total, 47,784 in $E_H$ and 17,864 in $E_M$.   A tile is 256 source pixels across, about 25\,mm. Its width in mm follows the pixel size of its block, read from the block's reference affine map, and stays within about 3\% of 25\,mm. Each tile is resampled to $128{\times}128$ square pixels of about 0.2\,mm before the 2D encoder, and the MRI is encoded on an isotropic $128^3$ grid of about 1.4\,mm. 

\noindent\textbf{Loss and negatives.} Equation~(\ref{eq:loss}) uses $\tau = 0.07$ and $\lambda = 1$. Its second term pulls the score $s^{+}$ of the reference pose toward 1 (Section~\ref{sec:training}).  With $\lambda = 0$, the dual encoder is as accurate: its joint accuracy is 0.740 / 0.767 against 0.747 / 0.760 with the term, in single-section mode on the 150 development sections.    The wrong poses lie at least 15$^\circ$ or 15\,mm from the reference pose, and they are mined in two rounds, 8 per tile in round 1 and 32 in round 2. Round 1 scores the translator's MRI view of each tile with masked NCC, and round 2 uses the first dual encoder. The translator is trained first, for each fold on its four training hemispheres only, and then kept fixed. Each tile is scored at its own reference and mined poses and at those of the other tiles of its batch. This gives up to 144 poses with the round-1 negatives and up to 528 with the round-2 negatives. One of them is its own reference pose, so its set $\mathcal{N}$ has up to 527 poses in round 2. Poses within 15$^\circ$ and 15\,mm of a tile's own reference pose are left out of its negatives.   Training is leave-one-hemisphere-out with two seeds, and each fold keeps the weights of a fixed final step, with no validation hemisphere (Section~\ref{sec:setup}, Appendix~\ref{app:loso}).

\noindent\textbf{Training.} The two encoders of a dual encoder are trained together with Adam at a learning rate of $3{\times}10^{-4}$, for 4000 steps with batches of 16 tiles. The translator is trained with AdamW at a learning rate of $2{\times}10^{-3}$ and a weight decay of $10^{-4}$, also for 4000 steps with batches of 16 tiles. The translator is a U-Net with three levels and 32 channels at the first level. It maps a 128-pixel RGB tile to a one-channel MRI view. Its loss is an L1 loss plus one minus the Pearson correlation over the MRI's tissue pixels, without an adversarial term.  All networks are implemented in PyTorch. 

\noindent\textbf{Tiles.} Training tiles are drawn under the same 15\% foreground gate as the scored tiles. In single-section mode, encoding the whole section at once loses 12 / 20 points of rotation accuracy on LFB and 11 / 21 on H\&E (Appendix~\ref{app:tiling}).  

\noindent\textbf{Global search and funnel.} The global search scores 2048 rotations, about 14$^\circ$ apart, at all $64^3$ translations, about 3\,mm apart, which gives $5.4{\times}10^{8}$ poses per section. The coarse-to-fine funnel then keeps 22,528 poses, then the 768 basins that non-maximum suppression leaves, and then 256 shortlisted poses. The finishing stage keeps 48 of them after the first of its two rounds of local ascent and the top 4 after the second.  

\noindent\textbf{The score in detail.} The MRI descriptor sampled at a point is used as interpolated, without renormalization. A pixel that falls outside the volume contributes zero, so a pose that leaves the volume lowers its score instead of raising it. The global FFT stage scores the unweighted mean of Equation~(\ref{eq:score}) over a section's tiles. The later stages of the funnel re-rank the remaining candidates with the same tile score but exclude each candidate's lowest-scoring tile. In single-section mode, they also weight every tile by its foreground fraction.   

\noindent\textbf{Block vote, line fit, and refinement.} The block vote uses the top-4 rotations of every section, a tolerance of 15$^\circ$, and at least 3 agreeing sections. The line fit acts on a block only if at least 8 of its sections, including any it replaces, lie within 2.5\,mm of a line through two of them. For each section, it finds the line through two of the others that has the most of them within 2.5\,mm and, if at least 7, fits a least-squares line to these. It replaces a section more than 2.5\,mm from the fit's value by that value.     The refinement adjusts one shared rotation and offset for up to 8 evenly spaced sections by maximizing MIND-SSC with radius 2 and dilation 2 (Appendix~\ref{app:baselines}). It skips a block whose sections are small or contain less than 10\% tissue (Appendix~\ref{app:data}).  With the LFB-only dual encoder of Table~\ref{tab:clean}, it is close to neutral (Appendix~\ref{app:realmri}). 

\noindent\textbf{Where the settings come from.} Appendix~\ref{app:settings} says how the settings were chosen.   With an earlier LFB-only dual encoder, a foreground gate of 10\% or 20\% in place of 15\% changed rotation accuracy in single-section mode on the 30 development sections of one hemisphere by a net of at most one section. Thirty sections can detect only a large effect, so neither gate replaced the 15\% one.

%% file: sections/supp_results.tex
\section{Additional results and ablations}\label{app:results}

Every accuracy in this appendix is measured as in Section~\ref{sec:setup} (Appendix~\ref{app:protocol}).  
\input{tables/tab_content}
\begin{figure*}[t]
\centering
\includegraphics[width=0.8\textwidth]{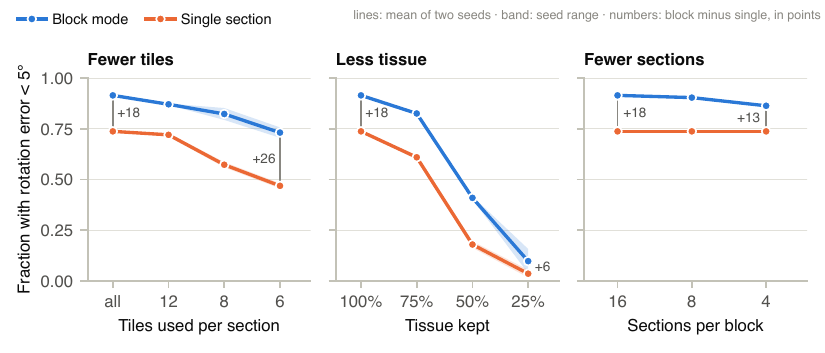}
\caption{\textbf{Missing content.} Rotation accuracy within 5$^\circ$ before the refinement, mean over blocks (Table~\ref{tab:content}).}
\label{fig:content}
\end{figure*}

Table~\ref{tab:content} gives the numbers behind Figure~\ref{fig:content}.   The pre-registered rule for these runs (Appendix~\ref{app:rules}) compares, for each block, the step at which each mode's rotation accuracy halves, interpolated linearly between steps.  Applied to these accuracies, the rule gives \emph{compensating} on both seeds for tiles and for tissue. That is, the accuracy of block mode halves at a later step than that of the Single columns of Table~\ref{tab:content}, where each section is solved alone. A constant lead alone would produce 15 to 74\% of that effect, and the remainder is significant on one seed only, for tiles and for tissue.  In the rows of Table~\ref{tab:content} with fewer sections per block, the scored sections of a block are solved together in one smaller block. The pre-registered test of how many sections suffice instead solves each scored section in its own block of $K$ sections, itself and $K{-}1$ evenly spaced unscored sections of its block, and then refines it. At $K=4$, that test reaches 0.853 / 0.840 joint accuracy, above its threshold of 0.70 on both seeds and about 6 points below the full block, which reaches 0.920 / 0.887.    Applied to these accuracies, the rule of that test gives the conclusion that four evenly spaced sections suffice. At $K=8$, the line fit (Section~\ref{sec:inference}), expected to be inactive at this size, replaced one of the 300 scored poses (150 sections, two seeds), and the conclusion holds with that pose counted as a miss. 

\noindent\textbf{Intervals.} Block-bootstrap 95\% intervals for the released LFB block cells of Tables~\ref{tab:main} and~\ref{tab:main_devfull}, seed 0 / seed 1: development 0.920 [0.842, 0.981] / 0.887 [0.801, 0.960], and 0.927 [0.850, 0.987] / 0.927 [0.849, 0.987] on the acquired 0.4\,mm MRI.  The full pool gives 0.933 [0.877, 0.977] / 0.904 [0.838, 0.961], and 0.935 [0.874, 0.984] / 0.934 [0.874, 0.983] on the acquired MRI. The test set gives 0.880 [0.797, 0.948] / 0.853 [0.764, 0.929], and 0.953 [0.911, 0.986] / 0.920 [0.841, 0.980] on the acquired MRI. For the released H\&E block cells of Table~\ref{tab:main}, the test set gives 0.764 [0.653, 0.869] / 0.811 [0.707, 0.908], and 0.774 [0.663, 0.880] / 0.818 [0.724, 0.905] on the acquired MRI.   

\begin{table*}[t]
\centering\scriptsize\setlength{\tabcolsep}{3.5pt}
\caption{\textbf{Accuracy on the development set and the full pool.} Seed 0 / seed 1, clean protocol, two-stain dual encoder, as in Table~\ref{tab:main}, which gives the test set. Errors are measured against the rigid pose that best fits NextBrain's alignment over the tissue, translation at the tissue center.  The full pool contains up to 16 sections of each of the 68 development blocks, 1050 of the 1480 LFB sections they contain. Indented: acquired 0.4\,mm MRI.}
\label{tab:main_devfull}
\begin{tabular}{@{}lr ccccc ccccc@{}}
\toprule
 & & \multicolumn{5}{c}{Block mode} & \multicolumn{5}{c}{Single section}\\
\cmidrule(lr){3-7}\cmidrule(l){8-12}
Setting & $n$ & Joint$\uparrow$ & \shortstack{Rot.\\$<5^\circ$\,$\uparrow$} & \shortstack{Trans.\\$<2$\,mm\,$\uparrow$} & \shortstack{Med.\\rot.\ ($^\circ$)\,$\downarrow$} & \shortstack{Med.\\trans.\ (mm)\,$\downarrow$} & Joint$\uparrow$ & \shortstack{Rot.\\$<5^\circ$\,$\uparrow$} & \shortstack{Trans.\\$<2$\,mm\,$\uparrow$} & \shortstack{Med.\\rot.\ ($^\circ$)\,$\downarrow$} & \shortstack{Med.\\trans.\ (mm)\,$\downarrow$}\\
\midrule
LFB, development & 150 & 0.920/0.887 & 0.920/0.900 & 0.960/0.913 & 1.62/1.55 & 0.60/0.57 & 0.747/0.760 & 0.753/0.760 & 0.860/0.860 & 2.56/2.51 & 0.60/0.56\\
\quad acquired 0.4\,mm MRI & 150 & 0.927/0.927 & 0.927/0.927 & 0.973/0.960 & 1.31/1.48 & 0.55/0.54 & 0.813/0.780 & 0.820/0.787 & 0.887/0.880 & 2.20/2.18 & 0.51/0.53\\
LFB, full pool & 1050 & 0.933/0.904 & 0.943/0.921 & 0.965/0.939 & 1.63/1.56 & 0.59/0.59 & 0.796/0.783 & 0.803/0.786 & 0.908/0.896 & 2.50/2.49 & 0.54/0.54\\
\quad acquired 0.4\,mm MRI & 1050 & 0.935/0.934 & 0.937/0.937 & 0.987/0.969 & 1.35/1.45 & 0.54/0.53 & 0.839/0.823 & 0.843/0.826 & 0.930/0.924 & 2.27/2.18 & 0.48/0.49\\
\midrule
H\&E, development & 147 & 0.810/0.789 & 0.844/0.878 & 0.878/0.810 & 1.79/1.78 & 0.73/0.73 & 0.619/0.680 & 0.633/0.694 & 0.769/0.782 & 3.44/3.41 & 0.63/0.65\\
\quad acquired 0.4\,mm MRI & 147 & 0.782/0.789 & 0.810/0.803 & 0.816/0.823 & 1.66/1.64 & 0.68/0.67 & 0.653/0.646 & 0.673/0.667 & 0.782/0.769 & 2.99/3.14 & 0.67/0.62\\
H\&E, full pool & 1044 & 0.793/0.797 & 0.841/0.880 & 0.868/0.831 & 1.79/1.78 & 0.76/0.75 & 0.660/0.664 & 0.670/0.678 & 0.778/0.799 & 3.39/3.39 & 0.65/0.66\\
\quad acquired 0.4\,mm MRI & 1044 & 0.773/0.759 & 0.800/0.788 & 0.814/0.803 & 1.75/1.85 & 0.67/0.69 & 0.674/0.666 & 0.687/0.679 & 0.798/0.770 & 2.93/3.01 & 0.61/0.61\\
\bottomrule
\end{tabular}
\end{table*}

\subsection{Second stain}

\noindent\textbf{One dual encoder per stain.} In Table~\ref{tab:he}, the H\&E-only dual encoder is trained with only the stain changed (each fold reuses its own LFB translator and mines both rounds of training negatives on H\&E). The LFB-only dual encoder is trained with the same recipe on LFB alone. The H\&E-only dual encoder stays 15 to 32 points below the LFB-only one on the same sections, on every set, both MRI volumes, and both seeds, with block-level $p < 0.011$ in every row.  

\noindent\textbf{The trade-off of training on both stains, paired per section.} Against the LFB-only and H\&E-only dual encoders, the two-stain dual encoder of Table~\ref{tab:main} loses some accuracy on LFB and gains some on H\&E.  On the 300 LFB sections of the test set, it costs 3.0 / 7.3 points in block mode (0.910 / 0.927 to 0.880 / 0.853, exact block sign test $p = 0.55$ / $0.016$). In single-section mode, it costs nothing ($+1.0$ / $0.0$ points, $p = 0.82$ / $1.0$). On the development sections, the block cost is 3.3 / 7.3 points ($p = 0.69$ / $0.063$). On the 297 H\&E sections of the test set, it gains 0.3 / 13.1 points in block mode (0.761 / 0.680 to 0.764 / 0.811, $p = 0.45$ / $0.039$). With the acquired 0.4\,mm MRI, no LFB cost remains on the test set ($+1.3$ / $+1.3$ points), and on the development sections the change is $+0.7$ / $-1.3$ points, none significant. The rule for this choice (Appendix~\ref{app:rules}) was written down after the expected cost and gain had been measured on the test set (Appendix~\ref{app:data}). It reads joint accuracy in block mode on the test set with the synthesized MRI.  It adopts the two-stain dual encoder with a stated cost if its LFB accuracy falls 2 to 8 points below that of the LFB-only dual encoder on either seed. Its LFB accuracy must not fall more than 8 points below on the other seed, and its H\&E accuracy must stay at 0.50 or above on both seeds. Applied to the block-mode accuracies of the synthesized MRI on the test set above, the rule gives adoption with a stated cost.

\begin{table*}[t]
\centering\footnotesize\setlength{\tabcolsep}{4pt}
\caption{Second stain under the clean protocol with one dual encoder per stain (the control for the two-stain dual encoder of Table~\ref{tab:main}): joint accuracy of the H\&E-only and of the LFB-only dual encoder on the same sections, seed 0 / seed 1. Errors are measured as in Table~\ref{tab:main}. In every row H\&E is below LFB with block-level $p < 0.011$. The sets are smaller than the LFB sets of Tables~\ref{tab:main} and~\ref{tab:main_devfull}. Exclusions fixed before the H\&E runs remove H\&E images that are blank, flat gray, or copies of the LFB section. The H\&E runs also leave out three marginal sections that LFB solves on three or four tiles. The single-section rows are evaluated on the sections both stains select, because removing sections moves the single-section selection onto three other sections in the development set and in the test set.}
\label{tab:he}
\begin{tabular}{@{}llccccc@{}}
\toprule
& & & \multicolumn{2}{c}{Synthesized MRI} & \multicolumn{2}{c}{Acquired 0.4\,mm MRI}\\
\cmidrule(lr){4-5}\cmidrule(l){6-7}
Mode & Set & $n$ & H\&E & LFB & H\&E & LFB\\
\midrule
Block mode & development & 147 & 0.755/0.762 & 0.952/0.959 & 0.694/0.633 & 0.918/0.939\\
 & full pool & 1044 & 0.737/0.766 & 0.943/0.953 & 0.691/0.625 & 0.918/0.943\\
 & test set & 297 & 0.761/0.680 & 0.909/0.929 & 0.764/0.704 & 0.939/0.909\\
\midrule
Single section & development & 147 & 0.517/0.558 & 0.741/0.762 & 0.531/0.524 & 0.769/0.762\\
 & test set & 297 & 0.545/0.552 & 0.704/0.734 & 0.606/0.556 & 0.774/0.747\\
\bottomrule
\end{tabular}
\end{table*}

\noindent\textbf{Qualitative results.} Figures~\ref{fig:qualitative_he} and~\ref{fig:qualitative_he_s1} show H\&E sections in the same way as Figures~\ref{fig:qualitative} and~\ref{fig:qualitative_s1} show LFB sections. The same pre-registered rule chooses six of the 297 H\&E sections of the test set, and every pose is read from the runs behind the released H\&E rows of Table~\ref{tab:main}. Block mode has gross misses, beyond $30^\circ$ or 20\,mm, on 12 / 22 of the 297 sections, and single-section mode on 39 / 39 (seed~0 / seed~1). 

\input{tables/fig_qualitative_he}
\input{tables/fig_qualitative_he_s1}

\subsection{Tiles versus the whole section}\label{app:tiling}

\noindent\textbf{The comparison.} This ablation tests whether a section must be encoded tile by tile, as the encoders are trained and a section is scored (Section~\ref{sec:training}). The alternative encodes the whole section once, as one image padded to a square, and each tile takes its descriptors from the resulting descriptor field at its position. The tiles, their weights in the score, the search, and every setting remain unchanged, and nothing is retrained, since both encoders are fully convolutional. Table~\ref{tab:tiling} uses the two-stain dual encoder of Table~\ref{tab:main}. 

\noindent\textbf{Result.} Encoding the whole section at once loses 12.0 / 20.0 points of rotation accuracy on LFB and 10.9 / 21.1 on H\&E (seed 0 / seed 1). Accuracy falls on both seeds and both stains, with block-level $p<10^{-4}$ for each stain. The tile rows equal the single-section cells of Table~\ref{tab:main_devfull}.  

\noindent\textbf{The encoder's input.} The normalization layers of the 2D encoder pool their statistics over its whole input. Its training tiles are at least 15\% foreground, and the tiles scored here are about half foreground. Padded to a square, whole sections have a median of 14\% foreground, so the encoder normalizes over considerably more background than it does with tiles. 

\begin{table}[t]
\centering\footnotesize\setlength{\tabcolsep}{3pt}
\caption{Tiles versus the whole section in single-section mode on the development sections of Table~\ref{tab:main_devfull}, seed 0 / seed 1. Change: rotation accuracy of the whole section minus that of the tiles, in points. $p$: block-level, both seeds together. Bold: the better encoding of each stain.}
\label{tab:tiling}
\input{sections/tab_tiling}\end{table}

\subsection{Mined against random negatives}\label{app:mining}
\begin{table}[t]
\centering\footnotesize\setlength{\tabcolsep}{3pt}
\caption{Mined against random wrong poses, in single-section mode on the 150 development sections, seed 0 / seed 1. Mined: by the search with the first dual encoder (round 2, ours) or with the translator (round 1). Random: a uniform rotation and a center in the tissue. Changes are in points of joint accuracy, with the exact sign test over blocks. Bold: best.}
\label{tab:mining}
\begin{tabular}{@{}lcc@{}}
\toprule
\input{tables/tab_mining_rows}\end{tabular}
\end{table}

\noindent\textbf{The test.} Our dual encoders train on wrong poses that the global search mines (Section~\ref{sec:training}). To test whether the mining affects accuracy, we retrain only the two-stain dual encoder of every fold, on both seeds. The first retraining run replaces the 32 mined poses of each tile with 32 random poses. Each has a uniform rotation and a center drawn uniformly in the tissue of the tile's hemisphere. Like a mined pose, it lies at least 15$^\circ$ or 15\,mm from the reference pose. The draw is made once, so both seeds use the same poses. The second retraining run keeps the first 8 of them, and the third uses the 8 poses per tile that the translator mines in the first mining round. Every tile, batch, and other setting is unchanged, and every dual encoder, ours included, is trained from scratch. Each retrained dual encoder is evaluated in single-section mode on the 150 development sections, paired per section with the dual encoder of Table~\ref{tab:main_devfull}, under a rule fixed before any run (Appendix~\ref{app:rules}). The rule compares joint accuracy per seed with the exact sign test over blocks. The mined poses count as \emph{better} if they gain at least 5 points on both seeds with $p<0.05$ on at least one seed, and the random poses count as \emph{better} only if they gain at least 5 points on both seeds with $p<0.05$ on both. A change below 5 points on both seeds counts as \emph{no change}, and anything else as \emph{mixed}. The same rule compares the second mining round with the first, the second round taking the place of the mined poses.  

\begin{table}[t]
\centering\footnotesize\setlength{\tabcolsep}{2pt}%
\caption{Dual encoders trained without mined poses, in single-section mode on the 150 development sections, seed 0 / seed 1. In-batch negatives only: the negatives of a tile are the other tiles of its batch at their reference poses, about 15 per tile against 527 for ours. Similarity imitation: the dot product is trained to imitate LC2 at the reference pose and the 32 random poses of Table~\ref{tab:mining}. Changes are in points of joint accuracy, with the exact sign test over blocks.}
\label{tab:nomining}
\begin{tabular}{@{}lcc@{}}
\toprule
\input{tables/tab_nomining_rows}\end{tabular}
\end{table}

\noindent\textbf{Result.} Random poses cost 14.7 / 14.0 points of joint accuracy, with block-level $p<0.05$ on both seeds (exact sign test), so the mined poses count as \emph{better}, the outcome the pre-registered plan expected (Table~\ref{tab:mining}).  Training on the first round alone costs 16.0 / 14.0 points, with block-level $p<0.05$ on both seeds, and the second round also counts as \emph{better}. The score that mines the poses and the number of poses change together there, so this measures the second round as a whole. With 8 poses per tile, random poses perform as well as those the translator mines in the first round ($-1.3$ / $+2.0$ points for the mined ones, \emph{no change} under the same rule). Outside the rule, 32 random poses per tile instead of 8 change joint accuracy by $0.0$ / $+2.0$ points. Across the five folds, the median tissue fraction of the MRI plane under a tile is 0.91 to 0.92 for random poses and 0.90 to 0.91 for mined ones. However, the median rotation from the reference pose is 132$^\circ$ against 46 to 48$^\circ$. Both medians are measured on the training tiles, and the rotation is the angle between a wrong pose and the reference pose that training uses for its tile (Appendix~\ref{app:protocol}).  

\noindent\textbf{Training without mined poses.} Two more retraining runs, pre-registered after the test above, remove the mined poses altogether (Table~\ref{tab:nomining}). The first keeps our loss but trains on no wrong pose. The only negatives of a tile are the other tiles of its batch at their own reference poses, as in CoMIR~\cite{pielawski2020comir}. This yields about 15 negatives per tile, against 527 for ours and for 32 random poses. Its comparisons with those two cannot separate the wrong poses from the larger number of negatives. The second keeps the training tiles, batches, and 32 random poses of the test above. Instead of our loss, it trains its dot product to imitate the LC2 similarity, as DISA does~\cite{ronchetti2023disa} (Appendix~\ref{app:baselines}). LC2 is the fraction of the local variance of the histology that a linear fit of the MRI's intensity and gradient magnitude explains. The imitation dual encoder excludes one tile in 20 (about 5\%) from its loss, to test its fit on tiles it never trained on. It trains on about 95\% of the tiles the others use.  The random poses are one draw, shared by both seeds and by both dual encoders that use them, and the in-batch dual encoder uses none. The two seeds therefore measure training noise only, not the variation between draws. 

\noindent\textbf{Fit to their own objectives.} Before any accuracy was computed, each of the 10 dual encoders of each retraining run was required to fit its own training objective, with thresholds fixed in advance. All 20 passed. The parentheses give each threshold, and some also give the chance level.  The in-batch dual encoders were measured on a fixed set of their fold's training tiles. They reach in-batch rank-1 0.988 to 0.997 (at least 0.50, chance about 0.07), so almost every tile scores its own reference pose highest among those of its batch. Their tile-mean descriptors still differ across tiles. The spread, one minus the median cosine between the tile-mean descriptors of two tiles, is 0.32 to 1.39 times that of our dual encoder of the same fold and seed (at least 0.25 times). With every histology tile set to zero, they drop to zero-input rank-1 0.07 to 0.19 (at most 0.20). Our dual encoders of the same folds and seeds, scored the same way as a control, reach 0.993 to 0.997 in-batch and 0.06 to 0.16 with zero input. Before training, exact LC2 ranked the reference pose first among the 33 poses of a tile on 0.745 to 0.785 of 200 training tiles of each fold (at least 0.25, chance 0.03). On the held-out tiles, which the imitation dual encoders never trained on, their dot product and LC2 have a weighted Pearson correlation of 0.62 to 0.65 (at least 0.50). There they rank the reference pose first among the 33 poses with rank-1 0.39 to 0.52 (at least 0.25, chance 0.03), against 0.70 to 0.75 for exact LC2.  These are tile-level numbers, and wherever a pose enters them, it is the reference pose that training uses for each tile (Appendix~\ref{app:protocol}). 

\noindent\textbf{Rules.} Six rules, three per retrained dual encoder, were fixed before any run (Appendix~\ref{app:rules}) and are applied with no family-wise claim. Each compares joint accuracy per seed, paired per section, with the exact sign test over blocks. When a retrained dual encoder is compared with ours or with 32 random poses, ours or the random poses count as \emph{better} if they gain at least 5 points on both seeds with $p<0.05$ on at least one seed. The retrained one counts as \emph{better} only with $p<0.05$ on both. The asymmetry is deliberate, as in the test above. Each retraining run removes a part of our recipe, and a reason to remove a part must meet a stricter criterion than a reason to keep it. A change below 5 points on both seeds counts as \emph{no change}, and anything else as \emph{mixed}. Against MIND-SSC, a retrained dual encoder counts as \emph{above} or \emph{below} only with at least 5 points and $p<0.05$ on both seeds, and as not resolved otherwise. Against ours, the plan called \emph{no change} or a better retrained dual encoder informative. Against MIND-SSC, it called \emph{below} informative for the in-batch dual encoder and \emph{above} for imitation. Against 32 random poses, it called a better in-batch dual encoder informative, and for imitation \emph{no change} or a better imitation dual encoder. The plan also assigned each outcome a probability before the run, printed below as its prior. 

\noindent\textbf{In-batch negatives only.} Of the sections, this dual encoder places 0.587~/~0.620 (Table~\ref{tab:nomining}). Ours, with 527 negatives per tile against about 15, gains 16.0~/~14.0 points over it ($p = 0.0013$~/~$0.0015$) and counts as \emph{better}, the likely outcome (prior 0.80). The in-batch dual encoder gains 12.0~/~15.3 points over MIND-SSC, at 0.467 ($p = 0.13$~/~$0.035$, 95\% block-bootstrap intervals [$+1.9$, $+22.4$]~/~[$+5.3$, $+25.3$] points). It is therefore not resolved against MIND-SSC, the likeliest outcome (prior 0.55).  The 32 random poses, also with 527 negatives per tile, change joint accuracy by $+1.3$~/~$0.0$ points against it ($p = 0.83$~/~$1.0$, intervals [$-6.0$, $+8.8$]~/~[$-8.1$, $+8.5$] points), within 5 points on both seeds, \emph{no change} (prior 0.30).  The in-batch dual encoder's accuracies equal those of the first mining round alone in Table~\ref{tab:mining}, but the two do not place the same sections. 

\noindent\textbf{Similarity imitation.} This dual encoder fits LC2 on tiles it never trained on, but places only 0.027~/~0.033 of the sections in single-section mode. Ours gains 72.0~/~72.7 points over it ($p<10^{-4}$~/~$<10^{-4}$) and counts as \emph{better}, the likely outcome (prior 0.90). Trained without mined poses, the imitation dual encoder falls 44.0~/~43.3 points below MIND-SSC ($p<10^{-4}$~/~$<10^{-4}$). It counts as \emph{below}, the direction the plan expected, though not its likeliest outcome (prior 0.35).   With the same training tiles, batches, and random poses, our loss gains 57.3~/~58.7 points over imitation ($p<10^{-4}$~/~$<10^{-4}$) and counts as \emph{better}, the likely outcome (prior 0.70). 

\section{How the inference settings were chosen}\label{app:settings}

We chose the inference settings on the development sections (Section~\ref{sec:setup}), except the refinement's gate and the number of sections the refinement uses. These two were chosen in pilot runs on hemisphere-5 blocks outside the development set, among them all 15 hemisphere-5 blocks of the test set (Appendix~\ref{app:data}). The line fit's two thresholds are the code's defaults. Appendix~\ref{app:method} lists the main settings.

\section{Baselines in detail}\label{app:baselines}

Every accuracy on sections in this appendix is measured as in Section~\ref{sec:setup} (Appendix~\ref{app:protocol}).

Within one modality, each tile-based method receives the MRI plane on the histology grid inside the histology foreground, so its tiles are identical in both single-section columns of Table~\ref{tab:baselines_untouched}. Direct pose regression after DeepSlice~\cite{carey2023deepslice} and SVRnet~\cite{hou2018svr} predicts each tile's pose as anchor points under a geodesic loss~\cite{salehi2019svr} and takes the median of the section poses its tiles imply. The reference tile poses combined this way would place every test section correctly.  A ResNet-18~\cite{he2016resnet} trains for 4000 steps of batch 16 (Adam at $3\times10^{-4}$, cosine decay to zero, gradient clip 1.0), 8.9 passes over the training tiles within one modality and 4.4 over the two-stain tiles cross-modally. The checkpoint is chosen on a tenth of the training hemispheres' tiles. The share of its own training tiles that it places within $5^\circ$ and 2\,mm of their training pose is 0.000 to 0.006 within one modality and 0.000 to 0.002 cross-modally. Its mean distance to that pose is 8.2 to 13.2$^\circ$ and 14.0 to 31.6\,mm. On the test sections it scores 0.000 in both columns, within $5^\circ$ and 2\,mm and within $10^\circ$ and 5\,mm. A regressor that does not fit its training data says nothing about the task, so Table~\ref{tab:baselines_untouched} has no row for it.  On the training tiles, each prediction is compared with the tile's own reference pose (Appendix~\ref{app:protocol}), the pose the regressor is trained to predict.  

\noindent\textbf{Settings of the training-free scores.} Masked NCC and MIND-SSC receive the tiles our method receives (Section~\ref{sec:setup}), each converted to one gray image, the mean of its three color channels.  They run in our global search and, in block mode, through our block vote, re-solve, line fit, and refinement. The search uses our settings (2048 rotations, $64^3$ translations, the same funnel and local ascent), none tuned for them.  Masked NCC, the Pearson correlation over the tile's foreground pixels, runs on the gray tile and again with its tissue intensities inverted, and the higher final score is kept. In block mode, a section keeps the polarity whose best pose scores higher and reuses it in its re-solve.  The MIND-SSC of a rendered plane is not a slice of a MIND-SSC volume, so the exact score cannot be encoded once and sliced. It is computed on two shortlists of 256 poses each from our search. One is made with masked NCC on the gray tile, the other with a MIND-SSC volume averaged over offset directions (four radii from 1.4 to 5.6\,mm). Our method uses one shortlist, the 256 poses of its learned score. At each pose, the MRI plane is rendered and compared with the tile, both at 96 pixels, by the MIND-SSC of our refinement (radius 2 and dilation 2). The score is minus the mean absolute difference of the two MIND-SSC images over pixels that are foreground in the tile and tissue in the render. The best 256 are then improved by our local ascent.  MIND-SSC is built from squared intensity differences, so inverting all intensities leaves it unchanged, and it runs on the gray tile alone.  The only setting of these two scores chosen from data is the channel normalization of that MIND-SSC volume. It was selected among four by the rank of each tile's reference pose among random poses, on a hemisphere other than the one being solved.  

\noindent\textbf{Score or search.} Of the 187 test sections that MIND-SSC misses, 33 have a higher exact score at the reference pose than at the returned pose. For them, the search therefore missed a pose that the score prefers. Here the reference pose is the one that training uses (Appendix~\ref{app:protocol}), and it meets the joint criterion for all 33 of them. None of them has a shortlisted pose within 5$^\circ$ of the reference pose. For the 113 sections that MIND-SSC places, the nearest shortlisted pose lies within 19.9$^\circ$, and for 17 of the 33 it lies farther. The other 154 have a higher score at the returned pose, but so do 112 of the 113 placed sections, where the local ascent ends at a score up to 0.084 above that of the reference pose. For 14 of the 154 the margin is larger than 0.084, or for 13 if we leave out one section whose score at the reference pose is penalized for a tile with insufficient overlap. However, 0.084 is the largest value seen on 113 sections, not a bound, so this comparison cannot determine for any of the 154 whether the score or the search is the limiting factor. 

\noindent\textbf{MIND-SSC on our shortlist.} The two analyses below were pre-registered and run after the main results (Appendix~\ref{app:rules}). The comparison above uses two poses per section. To hold the candidates fixed instead, we gave the exact MIND-SSC stage the 256 poses that our learned score shortlists for each test section, in place of its own 512 poses, and ran our finishing stage on the same poses in the same task (Table~\ref{tab:mindlsl}). MIND-SSC's deficit then splits exactly into the share $d$ of our finishing stage, which covers the score and the local ascent around it (at 128 pixels for ours and 96 for MIND-SSC), and the share $g$ of our shortlist. The exact score cannot drive a search, so this tests MIND-SSC as the final ranker of our candidates. Run again in this task, our finishing stage reproduced the outcome of Table~\ref{tab:baselines_untouched} on all 300 test sections for both seeds, and the same pose, within 0.01$^\circ$ and 0.01\,mm, on 300 / 299. Both shares are significant on both seeds (exact block sign test, $p<10^{-3}$). Within 20$^\circ$ and 20\,mm, $d$ is still 13.3 / 10.3 points, so MIND-SSC prefers another of our candidates. MIND-SSC keeps 142 / 145 of the 212 / 218 sections that our finishing stage places, and places 17 / 16 of the 88 / 82 that our finishing stage misses. For these misses, the nearest of our 256 rotations lies a median 8.6$^\circ$ / 8.6$^\circ$ from the reference rotation that training uses, and within 5$^\circ$ on 8 / 14. Where our method places a section and MIND-SSC on its own poses misses it, the nearest rotation among those 512 poses lies a median 11.8$^\circ$ / 10.9$^\circ$ from that rotation, against 5.9$^\circ$ / 6.1$^\circ$ where both place it. The order $J_{\mathrm{R}} > J_{\mathrm{ML}} > J_{\mathrm{MB}}$ holds in every hemisphere on both seeds except hemisphere 2 on seed 0, where $J_{\mathrm{R}} = J_{\mathrm{ML}}$. The whole task takes a median of 77 / 74\,s per section on an A100 GPU. 

\begin{table*}[t]
\centering\footnotesize
\caption{Exact MIND-SSC on our shortlist, on the 300 test sections, per seed of our dual encoder. $J_{\mathrm{MB}}$: MIND-SSC on its own 512 poses, its cell of Table~\ref{tab:baselines_untouched}. $J_{\mathrm{ML}}$: on the 256 poses of our learned shortlist. $J_{\mathrm{R}}$: our finishing stage on the same poses, in the same task. $k$: sections whose outcome equals that of Table~\ref{tab:baselines_untouched}. $d = J_{\mathrm{R}} - J_{\mathrm{ML}}$ and $g = J_{\mathrm{ML}} - J_{\mathrm{MB}}$ in points of joint accuracy, with block-bootstrap 95\% intervals. Lower rows: $d$ and $g$ at looser thresholds of the rotation and translation errors. The subscript is the seed.}
\label{tab:mindlsl}
\begin{tabular}{@{}lcccccc@{}}
\toprule
Seed & $J_{\mathrm{MB}}$ & $J_{\mathrm{ML}}$ & $J_{\mathrm{R}}$ & $k$ & $d$ [95\% CI] & $g$ [95\% CI] \\
\midrule
0 & 0.377 & 0.530 & 0.707 & 300 & +17.7 [+10.1, +25.6] & +15.3 [+9.4, +21.7] \\
1 & 0.377 & 0.537 & 0.727 & 300 & +19.0 [+10.9, +27.4] & +16.0 [+10.5, +21.8] \\
\midrule
Threshold & $d_0$ & $g_0$ & $d_1$ & $g_1$ & & \\
5$^\circ$, 2 mm & +17.7 & +15.3 & +19.0 & +16.0 & & \\
10$^\circ$, 5 mm & +15.3 & +21.3 & +15.7 & +21.3 & & \\
15$^\circ$, 10 mm & +15.3 & +22.7 & +13.3 & +23.7 & & \\
20$^\circ$, 20 mm & +13.3 & +22.7 & +10.3 & +24.3 & & \\
\bottomrule
\end{tabular}
\end{table*}

\noindent\textbf{Mutual information.} Mutual information~\cite{viola1997alignment,maes1997multimodality,studholme1999overlap,pluim2003survey}, the classical similarity for multimodal registration, can be computed by FFT at every translation~\cite{ofverstedt2022fastmi}. We ran it in our global search as the pose score, computed from 8 gray levels on each side. The levels of a tile are fit to the gray tile that masked NCC and MIND-SSC score, so that only the similarity differs from theirs. The FFT method~\cite{ofverstedt2022fastmi} fits its levels to the color image, which would give mutual information more input than the other training-free scores receive, and uses mini-batch k-means, which is stochastic. We use a deterministic 1D k-means, initialized at quantiles of the distinct values and stopped after at most 100 iterations. The volume fit reached that cap for hemisphere 4 at 8 levels and for all five hemispheres at 16, and no tile fit reached it. Each level set is blurred and interpolated trilinearly as a separate image, so the counts are those of partial-volume interpolation~\cite{maes1997multimodality}, exact at every voxel translation. Voxels outside the MRI volume count as air, the level whose center is nearest 0, so every pose's joint histogram has the same total weight. Relabeling the levels leaves mutual information unchanged, so unlike masked NCC it requires no second run with inverted intensities. The search takes a median of 108\,s per section at 8 levels and 360\,s at 16 on an A100 GPU. With masked NCC in its place, the same code reproduced the archived records of 2 test sections per fold on each of two A100 GPU models. Before the test sections, mutual information was required to place at least 15 of 30 development sections within one modality, with one retry at 16 levels. These are the first 6 of each hemisphere, where masked NCC and MIND-SSC each place 27. It placed 3 at 8 levels and 3 at 16 (0 and 0 cross-modally), so it was not run on the test sections. Within one modality it scores the reference pose above the returned one on all 27 of its misses at both level counts, so on these sections the search missed a pose that mutual information prefers. Cross-modally it does so on 15 of its 30 misses at 8 levels and 14 of 30 at 16. Here the reference pose is the one that training uses (Appendix~\ref{app:protocol}), and it meets the joint criterion on all 30 development sections. 

Before that, mutual information had failed a check on a phantom, in which tiles are cut from a synthetic volume at a random pose and their intensities remapped by $4x(1-x)$, which masked NCC cannot model. Our search with mutual information returned a pose 132.4$^\circ$ from the planted one at 8 levels and 132.9$^\circ$ at 16, although mutual information scores the planted pose higher (0.550 against 0.253 at 8 levels, 0.640 against 0.313 at 16). At 8 levels, the planted pose is eliminated at the first stage, where tiles and volume are blurred and resized. It scores 0.0082 there, below the best pose of every one of the 2048 rotations (median 0.0204), and the best pose at the planted rotation lies 80\,mm from the planted translation with 37\% of its tile weight outside the tissue. No pose within 16$^\circ$ and 6\,mm of the planted one survives that stage, while the next stage would rank the planted pose first. Masked NCC on the tiles before the remapping keeps the planted pose through every stage and returns it to within 0.03$^\circ$. Before any number on real sections existed, we attributed this to mutual information in our search, since every exactness check had passed, and let the development sections decide.  

\noindent\textbf{Translator + masked NCC.} This row is the masked-NCC row with each tile's gray image replaced by the MRI view that our translator synthesizes from it, as in our first mining round (Section~\ref{sec:training}). The clean protocol trains one translator per fold, on LFB alone, and uses it for both stains. On seed~0 it is the translator that mined the first round of wrong poses for both seeds of our dual encoder, and on seed~1 a second translator, trained the same way with another seed. Each is paired with our dual encoder of the same seed. The row has no same-modality cell, because within one modality the section already is an MRI plane. Against masked NCC, its joint accuracy changes by $+3.7$ / $+6.3$ points, which is not significant (exact block sign test, $p = 0.41$ / $0.44$). It stays 15.3 / 12.7 points below MIND-SSC (same test, $p = 0.0026$ / $0.0015$), and our single-section mode leads it by 48.3 / 47.7 points ($p<10^{-12}$).   After reading the single-section result, we also ran the row in block mode under a pre-registered rule. There it reaches 0.353 / 0.377, below MIND-SSC's 0.667 (exact block sign test, $p = 0.0013$ / $0.00054$). Against masked NCC's 0.260 in block mode, its joint accuracy changes by $+9.3$ / $+11.7$ points, which is not significant ($p = 0.093$ / $0.052$).  

The anatomix row~\cite{anatomix2025} uses the frozen 5.9M-parameter anatomix encoder as the MRI encoder, with a 136-parameter linear probe and our 2D encoder trained with our recipe, separately for each modality. Its heads (the probe and the 2D encoder of each modality) are trained under the clean protocol exactly as our dual encoders are (Appendix~\ref{app:loso}), the cross-modal one on both stains. At section level the row reaches 0.727 / 0.673 within one modality and 0.357 / 0.210 cross-modally, below our single-section mode in both columns (block-level $p<10^{-4}$). The cross-modal cell is comparable to MIND-SSC on seed 0 and below it on seed 1.
  A cross-modal head trained on LFB alone reaches 0.290 / 0.193. Against it, training on both stains, at twice the training data, brings no change that is significant on both seeds ($+6.7$ / $+1.7$ points, exact block sign test $p = 0.043$ / 0.70, a gain on seed 0 alone).  Only LFB-only heads were run on the development sections, where they reach 0.833 / 0.787 within one modality and 0.353 / 0.253 cross-modally.   Used training-free, anatomix scores 0.000 on the same-modality problem of the development sections.   External learned matchers (MatchAnything~\cite{he2025matchanything}, MINIMA~\cite{ren2025minima}) did not converge as a local refinement in our matching loop, in same-modality controls where the histology is replaced by its MRI plane. We ran them on up to 40 blocks of hemisphere 5, among them test blocks, each initialized 5$^\circ$ and 2\,mm from the reference pose that training uses (Appendix~\ref{app:protocol}). In these controls only one block ended within 2$^\circ$ and 1\,mm of that pose.  

LoFTR-S2V~\cite{frolov2025needles} takes each whole section as one 128-pixel image through its own adapter. Fine-tuned on our folds with both stains, for as many training steps as its fine-tuning on LFB alone, it reaches 0.017 / 0.017 on the 300 test sections (Table~\ref{tab:baselines_untouched}). Fine-tuned on LFB alone, it reaches 0.013 / 0.007, so training on both stains, at twice the training data, brings no significant change (exact block sign test $p = 1$ / 0.45). Fine-tuned instead on the same-modality problem, with its own recipe unchanged, it reaches 0.187 / 0.223, 17~/~22 points above its LFB-only cross-modal result (block-level $p<10^{-4}$). This is below every other same-modality cell of Table~\ref{tab:baselines_untouched} except EUReg's.   On the development sections, where only the LFB-only models were run, the one fine-tuned cross-modally on LFB scores 0.000 on MRI sections.  The same-modality result falls in neither band of its pre-registered rule on either set ($\ge 0.30$ on both seeds, or $<0.05$ on both). The row is therefore reported as numbers alone. 

\noindent\textbf{EUReg.} EUReg~\cite{eureg2025} registers a 2D ultrasound frame to a 3D volume. It assumes that the frame starts at the central plane of the volume and regresses the rigid offset from there. We used its released model and loss unchanged, in the variant of its paper for a volume larger than the frame. We trained the model in our own loop, with the optimizer, learning rate decay, and batch size of its released script. Training followed the clean protocol for each fold, on the data of the LoFTR-S2V rows, with both stains for the cross-modal column. We kept the weights of the final step. The volume is the 128-voxel cube of those data, at 1.42 or 1.48\,mm per voxel against 0.8\,mm in EUReg's prostate data, and each section is one image. Each model was trained for as many updates as fit in 4 hours on an A100 GPU, a number fixed before any accuracy was read. The budget is 110,514 updates for the row without initialization and 381,254 for the initialized row, against 600,000 in EUReg's paper. Cross-modally this is 89 to 95 passes over the training sections for the row without initialization and 309 to 330 for the initialized row. It is 178 to 190 and 615 to 657 within one modality, where training uses one stain. Without initialization, every section starts at the central sagittal plane of the volume, and training uses each section at its reference pose only. With initialization, each section starts at a pose drawn uniformly within $\pm$10$^\circ$ about each axis, $\pm$4\,mm along both axes of the section, and $\pm$8\,mm along its normal from the reference pose that EUReg is trained to recover. This is the range of EUReg's prostate benchmark (Appendix~\ref{app:euregbench}).  The initial poses are the same for both seeds and both modalities. The volume is a crop of the cube around the plane of the initial pose, and training draws a new initial pose in the same range for every sample. Without initialization, EUReg places no section in either modality, even within 10$^\circ$ and 5\,mm. The initial poses alone place no section. From them EUReg moves the sections toward the reference pose, from median errors of 9.7$^\circ$ and 5.6\,mm to 6.9$^\circ$~/~7.4$^\circ$ and 3.5~/~3.7\,mm within one modality and to 7.7$^\circ$~/~8.3$^\circ$ and 5.1~/~5.4\,mm cross-modally. It places only 0.057~/~0.060 and 0.007~/~0.010 of the sections. The pre-registered plan allowed a reason for the failure of the row without initialization only if the initialized row placed at least half of the sections within one modality on both seeds. It did not, so we give none. Appendix~\ref{app:euregbench} gives EUReg's accuracy on its own benchmark at the same thresholds, with the translation measured at the frame's center.   

\noindent\textbf{CoMIR-style negatives and DISA-style imitation.} The in-batch dual encoder is trained with only the other tiles of the batch as negatives (CoMIR-style, Appendix~\ref{app:mining}). It takes this choice of negatives from CoMIR~\cite{pielawski2020comir}, whose contrastive loss pulls the two modalities of an image pair together against the other pairs of the batch. Its encoders, score, temperature, and alignment term are ours. It takes none of CoMIR's 2D U-Nets, critic, rotation equivariance, or later registration within one modality, since our search slices an MRI encoded once in 3D, which a 2D encoder cannot provide. The imitation dual encoder is trained so that its dot product imitates the LC2 similarity (DISA-style). It takes this objective from DISA~\cite{ronchetti2023disa}, with LC2 computed as in DISA's code, averaged over three window sizes, and no contrastive term. Its encoders, poses, optimizer, and search are ours. It uses 2D windows of half-width 3, 5, and 7\,mm and no augmentation, while DISA trains one shared 3D network on augmented pairs with windows in voxels and registers locally. 

\noindent\textbf{Further details of Table~\ref{tab:baselines_untouched}.} For \emph{Ours}, the block-mode cell is taken from Table~\ref{tab:main} (LFB, test set). In block mode, MIND-SSC reaches 0.620 before the refinement and 0.667 after it.   Our single-section mode reaches 0.953 / 0.960 on the development sections within one modality and, with the LFB-only dual encoder of Table~\ref{tab:clean}, 0.733 / 0.753 cross-modally, a cost of 21 points.    Section~\ref{sec:baselines} reports a smaller cost for the same sections, because there the cross-modal value is that of the two-stain dual encoder of Table~\ref{tab:main_devfull}, which places more of them. Every baseline run on these sections stays below our 0.733 / 0.753 cross-modally, each learned one also trained on LFB alone. MIND-SSC reaches 0.467, anatomix 0.353 / 0.253, masked NCC 0.107, LoFTR-S2V at most 0.027 / 0.020, and pose regression 0.000. The drop of MIND-SSC from 0.467 there to 0.377 on the test set is not significant (unpaired $z=-1.29$, since the two sets share no block). A pre-registered rule says that a row transfers from the development set if it moves by at most 10 points (Appendix~\ref{app:rules}). Applied to these accuracies, the rule gives a transfer, as the drop is 9 points.    Table~\ref{tab:baselines_untouched} has two kinds of dash. The translator row's same-modality cell does not apply (Translator + masked NCC above). The block-mode cells of anatomix, LoFTR-S2V, and both EUReg rows were not run.

\section{LoFTR-S2V on its own benchmark}\label{app:nhbench}

\begin{table*}[t]
\centering
\small
\caption{LoFTR-S2V and our method on the reduced validation set of Needles \& Haystacks~\cite{frolov2025needles}, 912 problems, each slice cut from its own volume: mean average accuracy (mAA, in \%) of the angular pose error up to 5$^\circ$, 10$^\circ$, and 20$^\circ$, and the mean time per problem, from the loaded slice and volume to the pose. \emph{Paper}: the authors' published results, timed on an Intel i9-7920X CPU and an NVIDIA RTX 2080 Ti GPU, with means computed by us. The optimizer search is SLSQP from 2048 random initializations, their best method on this set (their Table 2). Their LoFTR-S2V row is the released model with GC-RANSAC and one pass of local optimization, its best row (their Table 4). \emph{Rerun}: the released model run by us at inference seeds 66~/~67, on an NVIDIA H100~/~A100 GPU. \emph{Ours}: training seeds 2032~/~2033, on an NVIDIA H100 GPU, its time without resampling the slice to a square image. Bold: the best value of each column, a pair counted by its mean.}
\label{tab:nhbench}
\begin{tabular}{@{}lccccc@{}}
\toprule
 & 5$^\circ$ & 10$^\circ$ & 20$^\circ$ & Mean & Time (s)\\
\midrule
\input{tables/tab_nh_rows}
\end{tabular}
\end{table*}

\begin{table*}[t]
\centering
\small
\caption{EUReg and our method on setting ii of the prostate ultrasound benchmark of EUReg~\cite{eureg2025}, 160 test frames from 8 test subjects. DE is the mean distance between the true and the estimated positions of the frame's center and of the four corners of a 44.8\,mm square centered on it, TE the translation error, and RE the norm of the three Euler angle errors. I-NCC and SSIM compare the frame with the volume resampled at the estimated pose, and P-NCC the estimated with the true pose parameters, all as the authors define them. Time is per frame, for the paper on an NVIDIA RTX 2080 Ti GPU with the volume encoded again for each frame in its released timing code, and for ours on an NVIDIA A100 GPU without the one-time encoding of the volume. A dash denotes a value that the paper does not report or our rerun did not record. \emph{Start pose}: the frame at the center of the volume. \emph{Paper}: the authors' published results. \emph{Rerun}: the released model with its own loss, trained by us at seeds 2032~/~2033 on NVIDIA A100 GPUs, at its final step and at the step that the released code selects on the test frames. \emph{Ours}: trained at the same two seeds. Bold: the best value of each column.}
\label{tab:euregbench}
\begin{tabular}{@{}lccccccc@{}}
\toprule
 & DE (mm) & I-NCC (\%) & SSIM (\%) & TE (mm) & RE ($^\circ$) & P-NCC (\%) & Time (ms)\\
\midrule
\input{tables/tab_eureg_rows}
\end{tabular}
\end{table*}

LoFTR-S2V~\cite{frolov2025needles} places almost no NextBrain section cross-modally (Table~\ref{tab:baselines_untouched}). To verify that this is not an artifact of our setup, we also ran it and our method on its own benchmark, Needles \& Haystacks~\cite{frolov2025needles} (Table~\ref{tab:nhbench}). Each problem provides one slice cut from a $128^3$ volume of one of many imaging domains, so slice and volume share their modality. It asks for the slice's rigid pose, with no initial pose. We use the authors' reduced validation set, the first 12 problems of each of its 76 volumes, and their metric. Its angular pose error is the larger of the rotation error and the angle between the true and the estimated translation, so it is not a distance in millimeters.

As the training script of LoFTR-S2V is not public, we ran the released checkpoint with the authors' code and settings unchanged, at inference seeds 66 (the authors' example) and 67. The rerun reproduces the published mean of 51.7\% within 0.3 points on both seeds. Our method reaches 88.1~/~87.8\% against 51.5~/~51.4\% for the rerun and 53.8\% published for the optimizer search. Averaged over the three thresholds per volume, our accuracy is higher than the released model's in 70 to 72 of the 76 volumes for each of the four pairings of seeds (exact sign test, $p<10^{-14}$). LoFTR-S2V is the fastest, 0.40~/~0.54\,s per problem in our rerun against 9.1~/~9.0\,s for our method, and the authors report 60\,s for their optimizer search.

\noindent\textbf{Changes for this benchmark.} Our single-section mode uses the same encoders as on NextBrain, trained from scratch on the benchmark's training set for a fixed 4000 updates of 16 slice-volume pairs with our loss. Each training pair has its own volume, so the wrong poses that a pair is compared with come from its own volume only. Training mines these wrong poses online instead of in two rounds, keeping at every update the poses of a new random set that score best under the current model. The search covers all rotations from 4096 initial orientations, followed by our translation search and local ascent, on one slice with no block. We fixed the settings before any validation run and used the weights of the final step. We resample each slice, whose pixel spacing can differ between its two axes, to equal spacing on a square image and assign zero weight outside it. The release's training and validation sets share one volume byte for byte. We removed it from our training (96 of 149,895 pairs). On their hidden test set, which we did not use, the authors report 52.1, 52.6, and 53.8\% for the released model with local optimization (their Table 5).

\section{EUReg on its own benchmark}\label{app:euregbench}

EUReg~\cite{eureg2025} starts each 2D ultrasound frame at the central plane of a 3D prostate volume and regresses its rigid pose. In its setting ii, each test frame is cut from its own volume, so frame and volume share their modality. The frame's pose is drawn within $\pm$10$^\circ$ about each axis, $\pm$4\,mm along both axes of the frame, and $\pm$8\,mm along its normal. The volumes are the public training and validation sets of the $\mu$-RegPro challenge~\cite{baum2023muregpro}, used as 65 training subjects and 8 test subjects. The test subjects are therefore the challenge's validation cohort, not its hidden test set. 

EUReg's paper reports a mean distance error (DE) of 1.97\,mm and 212 frames per second on an NVIDIA RTX 2080 Ti GPU (Table~\ref{tab:euregbench}). Its released training script does not run as published, so we retrained EUReg in our own loop. We used the optimizer, learning rate decay, and batch size of that script and kept the released model, loss, and frame sampler unchanged. We trained EUReg for the 600,000 updates of 6 frame-volume pairs that its paper specifies, whereas the released script runs 630,000. At its final step it reaches 2.04~/~2.20\,mm. The released code keeps the weights of the step with the lowest pose error on the test frames, and these reach 1.98~/~2.14\,mm. Our method, with the same encoders as on NextBrain, trained from scratch for 4000 updates of 16 frame-volume pairs and used at the final step, reaches 0.29~/~0.30\,mm. EUReg at its final step places 0.52~/~0.44 of the test frames within 5$^\circ$ and 2\,mm of the true pose, and our method all of them. These are the thresholds of Table~\ref{tab:baselines_untouched}, with the translation measured at the frame's center instead of the tissue center.  Averaged per subject, our DE is lower than EUReg's at the final step for all 8 test subjects in each of the four pairings of seeds (exact sign test, $p=0.008$). 

\noindent\textbf{Changes for this benchmark.} Our search covers only the range above, the range EUReg is trained on, on one frame with no block. Clamping EUReg's estimates to this range changes its DE at the final step by at most 0.01\,mm. Training mines the wrong poses of our loss online instead of in two rounds, keeping at every update the poses of a new random set that score best under the current model. We chose no weights or settings on the test frames. Before training, we read the test frames and their poses once to check the geometry of the data. Our time is 0.72~/~0.78\,s per frame on an NVIDIA A100 GPU. EUReg is considerably faster per frame, even on a slower GPU and with the volume encoded again for each frame. 

\section{Negative results}\label{app:negative}

The following ideas were tested with an earlier LFB-only dual encoder and discarded: more candidates per section, re-ranking the vote, averaging candidates, a third mining round, and larger descriptors. Encoding the whole section at once was also discarded, and Appendix~\ref{app:tiling} gives its result with the two-stain dual encoder. Appendix~\ref{app:rules} gives the pre-registered rule and outcome of each idea. 

\section{Failure analysis}\label{app:failures}

\noindent\textbf{What fails.} Figures~\ref{fig:failures} and~\ref{fig:failures_dev} show every failed block of seed~0 in block mode.  On the 300 test sections, 36 / 44 sections miss the joint criterion (seed~0~/~seed~1), in 11 / 14 of the 68 blocks: 27 / 40 on rotation, in 7 / 10 blocks, by $5.0^\circ$ to $9.6^\circ$ / $5.1^\circ$ to $93.7^\circ$, and 9 / 4 on translation only, by 2.0 to 3.6 / 2.0 to 2.4\,mm. Only seed 1 has gross misses on the test sections, 13 in two blocks of one hemisphere, with rotation errors of about $93^\circ$. On the 150 development sections, 12 / 17 sections fail, in 6 / 8 blocks: 12 / 15 on rotation, in 6 / 7 blocks, and 0 / 2 on translation only, by 2.1 to 2.8\,mm on seed 1. One development block per seed has a rotation error above $30^\circ$ ($55^\circ$ / $105^\circ$).  When the rotation is correct, the translation is correct 97\% / 98\% of the time on the test sections and 100\% / 99\% on the development ones.  Figure~\ref{fig:qualitative_s1} repeats Figure~\ref{fig:qualitative} for the second seed. 

\noindent\textbf{Where.} The final refinement skips blocks whose sections contain less than 10\% tissue. Most blocks that fail on rotation are among them: 4 of 7 / 6 of 10 rotation-failed blocks on the test sections and 5 of 6 / 4 of 7 on the development ones. Two of the four blocks with a rotation error above $30^\circ$ are among them too (5 to 8\% tissue, the other two 12\% and 24\%).  Little tissue also weakens the search (Section~\ref{sec:serial}). 

\begin{figure*}[t]
\centering
\includegraphics[width=0.8\textwidth]{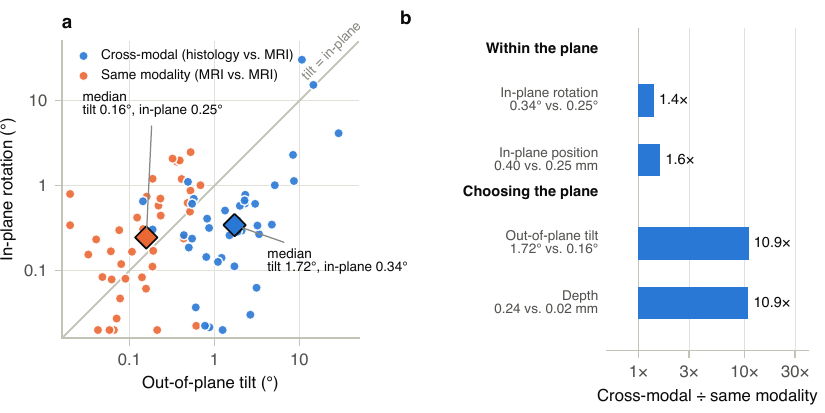}
\caption{Where the refinement reaches its MIND-SSC peak when initialized at the reference pose that training uses (Appendix~\ref{app:protocol}), for 40 blocks of one hemisphere, cross-modally and in a same-modality control. Errors are measured as in Table~\ref{tab:main}. A block's rotation error is taken against the mean of the reference rotations of its 8 sections, and each component of its translation error is the median over these sections. (a) Out-of-plane tilt versus in-plane rotation error per block (log scales, values below 0.02$^\circ$ drawn at 0.02$^\circ$, diamonds: medians). (b) Cross-modal median divided by same-modality median per component of the pose error. The modality gap increases the median out-of-plane errors (tilt and depth, 10.9 times each) considerably more than the in-plane ones (rotation 1.4 times, position 1.6 times).}
\label{fig:mechanism}
\end{figure*}

\noindent\textbf{How.} In the failed blocks the residual rotation is mostly out-of-plane tilt. Tilt exceeds the in-plane rotation in 5 of 7 / 10 of 10 rotation-failed test blocks and in 4 of 6 / 7 of 7 development ones. Initialized at the reference pose that training uses (Appendix~\ref{app:protocol}), the refinement reaches its MIND-SSC peak at a tilt error 11 times larger cross-modally than in a same-modality control (Figure~\ref{fig:mechanism}).  The cutting plane is only weakly constrained, because the anatomy varies slowly along its normal (Section~\ref{sec:intro}), which is likely why the modality gap first degrades the out-of-plane components (tilt and depth).

\input{tables/fig_failures}
\input{tables/fig_failures_dev}
\input{tables/fig_qualitative_s1}

\section{Runtime and compute}\label{app:runtime}

\noindent\textbf{Per section.} Each method of Table~\ref{tab:runtime} records its search time per section in its result file, and the runs of EUReg record none. For our single-section search the five hemispheres agree within 6\,s (medians 23.0 to 28.3\,s on seed 0 and 22.7 to 28.1\,s on seed 1). Same-modality times are close to the cross-modal ones (ours 23\,s, masked NCC 39\,s, MIND-SSC 77\,s). 

\noindent\textbf{Block pipeline.} Block mode solves up to 16 sections of a block, not only the scored ones. Timed on the LFB-only runs, a block of 16 sections takes 522\,s (Table~\ref{tab:runtime}), about 33\,s per section. For all five hemispheres this is 10.0 / 9.9 GPU-hours (seed 0 / seed 1) on the development blocks and 9.8 / 9.8 on the test blocks. The block refinement adds under 3 minutes per seed. The single-section runs take 1.0 / 1.1 GPU-hours for the 150 development sections and 2.1 / 2.0 for the 300 test sections. 

\clearpage
\noindent\textbf{Stages.} Table~\ref{tab:stages} times every stage of block mode, and every timed run reproduced the archived poses of its sections. The global search takes most of a block's time and the re-solve at the block's rotation most of the rest. Within the search, the first round of local ascent takes the longest. The timed median block, 512\,s, agrees with the 522\,s derived above. 

\noindent\textbf{Direct convolution.} On one A100 GPU, we replace the FFT of the coarse sweep with a dense direct convolution in 32-bit floats. We run both on random descriptors at the size of a section with 15 tiles. The sweep of all 2048 rotations then takes 70.2\,s against 7.0\,s, and the two score fields agree to about $10^{-6}$ of the largest score. The 25\,s search would therefore take about 88\,s, which Section~\ref{sec:serial} rounds to about 90\,s. We derived this number from the two sweep times and did not time the search end to end. With the GPU's reduced-precision convolution (TF32), the sweep takes 24.9\,s and the search would take about 43\,s. We did not check the scores of this faster convolution.    

\noindent\textbf{Training.} Timed on the LFB-only runs, one fold of the clean protocol trains in 1.6 GPU-hours (1.2 to 2.5 over the five folds). Of this time, the translator takes 0.56, the two mining rounds 0.33 and 0.40, the first dual encoder 0.08, and the two final dual encoders (one per seed) 0.25. All five folds take 8.1 GPU-hours.

\noindent\textbf{Memory.} Every method of Table~\ref{tab:runtime} uses less than 5\,GiB of GPU memory (6.6 and 6.9\,GiB reserved by the allocator for our search on a section and a block). The block refinement requires less than 0.1\,GiB. Each method was measured unmodified, in its LFB-only run, on the hemisphere containing the largest section (23 tiles). 

\noindent\textbf{How these were measured.} The GPU-hours are computed from the timestamps each pipeline step writes to its log, checked against the scheduler's accounting of the same job. The logged steps cover 99\% of the accounted time in every run, all on A100 GPUs.

\begin{table}[t]
\centering\footnotesize\setlength{\tabcolsep}{3pt}
\caption{Runtime per section (last row: per block) and peak GPU memory on one A100 GPU, cross-modal, on the 300 test sections. Median and 10th to 90th percentile over sections, seed 0 / seed 1 where a method has seeds. Times are taken from the runs of Table~\ref{tab:baselines_untouched} (learned methods trained on both stains) and cover the pose search on already-encoded inputs, without file reading or the one-time encoding of the volume. Memory is the peak allocated on one hemisphere (seed 0, masked NCC and MIND-SSC on its first 20 sections, block mode on its two blocks with the most tiles). $^\ast$Measured on the LFB-only run of the same method. $^\dagger$One whole block of 16 serial sections (the median, 5 to 16 are solved per block), with the vote and the re-solve of every section at the block's rotation. The archived runs record no per-block time, so this is a whole solve's time divided by its number of blocks, with the range over ten solves (five hemispheres, two seeds).}
\label{tab:runtime}
\begin{tabular}{@{}lccc@{}}
\toprule
Method & Median (s) & 10--90\% (s) & Memory (GiB)\\
\midrule
\input{sections/tab_runtime_rows}\end{tabular}
\end{table}

\begin{table*}[t]
\centering\footnotesize\setlength{\tabcolsep}{4pt}
\caption{Time of each stage of block mode on one A100 GPU over 20 timed blocks (ten test blocks, two per hemisphere, both seeds). Median and 10th to 90th percentile over each stage's units, and the stage's share of a block's solve time without the one-time encoding of the volume (median over blocks). The two volume rows give their share of a hemisphere's run. The refinement runs after the solve, outside the first row.}
\label{tab:stages}
\begin{tabular}{@{}llccc@{}}
\toprule
Stage & Unit & Median (s) & 10--90\% (s) & Share\\
\midrule
\input{sections/tab_stage_rows}\end{tabular}
\end{table*}

%% file: tables/tab_content.tex
\begin{table*}[t]
\centering\footnotesize\setlength{\tabcolsep}{4pt}
\caption{Serial sections versus a single section as content is removed: fraction of sections within 5$^\circ$ of the reference rotation, and within 5$^\circ$ and 2\,mm (joint), mean over the 68 development blocks, seed 0 / seed 1. No refinement is applied, for two reasons. First, the Single columns have none, because the refinement adjusts all sections of a block together. Second, the refinement runs only on blocks whose sections contain sufficient tissue, so as tissue is removed it would be skipped more often, which would overstate the drop. The intact row is one run shared by all three kinds of removal, and the Single columns do not change with the number of sections. The Single columns report the first stage of block mode, which solves each section alone with equal tile weights. Their intact row lies slightly below the single-section mode of Table~\ref{tab:main_devfull}, which weights tiles by their foreground and averages over sections rather than blocks.}
\label{tab:content}
\begin{tabular}{@{}llcccc@{}}
\toprule
& & \multicolumn{2}{c}{Rotation $<5^\circ$} & \multicolumn{2}{c}{Joint}\\
\cmidrule(lr){3-4}\cmidrule(l){5-6}
Content removed & Level & Block & Single & Block & Single\\
\midrule
None (intact) & none & 0.919/0.912 & 0.735/0.740 & 0.919/0.907 & 0.713/0.740\\
\addlinespace
Fewer tiles & 12 tiles & 0.875/0.868 & 0.712/0.729 & 0.875/0.859 & 0.705/0.722\\
 & 8 tiles & 0.794/0.853 & 0.586/0.560 & 0.787/0.826 & 0.574/0.560\\
 & 6 tiles & 0.706/0.757 & 0.456/0.482 & 0.695/0.724 & 0.433/0.467\\
\addlinespace
Less tissue & 75\% kept & 0.833/0.819 & 0.598/0.621 & 0.833/0.806 & 0.587/0.614\\
 & 50\% kept & 0.417/0.402 & 0.195/0.163 & 0.395/0.350 & 0.195/0.145\\
 & 25\% kept & 0.157/0.037 & 0.054/0.016 & 0.137/0.029 & 0.039/0.016\\
\addlinespace
Fewer sections & 8 per block & 0.897/0.912 & 0.735/0.740 & 0.863/0.865 & 0.713/0.740\\
 & 4 per block & 0.868/0.860 & 0.735/0.740 & 0.836/0.833 & 0.713/0.740\\
\bottomrule
\end{tabular}
\end{table*}

%% file: tables/fig_qualitative_he.tex
\begin{figure*}[tp]
\centering
\includegraphics[width=\textwidth]{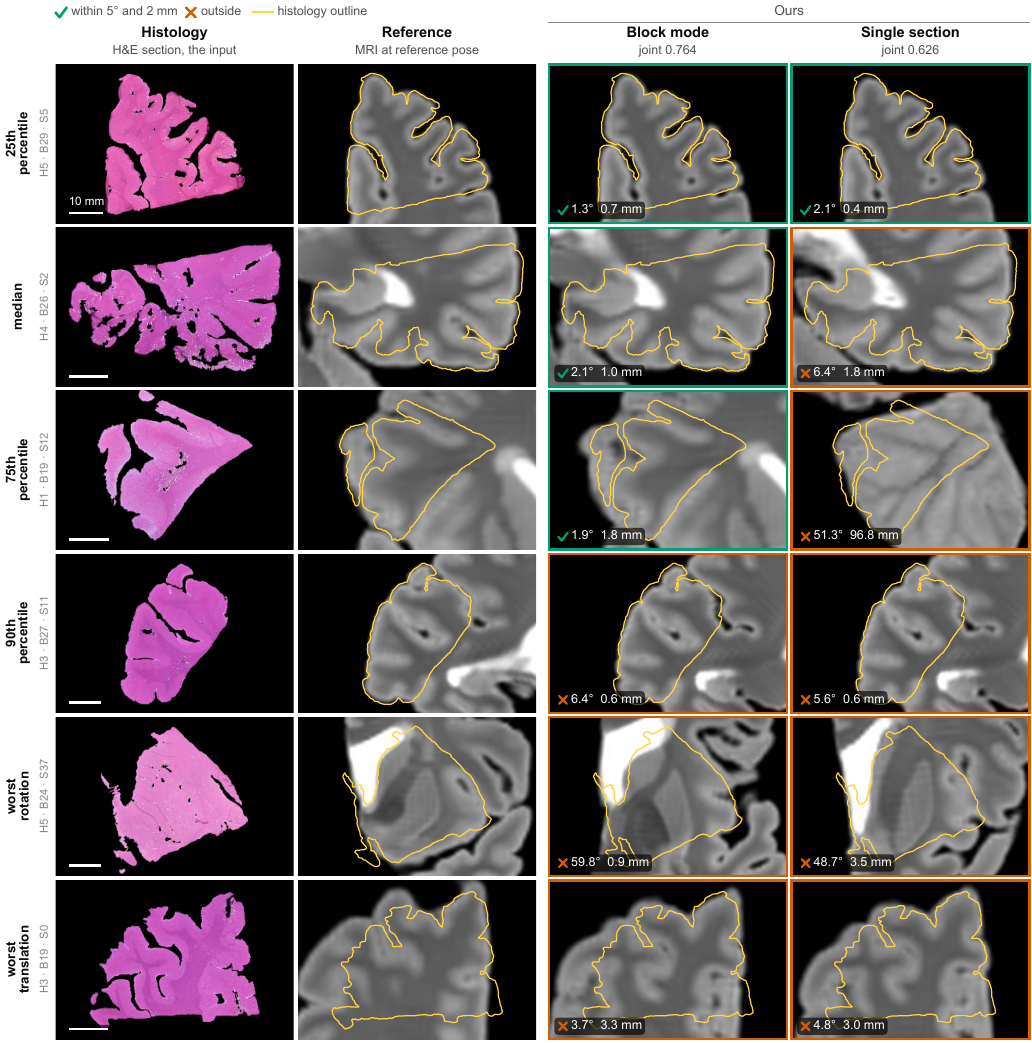}
\caption{\textbf{Registrations of H\&E sections on the test set, failures included.} Rows: six of its 297 H\&E sections, chosen by the rank of their error in block mode under the rule of Figure~\ref{fig:qualitative}, seed 0. Both modes use the two-stain dual encoder of Table~\ref{tab:main}. Cells as in Figure~\ref{fig:qualitative}, with the H\&E section as the input. A cell is black where its plane misses the brain. No baseline was evaluated on H\&E sections, so only our two modes are shown. The second seed: Figure~\ref{fig:qualitative_he_s1}. H, B, S: hemisphere, block, section. Scale bars: 10\,mm.}
\label{fig:qualitative_he}
\end{figure*}

%% file: tables/fig_qualitative_he_s1.tex
\begin{figure*}[tp]
\centering
\includegraphics[width=\textwidth]{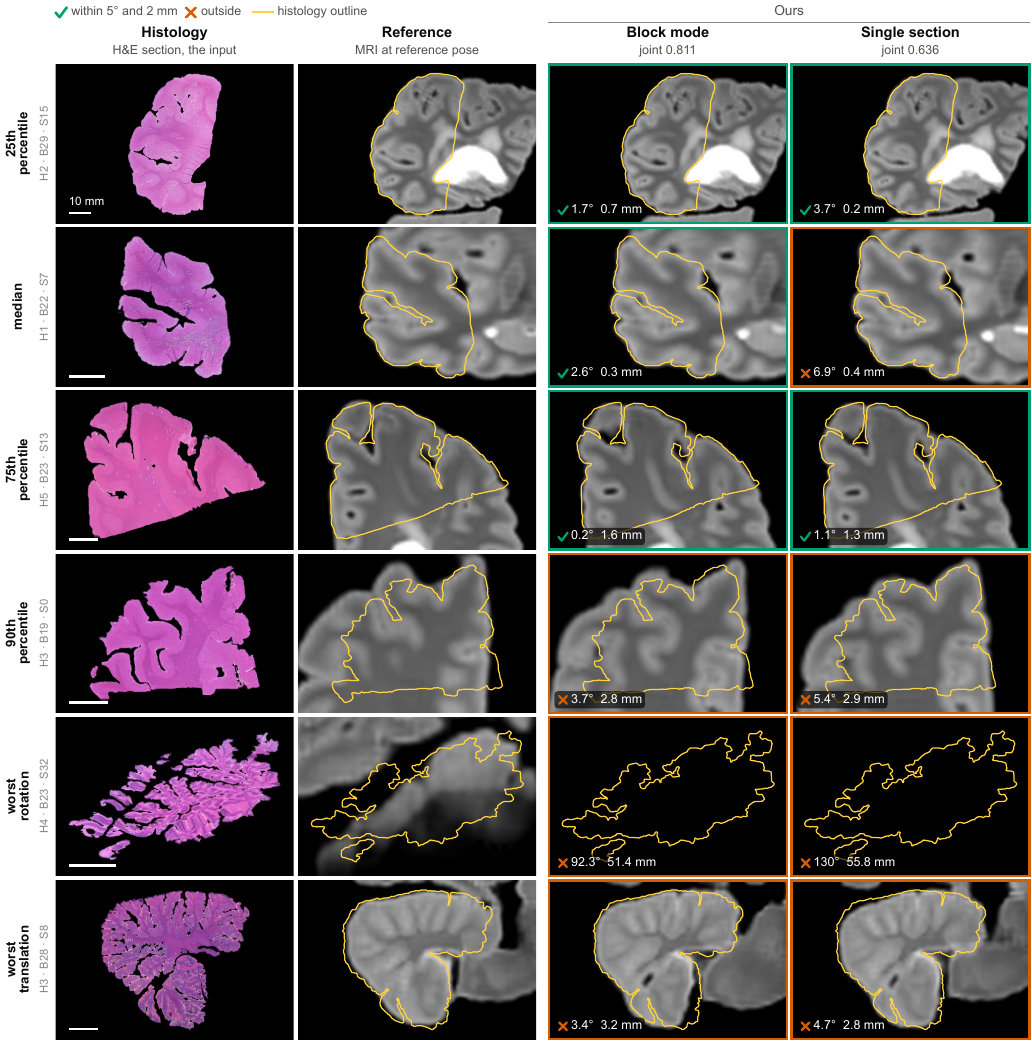}
\caption{\textbf{Registrations of H\&E sections on the test set, failures included.} Rows: six of its 297 H\&E sections, chosen by the rank of their error in block mode under the rule of Figure~\ref{fig:qualitative}, seed 1. Both modes use the two-stain dual encoder of Table~\ref{tab:main}. Cells as in Figure~\ref{fig:qualitative}, with the H\&E section as the input. A cell is black where its plane misses the brain. No baseline was evaluated on H\&E sections, so only our two modes are shown. The first seed: Figure~\ref{fig:qualitative_he}. H, B, S: hemisphere, block, section. Scale bars: 10\,mm.}
\label{fig:qualitative_he_s1}
\end{figure*}

%% file: sections/tab_tiling.tex
\begin{tabular}{@{}lcccc@{}}
\toprule
Encoding & Rot.\ $<5^\circ$ & Joint & Change & $p$\\
\midrule
\multicolumn{5}{@{}l}{LFB sections, $n = 150$, 68 blocks}\\
\quad Tiles & \textbf{0.753/0.760} & \textbf{0.747/0.760} & & \\
\quad Whole section & 0.633/0.560 & 0.620/0.547 & $-12.0$/$-20.0$ & $<10^{-4}$\\
\midrule
\multicolumn{5}{@{}l}{H\&E sections, $n = 147$, 67 blocks}\\
\quad Tiles & \textbf{0.633/0.694} & \textbf{0.619/0.680} & & \\
\quad Whole section & 0.524/0.483 & 0.510/0.456 & $-10.9$/$-21.1$ & $<10^{-4}$\\
\bottomrule
\end{tabular}

%% file: tables/tab_mining_rows.tex
Wrong poses per tile & Joint & \\
\midrule
32 mined, round 2 (ours) & \textbf{0.747/0.760} & \\
32 random & 0.600/0.620 & \\
8 random & 0.600/0.600 & \\
8 mined, round 1 & 0.587/0.620 & \\
\midrule
Change & Points & $p$\\
\midrule
Ours minus 32 random & $+14.7$/$+14.0$ & 0.00028/0.0094\\
Round 1 minus 8 random & $-1.3$/$+2.0$ & 0.81/0.61\\
Ours minus round 1 & $+16.0$/$+14.0$ & 0.00082/0.0043\\
\bottomrule

%% file: tables/tab_nomining_rows.tex
Dual encoder & Joint & \\
\midrule
In-batch negatives only (CoMIR-style) & 0.587/0.620 & \\
Similarity imitation (DISA-style) & 0.027/0.033 & \\
\midrule
Change & Points & $p$\\
\midrule
Ours minus in-batch only & $+16.0$/$+14.0$ & 0.0013/0.0015\\
In-batch only minus MIND-SSC & $+12.0$/$+15.3$ & 0.13/0.035\\
32 random minus in-batch only & $+1.3$/$0.0$ & 0.83/1.0\\
Ours minus imitation & $+72.0$/$+72.7$ & $<10^{-4}$/$<10^{-4}$\\
Imitation minus MIND-SSC & $-44.0$/$-43.3$ & $<10^{-4}$/$<10^{-4}$\\
32 random minus imitation & $+57.3$/$+58.7$ & $<10^{-4}$/$<10^{-4}$\\
\bottomrule

%% file: tables/tab_nh_rows.tex
Optimizer search, paper & 53.1 & 53.6 & 54.6 & 53.8 & 60\\
LoFTR-S2V, paper & 50.9 & 51.4 & 52.7 & 51.7 & 0.49\\
LoFTR-S2V, rerun & 50.8/50.6 & 51.2/51.0 & 52.5/52.7 & 51.5/51.4 & \textbf{0.40/0.54}\\
\midrule
Ours & \textbf{85.4/84.9} & \textbf{88.5/88.1} & \textbf{90.4/90.3} & \textbf{88.1/87.8} & 9.1/9.0\\
\bottomrule

%% file: tables/tab_eureg_rows.tex
Start pose, paper & 6.46 & 61.77 & 31.28 & 5.33 & 9.59 &--&--\\
EUReg, paper & 1.97 & 85.72 & 74.69 & 1.29 & 4.59 & 89.79 & \textbf{4.7}\\
EUReg, rerun & 2.04/2.20 & 85.57/84.89 & 74.49/72.40 & 1.30/1.39 & 4.88/5.46 & 88.63/85.12 &--\\
EUReg, rerun, step chosen on test & 1.98/2.14 & 85.60/84.73 & 74.68/71.63 & 1.29/1.42 & 4.67/4.98 & 89.59/87.40 &--\\
\midrule
Ours & \textbf{0.29/0.30} & \textbf{90.65/90.61} & \textbf{96.21/96.20} & \textbf{0.16/0.16} & \textbf{0.68/0.71} & \textbf{99.83/99.81} & 719/782\\
\bottomrule

%% file: tables/fig_failures.tex
\begin{figure*}[p]
\centering
\includegraphics[width=\textwidth]{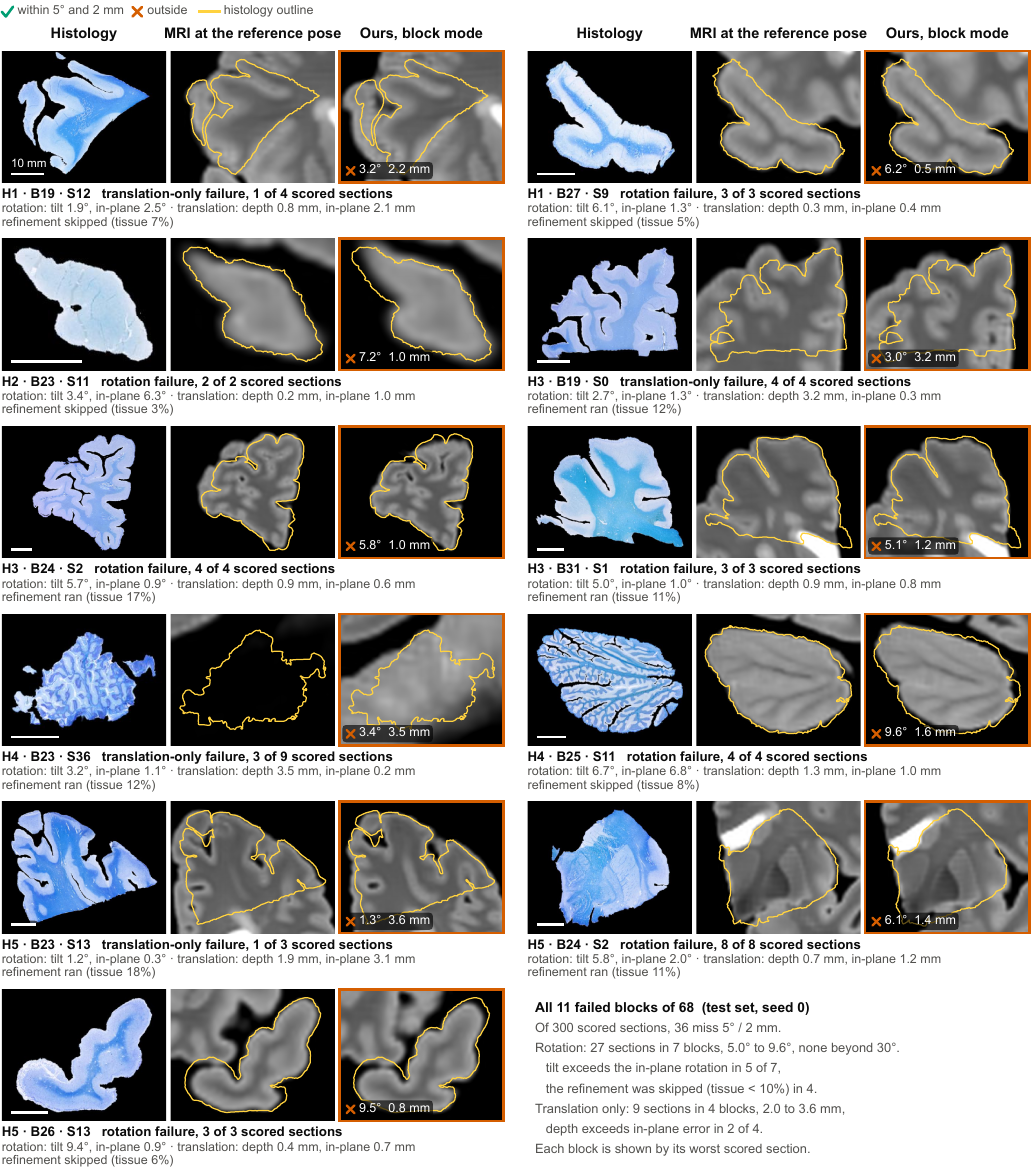}
\caption{\textbf{Every failed block in block mode on the test sections (seed 0). Nothing is selected.} Of the 300 scored sections, 36 miss the joint criterion ($5^\circ$ and 2\,mm), in 11 of 68 blocks, and each block is shown by its scored section with the largest error. Rotation fails for 27 sections in 7 blocks (5.0$^\circ$ to 9.6$^\circ$, none beyond $30^\circ$). The out-of-plane tilt exceeds the in-plane rotation in 5 of these 7 blocks. In 4 of the 7, the final refinement did not run, because its gate skips blocks whose sections contain less than 10\% tissue. Translation alone fails for 9 sections in 4 blocks (2.0 to 3.6\,mm), with the depth error above the in-plane error in 2 of 4. Cells, marks, and labels as in Figure~\ref{fig:qualitative}. A cell is black where its plane misses the brain.}
\label{fig:failures}
\end{figure*}

%% file: tables/fig_failures_dev.tex
\begin{figure*}[p]
\centering
\includegraphics[width=\textwidth]{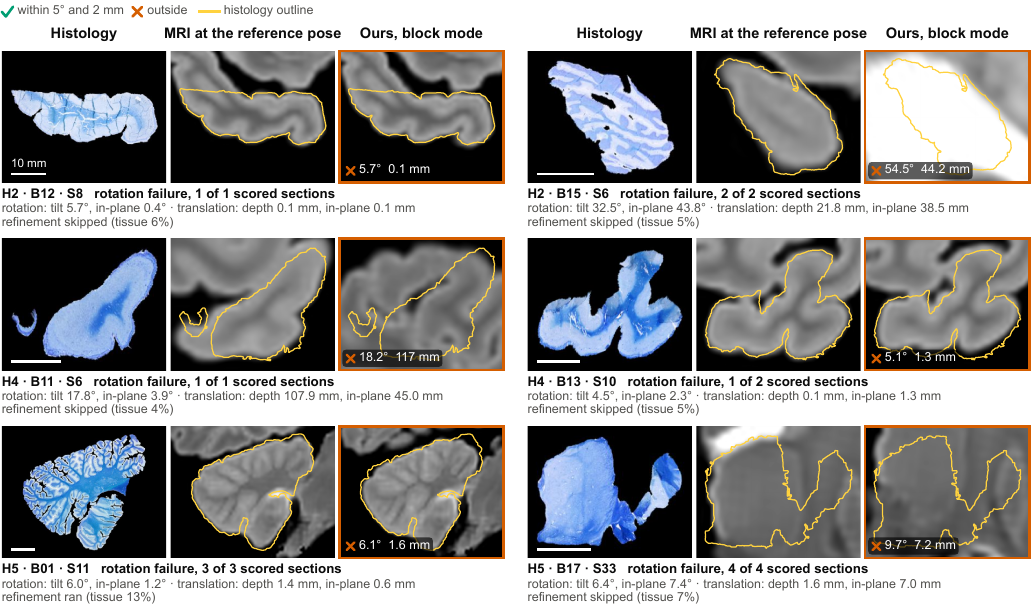}
\caption{\textbf{Every failed block in block mode on the development sections (seed 0). Nothing is selected.} Of the 150 scored sections, 12 miss the joint criterion ($5^\circ$ and 2\,mm), in 6 of 68 blocks, and each block is shown by its scored section with the largest error. Rotation fails for 12 sections in 6 blocks (5.1$^\circ$ to 54.7$^\circ$, 2 sections beyond $30^\circ$). The out-of-plane tilt exceeds the in-plane rotation in 4 of these 6 blocks. In 5 of the 6, the final refinement did not run, because its gate skips blocks whose sections contain less than 10\% tissue. The one block beyond $30^\circ$ is among these 5, with 5\% tissue. No section fails on translation alone. Cells, marks, and labels as in Figure~\ref{fig:qualitative}. A cell is black where its plane misses the brain.}
\label{fig:failures_dev}
\end{figure*}

%% file: tables/fig_qualitative_s1.tex
\begin{figure*}[t]
\centering
\includegraphics[width=\textwidth]{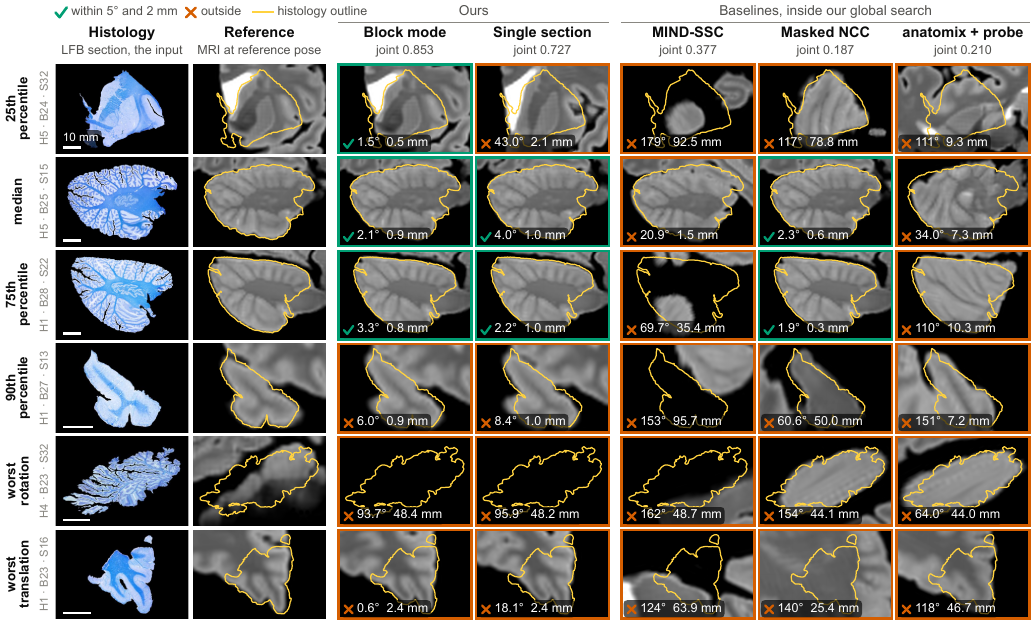}
\caption{\textbf{Registrations on the test set, failures included.} Rows: six of its 300 sections, chosen by the rank of their error in block mode under a pre-registered rule (Appendix~\ref{app:rules}), seed 1. Each cell: the 1\,mm MRI re-sliced on the plane that method returned, under the histology outline, with that section's rotation and translation error. \emph{Block mode} uses the block's serial sections, all other columns this section alone. A cell is black where its plane misses the brain. H, B, S: hemisphere, block, section. Scale bars: 10\,mm.}
\label{fig:qualitative_s1}
\end{figure*}

%% file: sections/tab_runtime_rows.tex
Masked NCC & 45.1 & 29--58 & 2.2\\
MIND-SSC & 72.4 & 47--94 & 3.2\\
anatomix + linear probe & 25.1/25.0 & 16--33 & 4.9$^{\ast}$\\
LoFTR-S2V (fine-tuned) & 0.24/0.25 & 0.20--0.31 & 1.9$^{\ast}$\\
Ours, single section & 25.0/25.1 & 17--32 & 4.7$^{\ast}$\\
Ours, one block (16 sections)$^\dagger$ & 522$^{\ast}$ & 477--574$^{\dagger\ast}$ & 4.9$^{\ast}$\\
\bottomrule

%% file: sections/tab_stage_rows.tex
One block (solve) & block & 511.6 & 417--653 & 100.0\%\\
Volume loading and 3D encoding & hemisphere & 0.45 & 0.36--0.53 & 0.01\% of a hemisphere\\
Section loading & section & 0.02 & 0.02--0.02 & 0.1\%\\
Tiling & section & 0.02 & 0.01--0.11 & 0.2\%\\
2D encoding & section & 0.02 & 0.01--0.02 & 0.1\%\\
Global search & section & 24.1 & 18--30 & 72.9\%\\
\quad coarse sweep of every pose & section & 7.2 & 5.3--9.1 & 21.7\%\\
\quad finer pass over the kept poses, then basins & section & 4.4 & 3.5--5.4 & 13.5\%\\
\quad local ascent, round 1 & section & 10.8 & 8--14 & 32.6\%\\
\quad local ascent, round 2 & section & 1.7 & 1.3--2.2 & 5.3\%\\
Block vote & block & 0.18 & 0.15--0.21 & ${<}0.1$\%\\
Re-solve at the block rotation & section & 8.5 & 7--11 & 26.8\%\\
Line fit & block & 0.03 & 0.02--0.03 & ${<}0.1$\%\\
Write-out & block & 0.03 & 0.02--0.03 & ${<}0.1$\%\\
Refinement: volume & hemisphere & 0.45 & 0.37--0.52 & 0.01\% of a hemisphere\\
Refinement: section targets & section & 0.06 & 0.03--0.09 & 0.2\%\\
Refinement: block ascent & block & 0.57 & 0.47--0.65 & 0.1\%\\
Refinement: write-back & block & 0.39 & 0.33--0.48 & 0.1\%\\
\bottomrule

%% file: sections/supp_adni.tex
\section{Synthetic data (ADNI) details}\label{app:adni}
\input{tables/tab_adni}

\noindent\textbf{Setup.} From in vivo ADNI scans we render blocks of 8 sections at known poses, each block at an orientation drawn uniformly over all 3D rotations. FLAIR-like sections are registered into a T2-like volume (cross-contrast condition), or T2-like into T2-like (same-contrast control), all synthesized with SuperSynth (Section~\ref{sec:setup}).  The models are trained with the same recipe on 15 ADNI subjects (an earlier LFB-only dual encoder trained on NextBrain scores 0.000 on this contrast pair). Results are on 4 held-out subjects, 32 blocks per setting. As on NextBrain, translation is measured at the center of the tissue, here that of the section before it is warped. The reference is the exact pose of the rendered block.  In Figure~\ref{fig:adni}, each band is $\pm$1 standard error from a bootstrap over subjects, then blocks. 

\noindent\textbf{Full ladder and alignment errors.}  Table~\ref{tab:adni} gives the joint accuracy after the refinement for both deformation families and both seeds, and Section~\ref{sec:adni} discusses it. From 0 to 2\,mm the same-contrast control loses less than the cross-contrast condition in three of the four family and seed pairs. With elastic warps on seed 0, however, it loses 4.7 points against 3.9. At 3\,mm it has lost less in all four (Section~\ref{sec:adni}).  The residual error of NextBrain's own alignment is 0.99~$\pm$~0.51\,mm~\cite{casamitjana2025nextbrain}, a 3D error at manually placed landmarks, which includes annotation noise. The dataset releases its sections already nonlinearly aligned to the MRI, so this is the misalignment that remains between a real section and the MRI at its reference pose. On the severity axis of Table~\ref{tab:adni}, a released section is deformed by at least the stretch of the block's affine map:   median 0.16\,mm, 10th to 90th percentile 0.02 to 0.55\,mm. These values cover 4462 sections of 200 blocks (36 blank sections excluded), and 1.1\% of them lie above 1\,mm. For raw sections, before any such correction, two studies measured landmarks against the unsectioned tissue. For 50\,\textmu m sections of a primate brain (291 landmarks), the error was 1.08 $\pm$ 1.16\,mm after linear alignment and fell to 0.32 $\pm$ 0.28\,mm after nonlinear registration~\cite{choe2011accuracy}. For 4\,\textmu m paraffin sections of human prostate (162 landmarks), it was 1.44 $\pm$ 0.73\,mm after rigid, 0.56 $\pm$ 0.31\,mm after similarity, and 0.50 $\pm$ 0.28\,mm after affine alignment. Most of the rigid residual is shrinkage~\cite{gibson2013prostate}.  For brain sections against MRI, surface distances in nine macaques were 0.76 $\pm$ 0.41\,mm (boundary of gray matter and cerebrospinal fluid) and 0.54 $\pm$ 0.39\,mm (boundary of white and gray matter) after affine alignment. They fell to 0.39 and 0.28\,mm after diffeomorphic registration~\cite{ceritoglu2010lddmm}. In human whole-hemisphere sections, nonlinear registration lowered the median landmark error by only 19\% from the initial affine alignment, to 0.88\,mm (641 landmarks, intra- and inter-observer variability 0.71 and 0.79\,mm)~\cite{casamitjana2022robust}.  To our knowledge, no study directly reports the landmark residual of human brain paraffin sections after linear alignment, and each distribution that this paragraph gives with a standard deviation is wide. Each standard deviation is between about half of its mean and about its mean.    Panel (a) of Figure~\ref{fig:adni} shows one section at each severity of the elastic ladder, as input to the method, with the displacement field that warped it.

\noindent\textbf{Confirmation on new subjects.} On the 12 new subjects, accuracy undeformed and at 2\,mm (elastic and free-form warps together) is at or above the development level in seven of the eight cells (two conditions, two severities, two seeds). In the eighth, the same-contrast control undeformed on seed 1, accuracy is within two standard errors below that level. Applied to these accuracies, the pre-registered rule (Appendix~\ref{app:rules}) gives \emph{confirmed} on both seeds: cross-contrast 0.999 / 0.978 undeformed and 0.911 / 0.907 at 2\,mm.  

\noindent\textbf{Acquired pair.} The last check of Section~\ref{sec:adni} uses no synthesis. Sections cut from a subject's acquired FLAIR scan are registered into the same subject's acquired T1-weighted scan (cross-contrast condition), and sections cut from the T1-weighted scan into that scan (same-contrast control). Each condition is run undeformed, at 2\,mm elastic, and at 2\,mm free-form, which gives six settings. The two scans are separate acquisitions, aligned to each other by a rigid fit of their segmentations. Their residual misalignment affects every translation reference of this pair, so only rotation is compared. The 3 subjects are rendered with the block seeds of their synthetic pair, so each setting is paired block by block (24 blocks, 192 sections). On seed 0, rotation accuracy in the cross-contrast condition is 1.000 undeformed, 0.917 at 2\,mm elastic, and 0.875 at 2\,mm free-form, against 0.958, 0.958, and 0.917 for the synthetic pair. In the same-contrast control it is 1.000 in all three settings, against 1.000, 1.000, and 0.958. No difference exceeds 4.2 points, and the paired block-level $p$ is 1.0 in all six settings.  The acquired pair has its own dual encoders, trained with the same recipe on the acquired scans of 13 other subjects. In these settings the acquired pair uses its first dual encoder, the one trained on its round-1 poses, and the synthetic pair its final dual encoder.   Round 1 was chosen on six other subjects, at the second of two pre-declared looks: its rotation accuracy was higher in 6 of their 48 blocks and lower in none ($p = 0.031$). No correction for the second look was fixed in advance, and this $p$ passes an O'Brien-Fleming bound but narrowly fails a Pocock bound. Round 1 is also the round under which the acquired pair agrees with the synthetic pair. With the final dual encoder on both pairs, rotation accuracy on the acquired pair is 0.708, 0.375, and 0.250 in the cross-contrast condition. That is 25, 58, and 67 points below the synthetic pair ($p = 0.070$, 0.0005, and 0.0001). In the same-contrast control it is 1.000, 0.917, and 0.833 ($p = 1.0$, 0.5, and 0.375).

%% file: tables/tab_adni.tex
\begin{table*}[t]
\centering\footnotesize\setlength{\tabcolsep}{4pt}
\caption{ADNI deformation ladder: joint accuracy after the refinement, seed 0 / seed 1, 32 blocks of 4 held-out subjects per cell. Elastic: smoothed random displacements. Free-form: a B-spline lattice. Severity is the RMS displacement left after removing the best rigid motion, and 0\,mm is the same undeformed run for both families. Errors are measured against the exact pose of the rendered block, translation at the tissue center of the undeformed section.}
\label{tab:adni}
\begin{tabular}{@{}llccccc@{}}
\toprule
& & \multicolumn{5}{c}{Deformation (mm)}\\
\cmidrule(l){3-7}
Condition & Family & 0 & 0.5 & 1 & 2 & 3\\
\midrule
Cross-contrast & elastic & 0.938/0.969 & 0.938/0.969 & 0.930/0.965 & 0.898/0.859 & 0.480/0.402\\
 & free-form &  & 0.938/0.969 & 0.930/0.934 & 0.852/0.828 & 0.496/0.387\\
\addlinespace
Same-contrast & elastic & 0.992/1.000 & 0.992/1.000 & 0.984/0.996 & 0.945/0.945 & 0.621/0.668\\
 & free-form &  & 0.992/1.000 & 0.988/0.996 & 0.957/0.965 & 0.754/0.754\\
\bottomrule
\end{tabular}
\end{table*}

%% file: sections/supp_cue.tex
\section{Optional human cue}\label{app:cue}

Every accuracy in this appendix is measured as in Table~\ref{tab:main}. 

\begin{table*}[t]
\centering\footnotesize\setlength{\tabcolsep}{5pt}
\caption{Optional human cue with a simulated user, on the 300 test sections, seed 0 / seed 1, each run paired with the cue-free run of~Table~\ref{tab:main}. Joint accuracy is rotation $<5^\circ$ and translation $<2$\,mm. Errors are measured as in Table~\ref{tab:main}. \emph{Hard}: the search is confined to the cued region. \emph{Soft}: a penalty outside it. Gross misses are sections beyond 30$^\circ$ or 20\,mm, of 300. A dash means not run.}
\label{tab:cue}
\begin{tabular}{@{}lcccccc@{}}
\toprule
 & \multicolumn{4}{c}{Single section} & \multicolumn{2}{c}{Block mode}\\
\cmidrule(lr){2-5}\cmidrule(l){6-7}
Cue & Joint, hard & Joint, soft & Gross misses, hard & Within 10$^\circ$ and 5\,mm, hard & Joint, hard & Joint, soft\\
\midrule
\input{sections/tab_cue_rows}\end{tabular}
\end{table*}

\begin{table}[t]
\centering\footnotesize\setlength{\tabcolsep}{4pt}
\caption{A wrong user, soft cue, single section: every cue is offset by 1.5 times the size of its region, so the reference pose lies outside it. Joint accuracy on the 300 test sections, seed 0 / seed 1, against 0.707 / 0.727 without a cue. Right user: the soft cue of Table~\ref{tab:cue}. Fixed and Broken: sections that the wrong user's cue turns from a miss into a hit, and the reverse.}
\label{tab:cue_wrong}
\begin{tabular}{@{}lcccc@{}}
\toprule
Cue & Right user & Wrong user & Fixed & Broken\\
\midrule
\input{sections/tab_cue_wrong_rows}\end{tabular}
\end{table}

\noindent\textbf{Two cues.} A position cue is a ball of radius $r$\,mm around an approximate position of the center of the section's tiles. An orientation cue keeps the poses whose rotation lies within $\phi$ degrees of an approximate orientation of the cutting plane. Its plane-only variant bounds only the normal of the plane and leaves the in-plane rotation free. The two cues can be combined. 

\noindent\textbf{Integration of the cue.} The cue acts inside the unchanged search, before its candidates are ranked, and nothing is retrained. The hard form treats the region as a strict constraint. Its global search sweeps only the grid rotations near the cue and keeps only the translations inside the ball, and the later stages of the funnel are confined to the region.  The soft form subtracts from the score a penalty that is zero inside the region and grows with the squared distance outside it. Its weight, 0.031, was pre-registered at the measured scale of the score's extreme values and not tuned. The soft form's global search covers a window three region sizes wide, and the returned pose may lie outside the region.  In block mode the cue acts on every section's search, the re-solve at the block's rotation keeps only the position cue, and the final refinement is not constrained. With both cues disabled, the search is bit-identical to the cue-free search. 

\noindent\textbf{Simulated user.} No human participant was involved. The cue is drawn around the reference pose that training uses (Appendix~\ref{app:protocol}), with a random error, uniform inside the region and never at its center. That pose is always inside the cue. Each tissue block has one such error, shared by its sections, both modes, both seeds, and every cue size. The reference pose of the evaluation, the rigid pose that best fits NextBrain's alignment over the section's tissue (Section~\ref{sec:setup}), lies outside the cue on 29 of the 300 sections at 10\,mm with 10$^\circ$, 16 by orientation and 13 by position, and on at most 16 at any other cue size. It lies at most 2.8$^\circ$ beyond the orientation bound and 0.4\,mm beyond the position bound, so every cue contains a pose within 5$^\circ$ and 2\,mm of it. Like the simulated users that evaluate interactive segmentation~\cite{sofiiuk2021ritm}, ours is optimistic, because a real user can be wrong. For the soft form only, we also test a wrong user, whose error is 1.5 times the size of the region, so the reference pose lies outside every cue. Under the hard form this would be a certain miss for three of the four wrong cues. The fourth, the orientation cue of 10$^\circ$, would still contain a pose within 5$^\circ$ and 2\,mm of the reference pose on 152 of the 300 sections.  

\noindent\textbf{Setup and rule.} Each run repeats a cue-free run of Table~\ref{tab:main} with only the cue added, so every comparison is paired. The runs are cross-modal, on the 300 test sections and the synthesized MRI, with the two-stain dual encoder under the clean protocol. Before any section was solved under a cue, we pre-registered the rule and how the upper bound of each cue size is computed from the cue-free runs alone. A cue \emph{helps} if it adds at least 5 points of joint accuracy with block-level $p<0.05$ on both seeds. It has \emph{no effect} if it changes joint accuracy by less than 5 points on both, and is \emph{unresolved} otherwise (Appendix~\ref{app:rules}). By rejecting a returned pose, a hard cue can fix only the failures whose returned pose lies outside its region, and their number is the upper bound.  In single-section mode they are 18 / 25 of the 88 / 82 failures at 20\,mm and 61 / 56 at 10\,mm with 10$^\circ$ (seed 0 / seed 1). In block mode they are 5 / 25 of the 36 / 44 failures at 10\,mm with 10$^\circ$ and 3 / 23 at 10$^\circ$. The seeds differ there because seed 1 fails grossly on two blocks of one hemisphere, 13 sections that every cue can reject. In block mode on seed 0, no cue has more than 12 failures outside it (the plane-only cue has the most), and no position cue has more than 2. In block mode, no cue size can therefore reach 5 points on both seeds by rejecting returned poses alone.   

\noindent\textbf{No cue size passes the rule} (Table~\ref{tab:cue}). In single-section mode seven of the nine cue sizes change joint accuracy by at most 3.3 points. The two tightest orientation cues reach 0.750 / 0.747 at 10$^\circ$ (4.3 / 2.0 points, block-level $p = $ 0.085 / 0.13) and 0.763 / 0.753 at 10\,mm with 10$^\circ$ ($p = $ 0.003 / 0.28), both from 0.707 / 0.727. At 10\,mm with 10$^\circ$ the change is 5.7 / 2.7 of the 19 to 20 points that the upper bound allows, significant on seed 0 alone, so under the two-seed rule it is \emph{unresolved}. The orientation cue of 10$^\circ$ alone stays below 5 points on both seeds, so under the rule it has \emph{no effect}, as it does with an LFB-only dual encoder (Appendix~\ref{app:rules}). In block mode no hard cue size changes accuracy by more than 7.3 points, and the seeds disagree, $-0.3$ to $+3.0$ points on seed 0 and $+0.3$ to $+7.3$ on seed 1. The best hard cells are 0.910 / 0.927 at 10\,mm with 10$^\circ$ and 0.907 / 0.927 at 10$^\circ$, both from 0.880 / 0.853 ($p = $ 0.15 / 0.027 and 0.19 / 0.016). They are the two hard-cue cells of block mode with a significant gain, each on seed 1 alone, so under the two-seed rule both are \emph{unresolved}.  

\noindent\textbf{The cue removes gross misses, not imprecision.} Gross misses are sections beyond 30$^\circ$ or 20\,mm. With the 10\,mm and 10$^\circ$ cue, those in single-section mode fall from 21 / 24 of 300 to 0 / 0, and accuracy within 10$^\circ$ and 5\,mm rises from 0.870 / 0.877 to 0.943 / 0.953. 
At the 5$^\circ$ and 2\,mm criterion, accuracy changes less, because most gross misses become near misses. Of the single-section failures that a cue can reject, it fixes at most 30\% at every cue size, which is the pre-registered sign that the score, not the search, is the limiting factor for these sections.  The sections that still fail lie inside the cued region, which also contains a pose within 5$^\circ$ and 2\,mm of the reference pose. The absence of an initial pose is therefore not the limiting factor.  

\noindent\textbf{Block mode.} The cue changes which per-section results vote for the block's rotation, so between 2 and 25 sections are fixed and between 0 and 14 broken per cue size, most of them inside the cued region. On seed 1 every cue size corrects the two gross block failures, so the 13 gross misses fall to 0, and 0 to 10 of those sections become hits. Seed 0 has no gross block failure, so there the cue mostly redistributes near misses, and two cue sizes even create gross misses, 4 with the orientation cue of 20$^\circ$ and 2 with the plane-only cue.   

\noindent\textbf{Soft form.} Because a correct pose just outside the region is no longer rejected, in single-section mode the soft cue breaks at most 5 correct solutions per cue size, whereas the hard cue breaks up to 8. The soft cue also fixes fewer: at 10$^\circ$ it fixes 7 / 8 sections and breaks 2 / 2, whereas the hard cue fixes 16 / 9 and breaks 3 / 3. Its gain is therefore no larger, 0.723 / 0.747 at 10$^\circ$ ($p = $ 0.37 / 0.10) and 0.723 / 0.750 at 10\,mm with 10$^\circ$, and no soft single-section cell of the right user is significant on both seeds. In block mode the soft cue at 10\,mm with 10$^\circ$ reaches 0.927 / 0.927 ($p = $ 0.22 / 0.021), again on seed 1 alone. At 10$^\circ$ in block mode it gains exactly 5 points on seed 1, 15 of the 300 sections, so the rule gives \emph{unresolved}. Under the pre-registered comparison of the two forms, the soft cue is at least as good as the hard cue at 40\,mm, 20\,mm, and 10\,mm in single-section mode and at 10\,mm with 10$^\circ$ in block mode.  

\noindent\textbf{A wrong user.} With a wrong user the soft cue costs at most 1.0 points (Table~\ref{tab:cue_wrong}), and at the two tightest cues it even gains 4.3 to 5.3 points, where the right user's hard cue gains 2.0 to 5.7.  At this weight the soft cue is a weak prior that mainly determines where the global search looks, three region sizes wide. It leaves the rest to the data, so it is robust to a wrong user and, with the right user, fixes fewer sections than the hard cue.  

\noindent\textbf{Cost.} In the hard form, an orientation cue cuts the single-section search time from about 25\,s to about 15\,s. The 10\,mm with 10$^\circ$ cue reduces it to 2\,s in the hard form and to 12 to 13\,s in the soft form, whose global search covers a window three region sizes wide. A position cue alone saves no time, since the FFT scores all translations at once. 

\noindent\textbf{Not covered.} We did not test a real user or other kinds of cue such as landmarks, and the baselines were not given the cue. 

%% file: sections/tab_cue_rows.tex
No cue & 0.707/0.727 & 0.707/0.727 & 21/24 & 0.870/0.877 & 0.880/0.853 & 0.880/0.853\\
\midrule
Position, 40\,mm & 0.690/0.733 & 0.707/0.737 & 20/14 & 0.870/0.900 & 0.887/0.887 & --\\
Position, 20\,mm & 0.693/0.733 & 0.693/0.737 & 16/13 & 0.873/0.897 & 0.877/0.883 & --\\
Position, 10\,mm & 0.683/0.727 & 0.697/0.737 & 12/11 & 0.867/0.900 & 0.903/0.857 & --\\
Orientation, 45$^\circ$ & 0.693/0.737 & 0.697/0.727 & 17/17 & 0.867/0.897 & 0.887/0.860 & --\\
Orientation, 20$^\circ$ & 0.727/0.750 & 0.707/0.740 & 10/11 & 0.897/0.903 & 0.890/0.890 & --\\
Orientation, 10$^\circ$ & 0.750/0.747 & 0.723/0.747 & 9/12 & 0.930/0.933 & 0.907/0.927 & 0.890/0.903\\
Plane only, 20$^\circ$ & 0.733/0.753 & 0.700/0.740 & 13/15 & 0.897/0.900 & 0.903/0.887 & --\\
20\,mm + 20$^\circ$ & 0.710/0.760 & 0.707/0.747 & 2/2 & 0.900/0.913 & 0.887/0.897 & --\\
10\,mm + 10$^\circ$ & 0.763/0.753 & 0.723/0.750 & 0/0 & 0.943/0.953 & 0.910/0.927 & 0.927/0.927\\
\bottomrule

%% file: sections/tab_cue_wrong_rows.tex
Position, 20\,mm & 0.693/0.737 & 0.697/0.730 & 2/6 & 5/5\\
Plane only, 20$^\circ$ & 0.700/0.740 & 0.733/0.733 & 16/15 & 8/13\\
Orientation, 10$^\circ$ & 0.723/0.747 & 0.757/0.773 & 22/20 & 7/6\\
10\,mm + 10$^\circ$ & 0.723/0.750 & 0.750/0.780 & 21/21 & 8/5\\
\bottomrule

%% file: sections/supp_rules.tex
\section{Pre-registered decision rules}\label{app:rules}
Each paragraph below gives an experiment and the decision rule written down for it before it ran, with its thresholds and the branch its plan called informative. It then states the outcome that rule gives on the numbers this paper prints (Appendix~\ref{app:protocol}). A plan calls a branch likely if it expected that branch, and informative if that branch would change a claim or a decision. Some plans also give a prior, the probability they assigned to a branch in advance. Where this paper reports the numbers behind an outcome, the paragraph gives them or names the appendix that does. Several paragraphs also name a departure from the plan, or a rule that was written or changed after a first measurement. Each rule is a dated plan or log entry, not filed in a public registry, and \emph{pre-registered} has this meaning throughout the paper. The rules will be released with the code (Appendix~\ref{app:release}).   

\noindent\textbf{The first runs on the test set.} The test blocks were fixed in advance by excluding the 68 development blocks. The test set was first scored after the method was frozen as it then stood, with an earlier LFB-only dual encoder and one seed. Pilot runs of the refinement had already used all 15 test blocks of hemisphere 5, and the two-stain dual encoder came after these first runs (Appendix~\ref{app:data}). The plan required that joint accuracy after the refinement not fall more than two block-bootstrap standard errors below the development value. It also required that the refinement's gain on the test blocks be positive with block-level $p<0.05$. Two rules written down before the runs replaced that design, so neither requirement was applied. In single-section mode, the method had scored six tiles of a section until then. Scoring all tiles, adopted on the development sections before these runs, was required to add at least 20 points of rotation accuracy within 5$^\circ$, with block-level $p<0.05$, and to reach 0.60 in that accuracy. In block mode, on the 300 test sections, joint accuracy was required to reach 0.80, and rotation accuracy within 5$^\circ$ was required to reach 0.87. Each plan called a clear miss informative: a gain below 10 points for the first rule and joint accuracy below 0.75 for the second. Applied to these runs, both rules give their likely branch, \emph{confirmed}. This paper does not print the accuracies of these first runs. The test rows of Table~\ref{tab:main} come from later runs (Appendix~\ref{app:loso}). 

\noindent\textbf{The two-stain dual encoder.} Whether one dual encoder trained on both stains would replace the LFB-only dual encoder was decided on the test set in block mode with two seeds. The comparison was paired per section, with an exact sign test over the 68 blocks. The rule adopts it if LFB joint accuracy stays within 2 points of the LFB-only dual encoder on both seeds and H\&E reaches at least 0.60 on both. It adopts it \emph{with a stated cost} if LFB falls 2 to 8 points below on either seed. It leaves the decision open if LFB falls more than 8 points below on either seed or H\&E below 0.50. The plan called the two outer branches informative. The two-stain dual encoder's expected cost and gain had been measured on the test set before the rule was written down (Section~\ref{sec:setup}). On LFB, joint accuracy falls from 0.910 / 0.927 with the LFB-only dual encoder to 0.880 / 0.853, a cost of 3.0 / 7.3 points (exact block sign test $p = 0.55$ / $0.016$). H\&E reaches 0.764 / 0.811 (Appendix~\ref{app:results}). Applied to these accuracies, the rule gives adoption \emph{with a stated cost}, the middle branch and the likely one. 

\noindent\textbf{MIND-SSC in block mode.} The training-free MIND-SSC similarity was run through the whole of block mode with only the pose score swapped, on the 300 test sections, paired per section with our block mode. With $J$ its joint accuracy after the refinement, the rule classifies $J \ge 0.60$ as informative: the blocks would then carry much of the accuracy. The claim for the learned descriptor would then be narrowed. The rule classifies $J<0.60$ with ours ahead at block-level $p<0.05$ on both seeds as the likely branch, and ours not ahead at that level on at least one seed as no measurable advantage. Masked NCC is reported beside it and changes no branch. MIND-SSC reaches 0.667, and our block mode with the two-stain dual encoder remains 20 points higher ($p<0.01$ on both seeds, Section~\ref{sec:baselines}). Applied to these accuracies, the rule gives the informative branch. 

\noindent\textbf{Baselines at section level.} Each baseline of the first baseline plan was scored at section level, each section alone, cross-modally and within one modality. It ran first on the 150 development sections and then on the 300 test sections, under one rule per row. For masked NCC and MIND-SSC, the sentence that the similarity and not the search is the problem holds if a row reaches 0.70 within one modality and stays at or below 0.20 cross-modally. The plan called informative a cross-modal cell of 0.30 or more, or a same-modality cell below 0.50. Applied to the test-set accuracies, the sentence holds for masked NCC, at 0.760 within one modality and 0.187 cross-modally. MIND-SSC falls in the informative branch on both sets, at 0.377 cross-modally on the test set, so the paper states that sentence for masked NCC alone. For direct pose regression the plan called both same-modality branches informative. A same-modality cell of at least 0.30 would blame the modality, and one below 0.05 means that the regressor fails, not the modality. It scores 0.000 cross-modally and within one modality (Table~\ref{tab:baselines_untouched} has no row for it, see \emph{Direct pose regression} below). For anatomix, the cross-modal cell was expected below 0.10 on both seeds and would have been informative at 0.30 or more on both seeds, and the result falls in neither band. The same holds for LoFTR-S2V trained within one modality, whose rule requires at least 0.30 or below 0.05 on both seeds, and both rows are reported as numbers alone. Our single-section rows meet their rules. On the development set, the modality costs single-section mode 21 points, 0.953 / 0.960 within one modality against 0.733 / 0.753 cross-modally with the LFB-only dual encoder of Table~\ref{tab:clean}. This is considerably more than the 10 points below which its rule would have been informative. With the two-stain dual encoder the cost is 20 points (Section~\ref{sec:baselines}). On the test set our single-section mode stays above every baseline's cross-modal cell, the likely branch. A row transfers from the development set if it moves by at most 10 points on both seeds. MIND-SSC falls from 0.467 to 0.377. Applied to these accuracies, the rule gives a transfer, as the drop is 9 points. The paired test of the plan cannot run, because the two sets share no block. Instead, an unpaired $z$ test on block-bootstrap standard errors finds the drop not significant ($z=-1.29$, Appendix~\ref{app:baselines}). 

\noindent\textbf{The translator baseline.} A further baseline, whose rule was fixed before it ran, scores masked NCC on the MRI view that our LFB translator synthesizes from each tile. It runs in our global search, cross-modally on the 300 test sections. Two translators trained with different seeds give the two values, each paired with our dual encoder of the same seed, and the test is an exact sign test over the 68 blocks. The plan called informative a row within 10 points of our single-section mode on at least one value, and a lead of ours that misses $p<0.05$ on at least one value. Either would have stopped the analysis before any sentence was written. It also called informative a row at least 5 points above MIND-SSC and a loss of at least 5 points against masked NCC. The row counts as below MIND-SSC at 10 points or more under it with $p<0.05$ on both values. Against masked NCC, a change counts as a gain at 5 points or more with $p<0.05$ on both values, and as a loss at 5 points or more under it with $p<0.05$ on either. Anything else counts as \emph{no resolved change}. The row reaches 0.223 / 0.250, 15.3 / 12.7 points below MIND-SSC ($p = 0.0026$ / $0.0015$), and changes by $+3.7$ / $+6.3$ points against masked NCC ($p = 0.41$ / $0.44$). Applied to these accuracies, the rule counts it as below MIND-SSC with \emph{no resolved change} against masked NCC, and our lead holds ($p<10^{-12}$). After reading the single-section result, we also ran the row in block mode, under a rule fixed before it ran (the same two comparators and test, with no point margin against them). There it reaches 0.353 / 0.377, below MIND-SSC's 0.667 ($p = 0.0013$ / $0.00054$), and changes by $+9.3$ / $+11.7$ points against masked NCC's 0.260 ($p = 0.093$ / $0.052$). Applied to these accuracies, the rule counts it as below MIND-SSC in block mode too, with \emph{no resolved change} against masked NCC (Appendix~\ref{app:baselines}). 

\noindent\textbf{EUReg.} EUReg was scored on the 300 test sections both without initialization and initialized as in its prostate benchmark. Its cross-modal models were trained on both stains (Appendix~\ref{app:baselines}). Our single-section mode was required to outperform the row without initialization cross-modally by over 10 points with $p<0.05$ on both seeds, or the analysis would stop before any sentence was written. That row's same-modality cell under 0.05 on both seeds, the likely branch, would attribute the failure to the full range of poses of our task. A same-modality cell of at least 0.30 on both seeds, with a cross-modal cell under 0.05, would attribute it to the modality. Attributing the failure to the range required a check, the initialized row at 0.50 or more within one modality on both seeds. The plan called informative a smaller lead, a failed check, and any other outcome. Without initialization EUReg places no section, and initialized it places 0.057 / 0.060 within one modality (Appendix~\ref{app:baselines}). Applied to these accuracies, the rule confirms our lead, the likely branch, but the check fails, so the paper gives no reason why the row without initialization fails. The threshold of the check was fixed before EUReg's accuracy on its own benchmark had been computed at the joint criterion. There EUReg places 0.52 / 0.44 of its test frames within 5$^\circ$ and 2\,mm, with the translation measured at the frame's center (Appendix~\ref{app:euregbench}), so the threshold sat at the level EUReg reaches on its own data. Table~\ref{tab:baselines_untouched} prints the initialized row by our choice, made after this result was known, which the plan allowed. 

\noindent\textbf{Direct pose regression.} Table~\ref{tab:baselines_untouched} has no row for direct pose regression. Its same-modality checkpoints were already on record as fitting 0.000 to 0.006 of their own training tiles (Appendix~\ref{app:baselines}), so a row with its same-modality cell would stay out of the table in every branch. A rule then asked whether the 10 checkpoints behind its cross-modal cell, trained on both stains, fit their training tiles. Every one below 0.05, the likely branch, would let one sentence say so for both columns. Every one at 0.80 or more, the informative branch, would have given the table a row with its cross-modal cell alone. Any other fit would have kept the row out, with the fit stated in numbers. The rule's text was written before the fit was measured. By mistake, the measurement ran 29 minutes before the rule was frozen, and its result was first opened after the freeze. They fit 0.000 to 0.002, below 0.05, the likely branch, so the paper says that the regressor does not fit its own training tiles. 

\noindent\textbf{MIND-SSC on our shortlist.} After the main results, a rule was fixed for an analysis of the MIND-SSC row before any of its code existed. Exact MIND-SSC, computed on the MRI plane rendered at each pose (Appendix~\ref{app:baselines}), is given the 256 poses that our learned score shortlists for each of the 300 test sections. In the same task, our finishing stage runs on the same poses. The deficit splits into the share $d$ of our finishing stage and the share $g$ of our shortlist, evaluated per seed with an exact sign test over the 68 blocks. The likely branch was $d$ of 5 points or more with $p<0.05$ on both seeds. A $d$ within 5 points on both seeds, which would have attributed most of the deficit to our shortlist, and a reversed sign, to be reported but not adopted into our method, were called informative. The likely branch for $g$ was likewise 5 points or more with $p<0.05$ on both seeds. In contrast, a $g$ within 5 points on both seeds, or 5 points or more below zero with $p<0.05$ on both, was called informative. Also called informative was a $d$ under 5 points on both seeds at 20$^\circ$ and 20\,mm, which would have meant that MIND-SSC finds the correct neighborhood among our candidates but not the pose. Either share is compared with Table~\ref{tab:baselines_untouched} only if our finishing stage, run again, reproduces at least 297 of its 300 outcomes on both seeds, which the plan expected, and a failure was called informative. It reproduced all 300 on both seeds. The share $d$ is 17.7 / 19.0 points and $g$ is 15.3 / 16.0, both with $p<10^{-3}$ (Table~\ref{tab:mindlsl}), and at 20$^\circ$ and 20\,mm $d$ is 13.3 / 10.3 points. Applied to these accuracies, the rule gives both likely branches and no informative one. 

\noindent\textbf{Mutual information.} In the plan of the paragraph above, mutual information computed by FFT was to be the pose score of a row of Table~\ref{tab:baselines_untouched}, from 8 gray levels on each side. The levels of a tile are fitted by k-means to its gray image. Our single-section mode was required to outperform it cross-modally by over 10 points with $p<0.05$ on both seeds, or the analysis would stop before any sentence was written. The likely branch was a row at least 5 points below MIND-SSC with $p<0.05$. A smaller lead, a row at least 5 points above MIND-SSC, a loss against masked NCC, and a row that fails even within one modality were called informative. Before it ran on the test sections, the row was required to place at least 15 of 30 development sections within one modality, with one retry at 16 levels. Before any section was scored, the initialization of that k-means was changed, because the planned one failed on the MRI volumes, which are mostly empty. A check of the code on a phantom then failed at the first stage of our search. Before any number on real sections existed, we attributed this failure to mutual information in our search (Appendix~\ref{app:baselines}), since every exactness check of the code had passed. Only then, before the development sections were scored, did the plan assign a failure at both level counts the same prior as a pass and call a pass informative. The row placed 3 of 30 at 8 levels and 3 at 16, so it was not run on the test sections and none of its rules was applied (Appendix~\ref{app:baselines}). 

\noindent\textbf{Baselines trained on both stains.} Direct pose regression, anatomix, and LoFTR-S2V, the learned baselines of the first baseline plan, were retrained on both stains, as our two-stain dual encoder is. Each was compared cross-modally with its own LFB-only version on the 300 LFB test sections, per seed, with an exact sign test over the 68 blocks. The rule classifies a gain of at least 5 points on both seeds with $p<0.05$ on both as \emph{improved}. It classifies a loss of at least 5 points on either seed with $p<0.05$ on that seed as \emph{hurt}. Anything else is \emph{no resolved change}. Any sentence about the change must state that the two-stain version trained on twice as much data. If a two-stain baseline came within 10 points of our two-stain single-section mode on either seed, the analysis was to stop before any sentence was written, and none came that close. The plan called \emph{improved} informative, and stated that its test has no power for pose regression. For anatomix the cell moved from 0.290 / 0.193 to 0.357 / 0.210 ($p = 0.043$ / 0.70, a gain on seed 0 alone). For LoFTR-S2V it moved from 0.013 / 0.007 to 0.017 / 0.017, and pose regression stayed at 0.000 (Appendix~\ref{app:baselines}). Applied to these accuracies, the rule gives \emph{no resolved change} for every row, as no row changed significantly on both seeds and none fell. 

\noindent\textbf{H\&E as a second stain.} An earlier LFB-only dual encoder was first run unchanged on H\&E, on the 148 development sections whose H\&E image passed the plan's two image checks. The rule had bands at 0.60 and 0.20 of joint accuracy after the refinement. Both seeds were required to fall in the same band, or the lower one applied. At least 0.60, which would have allowed the LFB-only dual encoder to be claimed for both stains, was the informative band. Applied to these runs, the rule gives the lowest band on both seeds, below 0.20, the most likely one. An earlier H\&E-only dual encoder therefore followed, trained with only the stain changed. The recipe \emph{transfers} if this dual encoder stays within 10 points of the LFB-only one on both seeds. It is \emph{stain-specific}, the informative branch, if this dual encoder falls more than 10 points below with block-level $p<0.05$ on both seeds. The rule gives \emph{stain-specific}, that is, the same recipe is less accurate on H\&E. Table~\ref{tab:he} reports the same comparison with one dual encoder per stain, and no rule covers it. There the H\&E-only dual encoder stays 15 to 32 points below the LFB-only one with block-level $p<0.011$ in every row (Appendix~\ref{app:results}). 

\noindent\textbf{The acquired MRI.} Every model was retrained on the acquired 0.4\,mm MRI under the clean protocol with the LFB-only dual encoder. The acquired MRI was compared in block mode with the synthesized MRI on the 150 development sections, per seed. Hemisphere 5 was scored first as a go/no-go check, passed if the acquired MRI reached at least half the accuracy on the synthesized MRI on both seeds. The rule evaluates the difference in joint accuracy after the refinement, acquired minus synthesized, paired per section. It is \emph{equivalent} if within 5 points on both seeds, and \emph{worse}, the informative branch, if at least 5 points lower on both seeds with block-level $p<0.05$ on at least one. It is \emph{better} if at least 5 points higher on both, and \emph{mixed} otherwise. The check gives a pass, at 0.767 / 0.800 against 0.867 / 0.900. Block mode reaches 0.920 / 0.940 on the acquired MRI against 0.953 / 0.960 on the synthesized MRI, a difference of $-3.3$ / $-2.0$ points (block-level $p = 0.63$ / $0.75$). Applied to these accuracies, the rule gives \emph{equivalent} (Appendix~\ref{app:realmri}). No rule covers the rows of Table~\ref{tab:main} on the acquired MRI, which use the two-stain dual encoder and are higher on the LFB test sections. 

\noindent\textbf{The qualitative figure.} This rule selects rows and produces no number, so it has no informative branch. It ranks the 300 test sections by the error $e$ of our block mode on seed 0. Here $e$ is the larger of the rotation error divided by 5$^\circ$ and the translation error divided by 2\,mm, so that a section is correct when $e<1$. The rows are the sections at the 25th, 50th, 75th, and 90th percentiles of that order. Two more rows are the section with the largest rotation error and the section with the largest translation error among sections with a correct rotation. No two rows come from the same block. A row whose block is already taken moves to the next section in its own order, a percentile row toward the larger error, and nothing else selects a row. Figure~\ref{fig:qualitative} shows the rows this rule selects. Figure~\ref{fig:qualitative_s1} applies the same rule to seed 1, and Figure~\ref{fig:failures} shows every failed block of seed 0 by its section with the largest $e$. The rule was defined on the LFB-only runs and applied unchanged to the two-stain runs of Table~\ref{tab:main}. 

\noindent\textbf{Missing content.} Tiles, tissue (trimmed from its outer contour), or sections were removed step by step, and rotation accuracy within 5$^\circ$ before the refinement was measured per block on the 150 development sections. The two sides of the comparison are block mode and each section solved alone (the Single columns of Table~\ref{tab:content}). A block's half-accuracy step is the step at which its accuracy falls to half its value with all content, interpolated linearly between steps. The rule tests that step in block mode minus that of the sections solved alone, with the same sign on both seeds. It uses the paired sign-flip test over the 68 blocks, evaluated exactly, at $\alpha = 0.025$. At least 0.25 steps with $p<0.025$ is classified as the block \emph{compensating} for missing content. A difference within 0.10 steps whose 95\% interval lies inside $\pm 0.25$ is classified as \emph{separable}. At most $-0.25$ steps with $p<0.025$ is classified as the block needing content too, and anything else as inconclusive. The plan called the first three branches informative and the mere fall of accuracy uninformative. The paper reports runs with the two-stain dual encoder, and the rule applies to them unchanged. Applied to the accuracies of Table~\ref{tab:content}, the rule gives \emph{compensating} on both seeds for tiles and for tissue. The half-accuracy step is defined only for a block with a section within 5$^\circ$ at all content, both in block mode and with each section solved alone. The test runs over these blocks, 58 / 57 of the 68, for tiles and for tissue, and it draws random sign flips where the rule asks for all of them. A constant lead alone would produce 15 to 74\% of that effect, and the remainder is significant on one seed only, for tiles and for tissue, so the paper considers the result suggestive (Section~\ref{sec:serial}, Appendix~\ref{app:results}). The removal of sections has no branch, because a section solved alone cannot depend on the number of sections. 

\noindent\textbf{Four sections per block.} Each of the 150 development sections was solved in its own block of $K$ sections, itself and $K{-}1$ evenly spaced unscored sections of its block, and then refined. The rule evaluates joint accuracy at $K=4$ per seed. Both seeds at 0.70 or above support the claim that four sections suffice, both below 0.70 require the paper to state that at least eight serial sections are needed, and split seeds are inconclusive. The rule reports $K=8$ descriptively, and the claim must use the words \emph{evenly spaced}. The plan called a result below 0.70 informative. With the two-stain dual encoder and the same threshold, $K=4$ reaches 0.853 / 0.840, about 6 points below the full block (Appendix~\ref{app:results}). Applied to these accuracies, the rule gives the likely branch, that four evenly spaced sections suffice. The plan expected the line fit to be inactive in blocks this small. At $K=8$, the line fit replaced one of the 300 scored poses (150 sections, two seeds), and at $K=4$ none. The conclusion rests on $K=4$, so it holds with that pose counted as a miss (Section~\ref{sec:serial}). 

\noindent\textbf{The optional human cue.} A simulated user bounds where a section lies, how it is oriented, or both. The unchanged search either stays inside that region (the hard form) or incurs a penalty for leaving it (the soft form). Each cue size runs in single-section mode and in block mode on the 300 test sections, paired with the run without a cue. An upper bound on what a cue can fix is computed from the runs without a cue alone. A cue size \emph{helps} if it adds at least 5 points of joint accuracy with block-level $p<0.05$ on both seeds. It has \emph{no effect} if the change is below 5 points on both seeds, and it is \emph{unresolved} otherwise. Every cue size is reported, and none is chosen as the best. A second rule evaluates, in single-section mode, the share of the fixable failures that a cue fixes. A share of at least 75\% on both seeds indicates that the search limits them, and at most 40\% on both indicates that the score does. Two rules were added in the first, LFB-only experiment, after its hard single-section runs and before any soft run. They compare the soft form with the hard one and test a wrong user, whose error is 1.5 times the region. The soft form is \emph{at least as good} as the hard one if it gains no less and breaks no more on both seeds. The soft cue is \emph{robust} to that user if it costs at most 5 points on both seeds, and its penalty weight was fixed at 0.031 beforehand. In block mode only the two tightest orientation cues run in the soft form. The whole experiment was then rerun with the two-stain dual encoder under the same rules, recorded before any run of it. Its plan called informative a block-mode cue size that helps on both seeds, a single-section outcome that differs from the LFB-only one, and a share above 40\%. Applied to the accuracies of Appendix~\ref{app:cue}, the rule finds no cue size that helps. In single-section mode the hard cue of 10\,mm with 10$^\circ$ is \emph{unresolved}, significant on seed 0 alone, and the orientation cue of 10$^\circ$ has \emph{no effect}, as with the LFB-only dual encoder. No single-section outcome differs from that of the LFB-only dual encoder. In block mode the hard cues of 10$^\circ$ and of 10\,mm with 10$^\circ$ gain significantly on seed 1 alone, so both are \emph{unresolved}. On seed 1 every cue size corrects the two gross block failures, so their 13 gross misses fall to 0, and 0 to 10 of these sections become hits. The soft cue is \emph{unresolved} at the same two sizes in block mode, at 10$^\circ$ with a gain of exactly 5 points on seed 1. In single-section mode a cue fixes at most 30\% of the failures it can reject, at every cue size, which indicates that the score limits these sections. In single-section mode the soft cue breaks at most 5 correct solutions per cue size, whereas the hard cue breaks up to 8. The soft cue also fixes fewer. The soft form is \emph{at least as good} at 40, 20, and 10\,mm in single-section mode and at 10\,mm with 10$^\circ$ in block mode. The wrong user's soft cue costs at most 1.0 points, so it is \emph{robust}. 

\noindent\textbf{Mined against random negatives.} The two-stain dual encoder of every fold was retrained with the 32 mined wrong poses of each tile replaced by 32 random poses in the tissue. Like the mined ones, they lie at least 15$^\circ$ or 15\,mm from the reference pose. The retrained dual encoder was evaluated in single-section mode on the 150 development sections, per seed, as the change in joint accuracy, with an exact sign test over the blocks. The mined poses count as \emph{better} if they gain at least 5 points on both seeds with $p<0.05$ on at least one, and the change as \emph{no change} if below 5 points on both. Random poses count as \emph{better} if they gain at least 5 points on both with $p<0.05$ on both, and the outcome is \emph{mixed} otherwise. The bar is higher for the random poses, because a reason to drop a part of our recipe must be stronger than a reason to keep it. The same bands apply to 8 random poses against the 8 the translator mines, and to the first mining round against the second. The plan called \emph{better} for the mined poses likely and \emph{no change} informative, and named no likely or informative branch for the two other comparisons. Random poses cost 14.7 / 14.0 points, and training on the first round alone costs 16.0 / 14.0 points, each with block-level $p<0.05$ on both seeds. Applied to these accuracies, the rule gives \emph{better} for the mined poses, the likely branch, and \emph{better} for the second round. With 8 poses per tile, random poses perform as well as those the translator mines ($-1.3$ / $+2.0$ points for the mined ones), which the rule gives as \emph{no change} (Appendix~\ref{app:mining}). 

\noindent\textbf{Training without mined poses.} Planned after the result of the test above was known, two further retraining runs of the two-stain dual encoder of every fold remove the mined poses. One keeps our loss with only the other tiles of the batch as negatives (CoMIR-style), about 15 per tile against 527 for ours. The other trains its dot product to imitate LC2 (DISA-style) at the reference pose and 32 random poses. Each of the 10 dual encoders of each retraining run was first required to fit its own objective, and all did. One fit check, the spread of the in-batch dual encoders' tile-mean descriptors (Appendix~\ref{app:mining}), was measured on all 375 validation tiles of each fold, since no fold contains the planned 512. Each run was evaluated in single-section mode on the 150 development sections, per seed, as the change in joint accuracy, with an exact sign test over the blocks. Six rules apply, with no family-wise claim. Against ours and against 32 random poses, ours or the random poses count as \emph{better} if they gain at least 5 points on both seeds with $p<0.05$ on at least one, the retrained dual encoder only with $p<0.05$ on both. This is the higher bar of the paragraph above. Against MIND-SSC a dual encoder counts as \emph{above} or \emph{below} only with at least 5 points and $p<0.05$ on both seeds, and as not resolved otherwise. The plan called \emph{better} for ours likely, and \emph{no change} or a better retrained dual encoder informative, for both dual encoders. Against MIND-SSC it called \emph{below} informative for in-batch negatives and \emph{above} for imitation. Against 32 random poses it called a better in-batch dual encoder informative, and for imitation \emph{no change} or a better imitation dual encoder. On the accuracies of Appendix~\ref{app:mining}, ours counts as \emph{better} by 16.0~/~14.0 points over in-batch negatives and by 72.0~/~72.7 over imitation. In-batch negatives are not resolved against MIND-SSC, the likely branch. Against 32 random poses, they stay within 5 points with considerably fewer negatives per tile, \emph{no change}, whereas \emph{better} for the random poses was likely. Imitation, trained without mined poses, falls \emph{below} MIND-SSC by 44.0~/~43.3 points, the expected direction. Against imitation, 32 random poses with our loss count as \emph{better} by 57.3~/~58.7 points, the likely branch. 

\noindent\textbf{Tiles versus the whole section.} Encoding the whole section in one pass, each tile taking its descriptors from the single resulting descriptor field, was compared with encoding each tile separately. Both ran in single-section mode on the development sections of Table~\ref{tab:tiling}, 150 with LFB and 147 with H\&E, under the clean protocol with the two-stain dual encoder. Each stain was evaluated separately, for LFB first and H\&E second. The rule names the whole section a candidate for adoption only if it gains at least 3.0 points of rotation accuracy within 5$^\circ$, averaged over both seeds, with block-level $p<0.05$ and both seeds agreeing in sign. A descriptive label states that the loss found with an earlier LFB-only dual encoder replicates if the whole section loses at least 10 points with block-level $p<0.05$ and both seeds negative. The plan called informative a loss smaller than 10 points or any gain. The whole section loses 12.0 / 20.0 points on LFB and 10.9 / 21.1 on H\&E (seed 0 / seed 1), accuracy falls on both seeds, and block-level $p<10^{-4}$ for each stain (Appendix~\ref{app:tiling}). Applied to these accuracies, the rule does not name the whole section a candidate for adoption, and the label gives a loss that replicates, the likely branch on both stains. 

\noindent\textbf{Stage timers.} Every stage of block mode was timed, with timers that change no pose, on ten test blocks, two per hemisphere, chosen by a rule that uses no timing, with both seeds. An analysis is valid only if every timed run reproduces its archived poses. The timed median block is then compared with the 522\,s derived from the archived runs. It agrees if it lies between 430 and 630\,s, and otherwise it replaces that value, with the difference disclosed. The plan also fixed the share of a block that each stage was expected to take, and it called a disagreement, or a stage outside its expected share, informative. Every timed run reproduced its archived poses. The timed median block, 512\,s, lies in that range, so the rule gives agreement, the likely branch, and the 522\,s stays in Table~\ref{tab:runtime}. Every stage took its expected share, except that within the search the first round of local ascent took the longest, a row the plan had marked descriptive (Appendix~\ref{app:runtime}). 

\noindent\textbf{Deformation in the controlled experiments.} ADNI sections rendered in another contrast were warped by elastic or free-form fields of increasing severity and placed by block mode, beside a same-contrast control, with two seeds. The plan first fixed a primary endpoint that it replaced before the runs. That endpoint divides the breaking severity by 0.5\,mm, a constant the plan fixed in advance. The breaking severity is the smallest at which accuracy, there and at the next two severities, cannot be distinguished from an input-free baseline at block-level $p \ge 0.05$. A ratio above 3 is classified as deformation not being the binding constraint, and 1 to 3 as a reason for a deformation-aware step. A ratio below 1 is classified as a fragile method if our method breaks at least one step earlier than two training-free comparators. It is classified as an invalid series of severities if all three break within one step of each other. The plan called all three ranges of the ratio informative and the mere fall of accuracy with severity uninformative. The endpoint that replaced this ratio is the interaction of severity and condition, which the paper describes without a test. The results rest on 4 held-out subjects, and with fewer than 24 subjects the plan's own clause drops the ratio and reports the curve. An earlier LFB-only dual encoder trained on NextBrain scores 0.000 on this contrast pair, below 0.30, where the plan keeps only the same-contrast control. The recipe was retrained on 15 ADNI subjects instead, a departure from the plan, which had excluded this branch. Joint accuracy stays above 0.92 up to 1\,mm, drops at 2\,mm, and falls sharply at 3\,mm (Section~\ref{sec:adni}, Table~\ref{tab:adni}). Accuracy under elastic and free-form warps falls at the same severity. The same-contrast control falls less at 3\,mm. From 0 to 2\,mm it loses less in three of the four pairs of deformation family (elastic or free-form) and seed (Appendix~\ref{app:adni}). 

\noindent\textbf{New ADNI subjects.} The same controlled experiments were repeated on the 12 lowest-numbered of the new subjects whose rendering passed every check, undeformed and at 2\,mm, in both conditions and with both seeds. The rule assigns \emph{confirmed} if every new value lies within two bootstrap standard errors of its development value, in both conditions and at both severities, and a value above development confirms. It assigns a level shift, the informative branch, if an undeformed value falls more than two standard errors below in either condition, and a deformation shift if only a value at 2\,mm does. The second seed must reach the same branch. On the new subjects, cross-contrast accuracy is 0.999 / 0.978 undeformed and 0.911 / 0.907 at 2\,mm, with elastic and free-form warps together (Appendix~\ref{app:adni}). Applied to these accuracies, the rule gives \emph{confirmed} on both seeds. 

\noindent\textbf{More candidates per section.} With an earlier LFB-only dual encoder, the block vote counted 16 candidates per section instead of 4, followed by the block refinement. The rule adopts the change only if joint accuracy rises by at least 3 points with block-level $p<0.05$ and no accuracy falls by more than 2 points. A census is a count over the blocks the vote gets wrong. The plan called the census informative if it finds a candidate within 10$^\circ$ of the reference rotation on most of them, but no cluster of agreeing sections that close. Their sections then do not agree on it. Applied to that run, the rule does not adopt the change, because joint accuracy falls, and the change was discarded (Appendix~\ref{app:negative}). Applied to the six blocks the vote gets wrong without the change, the census gives that informative branch: a candidate lies within 10$^\circ$ on five of them, and a cluster on none. 

\noindent\textbf{Re-ranking the vote.} For an earlier LFB-only dual encoder, the plan was to refine each block's three leading vote candidates and the mirror of the winner, and to keep the candidate with the highest block MIND-SSC score. A mirror is the winner turned by 180$^\circ$ about one of its two in-plane axes. Departing from the plan, the run used six candidates per block: the winner, three runners-up, and both mirrors of the winner. It used the adoption rule of the paragraph above. Its informative branch was the reference rotation missing from the candidates of most wrong blocks, which would make the candidates, not their ranking, the limiting factor. Applied to that run, the rule does not adopt the change, because the gain stays below 3 points, and the census gives that informative branch. The idea was discarded (Appendix~\ref{app:negative}). 

\noindent\textbf{Re-ranking over more candidates.} The same re-ranking was then run over 16 candidates per section, more vote candidates, and the rotation of the best-scoring section. The rule adopts it only if joint accuracy rises by at least 3.0 points with block-level $p<0.05$ and no correct block is broken. The plan called both a gain and no gain informative. An addendum written before the result made the last condition the primary one. Applied to that run, the rule passes, but its comparison is the vote over the same 16 candidates per section without any re-ranking, which cannot show what the added candidates contribute. Against the settings then in use, the combination performs worse, and it was not adopted (Appendix~\ref{app:negative}). Not adopting it although the rule passes is a departure from the plan. 

\noindent\textbf{Re-ranking on held-out sections.} The candidates of that re-ranking were then ranked by their score on held-out sections of the block instead of the sections they had been fitted to. This ranking was compared against the same run with the in-sample choice. The three adoption criteria were those of the paragraph above. The plan called both outcomes informative and fixed a prediction for one block in advance. That run fails all three criteria of the rule, while the prediction holds. A breakdown outside the plan shows that the run also changed how a candidate's score was computed, that this change alone caused the drop in accuracy and the broken block, and that ranking on held-out sections changed no block. Re-ranking was discarded (Appendix~\ref{app:negative}). 

\noindent\textbf{Averaging candidates.} With an earlier LFB-only dual encoder and six tiles per section, single-section mode was changed at the end of its global search. There, the mean of the candidate rotations within 10$^\circ$ of the best one replaced the best one alone. The rule names it a candidate for adoption if rotation accuracy within 5$^\circ$ rises by at least 3.0 points with block-level $p<0.05$ and at least three of the five hemispheres do not lose. It classifies a change below 3.0 points or $p \ge 0.05$ as \emph{null}, and a loss of at least 3.0 points with $p<0.05$ as \emph{harmful}. The plan called a gain informative and the null likely. Applied to that run, the rule gives \emph{null}, and the idea was discarded (Appendix~\ref{app:negative}). 

\noindent\textbf{A third mining round.} For an earlier LFB-only dual encoder, the training negatives were mined a third time with the final dual encoder, and the dual encoder retrained on them was evaluated leave-one-hemisphere-out on tiles. The rule evaluates the median over the five folds of the share of tiles within 5\,mm of their own reference pose (Appendix~\ref{app:protocol}). At least 58\%, 5 points above the second round's 53.0\%, means mining remains beneficial, and 48 to 58\% that it has saturated. Below 48\% the third round is harmful and the second stays final. The plan called both tails informative. In that run the third round lies below 48\%, so the rule gives the harmful branch and the dual encoder keeps two rounds (Appendix~\ref{app:negative}). 

\noindent\textbf{Larger descriptors.} The two encoders of an earlier LFB-only dual encoder were widened from 8 to 16 descriptor channels and trained for 8000 steps instead of 4000. The rule evaluates the median over the folds of the share of tiles within 2\,mm of their own reference pose (Appendix~\ref{app:protocol}). A gain of at least 8 points means capacity was the limiting factor, a change within 8 points means it was not and scaling stops, and a loss of at least 8 points means overfitting. The plan named no informative branch. Departing from the plan, the run measured another statistic, how often each tile's own reference pose ranked first on held-out validation tiles, so the rule itself gives no branch. On that statistic the change lies inside the band, and scaling stopped (Appendix~\ref{app:negative}). 

\noindent\textbf{Native slides.} The test replaces each of 299 LFB test sections by its native slide. It places each slide by one in-plane similarity transform fitted to its released image. This is the best case of a rigid placement. It compares each slide with the released section solved again in the same runs, per mode and seed. This rerun differs from the LFB test row of Table~\ref{tab:main} on some sections (Appendix~\ref{app:native}). The rule reads the loss in joint accuracy. It \emph{holds} if the loss is at most 5 points on both seeds, \emph{breaks} if it is at least 15 points on both seeds, \emph{costs} if it lies between these bars on both seeds, and is \emph{mixed} otherwise. In single-section mode the plan called \emph{holds} the informative outcome and \emph{breaks} the likely one, with a prior of 0.45. In block mode, where the slides keep the release's alignment between them, it called \emph{holds} and \emph{breaks} both informative, with priors of 0.30 and 0.25. A section without an accepted placement is counted both as a miss and with the outcome of its released section, and an outcome stands only if both counts give it. More than 15 such sections give no outcome at all. Further questions, on our preparation of the slides and on a split of the loss into geometry and appearance, use control conditions, and Appendix~\ref{app:native} defines their branches. Since 17 of the 299 placements were not accepted, the rule gives no branch for any question of this test, so this primary reading is \emph{unresolved}. A second reading, labeled ``after the freeze, cap removed'', drops the limit of 15. It was added after the placements were fitted and before any pose was scored. The native slides lose 19.1 / 18.4 points in single-section mode and 15.4 / 17.1 points in block mode (Appendix~\ref{app:native}). These losses take each slide without an accepted placement as the method solved it, and each lies between the two counts (Table~\ref{tab:native_diff}). Applied to these losses, the second reading gives \emph{breaks} in single-section mode. In block mode one count leaves seed 0 at a loss of 14.0 points, so it gives no branch there. Under the second reading, the question on the preparation, our masking and rescaling of the slides, reads \emph{costly} in both modes, so the loss is ``warp plus preparation, not separated''. For the split question, the second reading gives \emph{interaction} in single-section mode and \emph{attributable} in block mode. Because the primary reading gives no branch, no split is quoted. The main text states four of the plan's five consequences, and Appendix~\ref{app:native} states all five. The consequence the main text leaves out is that the test sections were selected on their released images. The H\&E native row of Table~\ref{tab:main} repeats this test under its own pre-registered plan (Appendix~\ref{app:native}). Applied to its losses, that plan's rule gives the total loss no branch in either mode. 

%% file: sections/supp_native.tex
\section{Native slides}\label{app:native}

This appendix reports the pre-registered test of Section~\ref{sec:mainresults} on native LFB slides, and its last paragraph reports the repeat of this test on native H\&E slides. Errors are measured as in Section~\ref{sec:setup}, in every condition against the tissue of the released section (Appendix~\ref{app:protocol}). Pairs are seed~0~/~seed~1. The pre-registered rule gives no verdict for any question of this test, because its checks did not accept 17 of the 299 placements and it allows at most 15. We report this reading first. A second reading, labeled ``after the freeze, cap removed'', was added after the placements were fitted and before any pose was scored, and every sentence that uses it says so. 

\noindent\textbf{Data.} NextBrain's public data record (Appendix~\ref{app:release}) contains the scans of the glass slides from which the released sections were made, and label maps in the space of many LFB slides. We call such a scan a native slide. A native slide is not warped onto the MRI, and NextBrain has neither masked it nor brought it to the pixel size of the released images. The record names the block of every slide but not the slide that a released section shows, so we paired slides and sections with two tools. They agree on the order of the slides in every block and differ only on which surplus slides of some blocks were left out, which concerns one scored section. The test takes the 300 test sections of Section~\ref{sec:setup}, which were selected on their released images. One of them is excluded by name before any fit, because its native file in the record is a byte-identical copy of its neighbor's. The test therefore scores 299 sections in 68 blocks.

\noindent\textbf{Design (pre-registered).} The test asks whether the method of Table~\ref{tab:main} keeps its accuracy when each test section is replaced by its native slide. Nothing is retrained. With the two-stain dual encoders of the clean protocol, the method runs in both modes on both seeds, and only the image of a section changes. Every condition is drawn on the canvas of the released image, so the reference pose of the block and the scoring stay those of the release. The test has four conditions (Table~\ref{tab:native_given}). Condition R is the released section, solved again in the runs of the test. Condition N is the test condition. It is the native slide, which we prepare from the scan alone and place by one in-plane similarity transform. Two controls cross the two ways in which N differs from R. W carries the prepared native pixels onto the released shape with our own estimate of the release's warp. U carries the released pixels to where the rigidly placed native tissue lies. Every rate has 299 sections in its denominator, and a section that a condition cannot solve counts as a miss.

\begin{table*}[t]
\centering\footnotesize\setlength{\tabcolsep}{4pt}
\caption{What each condition of the native-slide test is given. R is the released section, and N is the native slide placed by one in-plane similarity transform fitted to the released image. W carries the native pixels onto the released shape, and U carries the released pixels to where the placed native tissue lies. A fitted item comes from that similarity, and a withheld item is not applied.}
\label{tab:native_given}
\begin{tabular}{@{}lllll@{}}
\toprule
Item & R & W & U & N\\
\midrule
\input{tables/tab_native_given_rows}
\end{tabular}
\end{table*}

\begin{table*}[t]
\centering\footnotesize\setlength{\tabcolsep}{4pt}
\caption{Joint accuracy in the native-slide test on 299 LFB test sections, seed 0 / seed 1. Block mode is given the release's stack alignment in every row. In block mode W and U come from a second run with its own R, which the last row gives. W and U carry only the smooth part of the release's warp. On 22 of the 297 scored sections with a fit, the tiles of W do not match those of the released section. A dash means not run. No cell is bold, because the conditions are not competing methods.}
\label{tab:native}
\begin{tabular}{@{}lcc@{}}
\toprule
Condition & Single section & Block mode\\
\midrule
\input{tables/tab_native_rows}
\end{tabular}
\end{table*}

\begin{table*}[t]
\centering\footnotesize\setlength{\tabcolsep}{4pt}
\caption{Paired differences in joint accuracy in the native-slide test, in points, seed 0 / seed 1, on 299 sections. Block mode is given the release's stack alignment. Each difference is read twice over the 17 sections without an accepted placement, once with them as misses of the first-named condition and once with the outcome of the second-named one. $p$ comes from the paired sign-flip test over the 68 blocks. U minus R and N minus W are the two geometry differences, and W minus R and N minus U are the two appearance differences. W and U carry only the smooth part of the release's warp, so a sharp local deformation stays in W minus R and is missing from U minus R and N minus W. On 22 of the 297 scored sections with a fit, the tiles of W do not match those of the released section. In block mode N minus W and N minus U cross two runs and are computed against the R of each run. An asterisk marks a value that lies outside its two counts.}
\label{tab:native_diff}
\begin{tabular}{@{}lcccc@{}}
\toprule
Difference & Points & 17 as misses & 17 as comparator & $p$\\
\midrule
\input{tables/tab_native_diff_rows}
\end{tabular}
\end{table*}

\begin{table*}[t]
\centering\footnotesize\setlength{\tabcolsep}{4pt}
\caption{The readings of the native-slide test. The first reading is the pre-registered rule, which gives no branch for any question, because its checks did not accept 17 of the 299 placements and it allows at most 15. The second reading is labeled ``after the freeze, cap removed'', and it gives the branch that both counts share. Block mode is given the release's stack alignment.}
\label{tab:native_reads}
\begin{tabular}{@{}llcc@{}}
\toprule
Mode & Question & First reading & Second reading\\
\midrule
\input{tables/tab_native_reads_rows}
\end{tabular}
\end{table*}

\noindent\textbf{Preparation and placement.} The rules of the preparation and their constants come from development blocks and were frozen before any test slide was prepared. We compute a tissue mask from the slide alone, bring the scan to the release's pixel size, set the background to zero, and save the image with the JPEG settings of the released images. We apply no color or intensity mapping. Our mask differs from NextBrain's in known ways, and each such difference counts toward the preparation. We then fit one similarity transform between the prepared slide and its released image over all tissue, by least squares. The similarity has a mirror state, an in-plane angle, a shift, and one isotropic scale. Over the 282 accepted placements the mirror state is set on 243, and the scale has a median of 1.177 (0.986 to 1.489). N therefore starts from the best rigid placement that the released image allows, and it lacks the anisotropic stretch, the shear, and the nonlinear part of the release's warp. What the similarity leaves of our smooth estimate of that warp is the withheld deformation. Its root-mean-square size over the tissue is 0.31 to 1.99\,mm, with a median of 0.63\,mm. This size is not a severity on the axis of Figure~\ref{fig:adni}, where the severity is what remains of a planted field after the best rigid motion.

\noindent\textbf{What each condition is given.} Table~\ref{tab:native_given} lists what each condition takes from the release. Five consequences follow, and they apply to every number of this test. First, the similarity is the least-squares fit to the released image, so N is the best case of a rigid placement. Second, in block mode each native slide takes its placement from its own released section. The slides of a block therefore keep the alignment to each other that NextBrain's registration to the MRI supplied. We call it block mode given the release's stack alignment, and a stack aligned without the MRI is outside this test. Third, the placement gives each slide its in-plane angle, and the released sections carry the same information. Fourth, the 300 test sections were selected on their released images, which had to yield enough tiles. Fifth, this test covers LFB sections, and the last paragraph of this appendix reports its repeat on H\&E sections. A reading added after the test was read, and not part of the plan, is that the search can represent the angle, the shift, and the mirror state of a slide, but no run of this test withheld them. The same reading notes that every slide is shown in the release's image frame, in which the encoders were trained. 

\noindent\textbf{Acceptance of a placement.} A placement is accepted only if it passes five checks, whose thresholds come from development blocks. The paired slide must match its released image better than both neighboring slides, and the matches must cover the tissue in all four quadrants. Few matches may lie at the bound of the search, and the mirror state must be clear. Where the record contains a label map of the slide, its edges must agree with the released labels. No check bounds the size of the deformation that the similarity leaves, because that deformation is what the test studies. The checks accept 282 of the 299 placements. Of the other 17, two have no fit at all, four have no consistent match, and five fail the neighbor check. Three have a label map that does not pair with its released label slice, and one each fails the label distance, the coverage, and the mirror check. On the development sections with a native slide, the same checks had not accepted 4 of 146 placements. A scored section without an accepted placement stays in the test and is shown to the method from its best fit. The two sections without any fit are shown as empty images, so they are forced misses of W, U, and N. N and W each have one more forced miss, because the placed image of one slide yields fewer than three tiles. In block mode, given the release's stack alignment, W, U, and N show a slide that is not scored and has no accepted placement as an empty image.

\noindent\textbf{The pre-registered rule.} The rule compares N with the R of the same run, per mode, with the paired sign-flip test over blocks. Accuracy \emph{holds} if the loss is at most 5 points on both seeds, and it \emph{breaks} if the loss is at least 15 points on both seeds. It \emph{costs} if the loss lies between the two bars on both seeds, and any other pattern is \emph{mixed}. For single-section mode the plan gave \emph{holds}, \emph{costs}, and \emph{breaks} the probabilities 0.15, 0.40, and 0.45. It called \emph{holds} the informative outcome and \emph{breaks} the likely one, so a break says little by itself. For block mode, given the release's stack alignment, it gave 0.30, 0.45, and 0.25 and called both \emph{holds} and \emph{breaks} informative. Each section without an accepted placement is counted twice, once as a miss of N and once with the outcome of R. A branch stands only if both counts give it. With more than 15 such sections the rule gives no branch at all. Two further questions use the controls, each with a prior of 0.5. The preparation is \emph{costly} if W loses more than 5 points against R on either seed, and the loss of N is then read as ``warp plus preparation, not separated''. The split question reads \emph{attributable} if the two geometry differences, U minus R and N minus W, lie within 5 points of each other on both seeds. Only then is the loss split. Otherwise the four differences are printed and no split is quoted. The plan calls this outcome \emph{interaction}.

\noindent\textbf{The readings.} The primary reading gives no verdict. Under the primary reading every question of this test is \emph{unresolved} and has no branch, because the plan's checks did not accept 17 of the 299 placements, more than the 15 that the rule allows. This reading depends on the placements alone. With 17 such sections the two counts differ by at most 5.7 points. The second reading carries the plan's label ``after the freeze, cap removed''. It drops the limit of 15 and keeps the rule that both counts must give the same branch. Applied to the differences of Table~\ref{tab:native_diff}, the second reading is \emph{breaks} in single-section mode. In block mode, given the release's stack alignment, one count leaves seed 0 at a loss of 42 of the 299 sections, which is 14.0 points, while a break requires a loss of 45 sections. The second reading therefore gives no branch in block mode. Its preparation question reads \emph{costly} in both modes. Under the second reading the split question reads \emph{interaction} in single-section mode and \emph{attributable} in block mode. Because the primary reading gives no branch, the plan prints the four differences and quotes no split, and we quote none (Table~\ref{tab:native_reads}).

\noindent\textbf{Result.} Table~\ref{tab:native} gives the joint accuracy of every condition, and Table~\ref{tab:native_diff} gives the paired differences. N reaches 0.512 / 0.542 in single-section mode against 0.702 / 0.726 for R, and it loses 19.1 / 18.4 points (block-level $p<10^{-4}$ on both seeds). In block mode, given the release's stack alignment, N reaches 0.712 / 0.686 against 0.866 / 0.856, and it loses 15.4 / 17.1 points ($p = 0.020$ / $0.0049$). Each of these losses is significant on both seeds. R is a rerun of the test sections in this test's runs, not the LFB test row of Table~\ref{tab:main}. Its joint outcome equals that of the runs behind that row on 297~/~296 of the 299 sections in single-section mode and on 289~/~296 in block mode. These reruns used A100 and H100 GPUs, between which the search is not reproducible bit for bit. Every loss is paired with this R, never with that row or the row \emph{Ours} of Table~\ref{tab:baselines_untouched}. Three sections of N are forced misses in every run. In block mode R is correct on two of these three, so two of the lost block-mode sections come from the placement and the preparation, not from the pose search.

\noindent\textbf{Block mode in detail.} In 18 of the 68 blocks at least one slide without an accepted placement is shown as an empty image, and these blocks contain 64 of the 299 scored sections. Without them, block mode, given the release's stack alignment, loses 40 / 48 of the remaining 235 sections. In two blocks the refinement's gate decides differently for N than for R. Without these two blocks, block mode loses 46 / 51 of 292 sections.

\noindent\textbf{The controls.} W and U ask whether the loss comes from the withheld geometry or from the preparation. Our estimate of the release's warp is smooth, and a check that planted a piecewise deformation on development slides showed that it misses such a deformation. W, U, and the withheld deformation therefore carry~only the smooth part of the warp. A sharp local deformation, such as a crack that the warp closed, stays in the difference between W and R and is missing from the two geometry differences. In this appendix, geometry means the part of the withheld deformation that our estimate captures. Appearance means the preparation, together with the deformation that the estimate misses and the error of the estimate. A tool that does not estimate the warp compares W with the released image. Over the 297 scored sections with a fit, the two images have a median Dice overlap of 0.967 and a median contour distance of 0.301\,mm. On 22 of these 297 sections, the tiles of W do not match those of the released section. In block mode, given the release's stack alignment, W and U were solved in a second run, which the plan made conditional on a block-mode loss of N of at least 5 points on one seed. That run has its own R (Table~\ref{tab:native}). The two differences that pair N with W or U cross the two runs and are computed against the R of each run. The two runs of R differ in their joint outcome on 0 / 3 sections. Under the second reading the preparation is \emph{costly} in both modes. The plan then reads the loss of N as ``warp plus preparation, not separated'', with U minus R as its only reading of the geometry. Table~\ref{tab:native_diff} prints all five differences, and we quote no split and no cause.

\noindent\textbf{Further readings.} The plan asks for these readings, and none of them carries a branch. N lies below R in 19 of the 20 cells of hemisphere, mode, and seed, and in block mode on seed 0 it is 3 sections above R in hemisphere 5. At the plan's loosest bars of 20$^\circ$ and 5\,mm, the loss shrinks to 9 / 8 of the 299 sections in single-section mode and to 10 / 3 in block mode, given the release's stack alignment. Counting each of the 17 sections without an accepted placement as a miss of N instead, the loss at these bars is 16 / 14 sections in single-section mode and 23 / 16 in block mode. Giving each of them the outcome of R instead, the loss is 8 / 4 and 6 / 0 sections.  With the sections split into thirds by the size of the withheld deformation, the highest third loses more sections than the lowest in 4 of the 4 combinations of mode and seed. The loss rises or stays level from each third to the next in only 1 of them, and in single-section mode the middle third loses most on both seeds.

\noindent\textbf{How exact a rigid placement is.} A rigid placement of a slide that differs nonrigidly from its released section is a convention. Fits to all tissue, to the tiles, and to half sections place the tissue center within a median of 0.575\,mm of each other (95th percentile 1.207\,mm), and their spread exceeds 2\,mm on 2 of the 282 accepted placements. The in-plane angle spreads by a median of 1.177$^\circ$ (95th percentile 2.759$^\circ$).

\noindent\textbf{Disclosures.} The plan was frozen before the test's own placements were fitted and before any test slide was prepared or shown to the method. While the test was designed, read-only probes fitted between 26 and 100 test sections to their native slides to find out what a similarity leaves, and we measured the appearance of 10 test sections. The two pairing tools read every native test slide at 1/48 of its resolution. No accuracy of the method on native slides had been seen at that time, and every constant comes from development blocks. Three acceptance thresholds were loosened before the freeze, on development sections and with the limit of 15 in view. The second reading was added after the count of 17 was known. Two checks of the stored images changed after the freeze and before any pose was scored. The test scored the test sections once more and chose no setting of the method. The two counts at 20$^\circ$ and 5\,mm were computed after the test was read, under a reporting rule written before they were computed. 

\noindent\textbf{Not covered.} This test covers LFB sections, and its repeat on H\&E sections is the last paragraph of this appendix. It does not cover a block of native slides aligned without the MRI, or a placement that withholds the angle, the mirror state, or the scale. Nor does it cover an affine placement, which would separate stretch from the nonlinear part of the warp. It has no same-modality condition. No baseline of Table~\ref{tab:baselines_untouched} was run on native slides, and we compare the native slides with no baseline. The test gives no dose effect and no tolerance in millimeters, so it places no point on the curve of Figure~\ref{fig:adni}.

\input{tables/supp_native_he_paragraph}

%% file: tables/tab_native_given_rows.tex
Pairing of slide and section & not needed & from the release & from the release & from the release\\
Canvas size of a block & the release's & the release's & the release's & the release's\\
Pixel size & given & given & given & given\\
Mirror, angle, shifts, scale & in the image & fitted to the release & fitted & fitted to the release\\
Stretch, shear, nonlinear field & in the image & our estimate, reapplied & withheld & withheld\\
Tissue mask & NextBrain's & from the slide alone & NextBrain's & from the slide alone\\
Alignment within a block & the release's & ours, to the release & the release's rigid part & the release's rigid part, per section\\
\bottomrule

%% file: tables/tab_native_rows.tex
R & 0.702/0.726 & 0.866/0.856\\
W & 0.682/0.649 & 0.806/0.769\\
U & 0.612/0.579 & 0.753/0.759\\
N & 0.512/0.542 & 0.712/0.686\\
R, second block run &--& 0.866/0.853\\
\bottomrule

%% file: tables/tab_native_diff_rows.tex
Single section &&&&\\
\quad N minus R & $-19.1$/$-18.4$ & $-20.4$/$-19.1$ & $-18.4$/$-17.1$ & $<10^{-4}$/$<10^{-4}$\\
\quad U minus R & $-9.0$/$-14.7$ & $-10.7$/$-16.1$ & $-8.7$/$-14.0$ & 0.021/$<10^{-4}$\\
\quad N minus W & $-17.1$/$-10.7$ & $-18.4$/$-11.4$ & $-16.7$/$-10.4$ & $<10^{-4}$/0.0024\\
\quad W minus R & $-2.0$/$-7.7$ & $-3.7$/$-8.7$ & $-1.7$/$-6.7$ & 0.41/0.00099\\
\quad N minus U & $-10.0$/$-3.7$ & $-11.4$/$-4.3$ & $-9.7$/$-3.0$ & 0.00025/0.042\\
\midrule
Block mode &&&&\\
\quad N minus R & $-15.4$/$-17.1$ & $-18.4$/$-20.4$ & $-14.0$/$-16.4$ & 0.020/0.0049\\
\quad U minus R & $-11.4$/$-9.4$ & $-15.7$/$-12.7$ & $-11.4$/$-8.7$ & 0.028/0.032\\
\quad N minus W & $-9.4$/$-8.7^{*}$ & $-12.4$/$-12.0$ & $-9.4$/$-9.4$ & 0.064/0.068\\
\quad W minus R & $-6.0$/$-8.4$ & $-9.0$/$-11.0$ & $-4.7$/$-7.0$ & 0.30/0.063\\
\quad N minus U & $-4.0$/$-7.7$ & $-7.0$/$-11.0$ & $-2.7$/$-7.7$ & 0.46/0.14\\
\bottomrule

%% file: tables/tab_native_reads_rows.tex
Single section & Total loss & unresolved & breaks\\
 & Preparation & unresolved & costly\\
 & Split & unresolved & interaction\\
\midrule
Block mode & Total loss & unresolved & no branch (breaks, mixed)\\
 & Preparation & unresolved & costly\\
 & Split & unresolved & attributable\\
\bottomrule

%% file: tables/supp_native_he_paragraph.tex
\paragraph{H\&E native slides.} The H\&E native row of Table~\ref{tab:main} repeats this test with the H\&E slides of the same test set, under its own pre-registered plan and amendments. It uses 289 sections, the 300 test sections chosen on released images, minus 11 named ones. Each H\&E slide is a separate glass slide next to its LFB slide, scanned on its own, and takes its best rigid placement on its released H\&E section. Every condition, the released sections included, is given each section's in-plane angle. Its checks did not accept 19 of the 289 placements. In place of the LFB limit of 15, under which no question would have a branch, its plan set its own limits before any pose was computed, 40 such placements for single-section mode and 19 for block mode, and block mode met its limit with no margin. The released H\&E sections of the same runs reach 0.633 / 0.644 in single-section mode and 0.772 / 0.806 in block mode. Native slides reach 0.436 / 0.495 in single-section mode and 0.557 / 0.647 in block mode, given the release's stack alignment. They lose 19.7 and 14.9 points in single-section mode and 21.5 and 15.9 in block mode, given the release's stack alignment, with a sign-flip test over blocks giving $p<10^{-4}$ on both seeds in single-section mode and $p<0.01$ on both seeds in block mode. Applied to these losses, the plan's rule gives the total loss no branch in either mode, because its two counts of the 19 sections without an accepted placement give \emph{breaks} and \emph{mixed}. The preparation is \emph{costly} in both modes, so the plan reads the loss of the native slides as warp plus preparation, not separated. The two geometry differences lie within the plan's tolerance of each other on both seeds in single-section mode but not in block mode, given the release's stack alignment. The preparation and split readings rest on a warp estimate that under-reads deformation on H\&E development data, so they understate geometry and overstate appearance. The H\&E tissue mask is carried over from LFB unchecked. The threshold of the mirror check is LFB's, kept after the rule meant to set an H\&E threshold failed its own check, and planted warps turned the mirror choice on some development sections. An accepted H\&E placement lacks the label check of an LFB one, because no H\&E label maps exist. Errors are measured against the tissue of the released H\&E section, whose agreement with true H\&E tissue was not checked separately.

%% file: sections/09_statements.tex
\section{Data, ethics, and code release}\label{app:release}
NextBrain is a public dataset of five postmortem human hemispheres, released by its authors~\cite{casamitjana2025nextbrain}. Its record on the UCL Research Data Repository is published under CC0~1.0. This record points to the GitHub repositories that we use. It also contains the scans of the glass slides that the tests of Appendix~\ref{app:native} read. The test on native LFB slides also reads the record's label maps in slide space.   NextBrain's authors report that the hemispheres were used following informed consent to use the tissue for research. They also report that the brain donation program and its protocols have received ethical approval for research from the National Research Ethics Service Committee London - Central. They add that the tissue is stored for research under a license issued by the Human Tissue Authority.  ADNI data are used under the ADNI data-use agreement. ADNI's study protocols require each participating site's institutional review board or research ethics board to approve the protocol and its consent documents. They also require written informed consent from every participant or an authorized representative. Every ADNI site's consent documents state that de-identified data will be shared with the scientific community for research. Appendices~\ref{app:nhbench} and~\ref{app:euregbench} also use the public Needles \& Haystacks benchmark data~\cite{frolov2025needles} and the $\mu$-RegPro prostate data~\cite{baum2023muregpro}.  The first is released under a CC~BY~4.0 license, and its authors report that both of its sources of human scans were approved by ethics committees and anonymized. The second comes from the SmartTarget biopsy trial at University College London Hospital. The $\mu$-RegPro data are released under a \mbox{CC BY-NC-SA 4.0} license for research use only, as the trial's ethics approval and patient consent require.  We used only these released data and collected no data from human subjects ourselves, so no ethics review of our own was carried out.  The method is a research tool, not a clinical device.

We will release the code and the trained two-stain dual encoders of Table~\ref{tab:main}, together with the lists of evaluated sections and the pre-registered decision rules of Appendix~\ref{app:rules}.